\pdfoutput=1

\documentclass[11pt]{article}

\usepackage[final]{acl}

\usepackage{times}
\usepackage{latexsym}
\usepackage{graphicx}
\usepackage{xspace, soul}
\usepackage{threeparttable}
\usepackage{booktabs}
\usepackage{multirow}
\usepackage{amsmath}
\usepackage{cleveref}
\usepackage{pifont}
\usepackage{amssymb,enumitem}
\usepackage{comment}
\usepackage{hyperref,graphicx}
\usepackage{marvosym}
\usepackage{booktabs, makecell, rotating, siunitx}
\usepackage{tcolorbox}  
\usepackage{listings}   
\usepackage{xcolor}     
\usepackage{colortbl}
\usepackage{subcaption}
\usepackage{longtable} 
\usepackage{pdflscape} 
\usepackage{tcolorbox}
\usepackage[T1]{fontenc}

\usepackage[utf8]{inputenc}

\usepackage{microtype}

\usepackage{inconsolata}

\usepackage{graphicx}

\title{\mollm: Generalizing Large Language Models for Multi-property Molecule Optimization}

\author{
 \textbf{Vishal Dey\textsuperscript{1}$^*$},
 \textbf{Xiao Hu\textsuperscript{1}$^*$},
 \textbf{Xia Ning\textsuperscript{1,2,3,4}}
\\
 \textsuperscript{1} Department of Computer Science
and Engineering, The Ohio State University, USA \\
 \textsuperscript{2} Translational Data Analytics Institute, The Ohio State University, USA \\
 \textsuperscript{3}Department of Biomedical Informatics, The Ohio State University, USA \\
 \textsuperscript{4} College of Pharmacy, The Ohio State University, USA
\\
 \small{
   \textbf{Correspondence:} \href{mailto:email@domain}{ning.104@osu.edu}
 }
}

\newcommand{\cmark}{\ding{51}}%
\newcommand{\xmark}{\ding{55}}%
\definecolor{lightgray}{gray}{0.9} 
\definecolor{lightergray}{gray}{0.95} 

\newcommand{\DrugMOpt}{\mbox{$\mathop{\mathtt{MolOpt\text{-}Instructions}}\limits$}\xspace}
\newcommand{\OpenMolIns}{\mbox{$\mathop{\mathtt{OpenMolIns}}\limits$}\xspace}
\newcommand{\MOptData}{\mbox{$\mathop{\mathtt{MuMOInstruct}}\limits$}\xspace}
\newcommand{\GTO}{\mbox{$\mathop{\mathtt{GT}}\limits$}\xspace}
\newcommand{\CSO}{\mbox{$\mathop{\mathtt{CS}}\limits$}\xspace}

\newcommand{\mollm}{\mbox{$\mathop{\mathtt{GeLLM^3O}}\limits$}\xspace}

\newcommand{\mollmTripleTask}{\mbox{$\mathop{\mathtt{GeLLM^3O\text{-}{3}}}\limits$}\xspace}
\newcommand{\mollmQuadTask}{\mbox{$\mathop{\mathtt{GeLLM^3O\text{-}{4}}}\limits$}\xspace}
\newcommand{\mollmTripleGen}{\mbox{$\mathop{\mathtt{GeLLM^3O\text{-}P(3)}}\limits$}\xspace}
\newcommand{\mollmQuadGen}{\mbox{$\mathop{\mathtt{GeLLM^3O\text{-}P(4)}}\limits$}\xspace}
\newcommand{\mollmSixGen}{\mbox{$\mathop{\mathtt{GeLLM^3O\text{-}P(6)}}\limits$}\xspace}

\newcommand{\mollmTripleTaskM}{\mbox{$\mathop{\mathtt{GeLLM^3O\text{-}{3}_{Mistral}}}\limits$}\xspace}
\newcommand{\mollmQuadTaskM}{\mbox{$\mathop{\mathtt{GeLLM^3O\text{-}{4}_{Mistral}}}\limits$}\xspace}

\newcommand{\mollmTripleTaskL}{\mbox{$\mathop{\mathtt{GeLLM^3O\text{-}{3}_{Llama}}}\limits$}\xspace}
\newcommand{\mollmQuadTaskL}{\mbox{$\mathop{\mathtt{GeLLM^3O\text{-}{4}_{Llama}}}\limits$}\xspace}

\newcommand{\mollmTripleGenM}{\mbox{$\mathop{\mathtt{GeLLM^3O\text{-}P(3)_{Mistral}}}\limits$}\xspace}
\newcommand{\mollmQuadGenM}{\mbox{$\mathop{\mathtt{GeLLM^3O\text{-}P(4)_{Mistral}}}\limits$}\xspace}
\newcommand{\mollmSixGenM}{\mbox{$\mathop{\mathtt{GeLLM^3O\text{-}P(6)_{Mistral}}}\limits$}\xspace}
\newcommand{\mollmTripleGenL}{\mbox{$\mathop{\mathtt{GeLLM^3O\text{-}P(3)_{Llama}}}\limits$}\xspace}
\newcommand{\mollmQuadGenL}{\mbox{$\mathop{\mathtt{GeLLM^3O\text{-}P(4)_{Llama}}}\limits$}\xspace}
\newcommand{\mollmSixGenL}{\mbox{$\mathop{\mathtt{GeLLM^3O\text{-}P(6)_{Llama}}}\limits$}\xspace}
\newcommand{\BDQ}{\mbox{$\mathop{\mathtt{BDQ}}\limits$}\xspace}
\newcommand{\BPQ}{\mbox{$\mathop{\mathtt{BPQ}}\limits$}\xspace}
\newcommand{\BDP}{\mbox{$\mathop{\mathtt{BDP}}\limits$}\xspace}
\newcommand{\DPQ}{\mbox{$\mathop{\mathtt{DPQ}}\limits$}\xspace}
\newcommand{\BDPQ}{\mbox{$\mathop{\mathtt{BDPQ}}\limits$}\xspace}
\newcommand{\MPQ}{\mbox{$\mathop{\mathtt{MPQ}}\limits$}\xspace}
\newcommand{\BMPQ}{\mbox{$\mathop{\mathtt{BMPQ}}\limits$}\xspace}
\newcommand{\HMPQ}{\mbox{$\mathop{\mathtt{HMPQ}}\limits$}\xspace}
\newcommand{\BDMQ}{\mbox{$\mathop{\mathtt{BDMQ}}\limits$}\xspace}
\newcommand{\BHMQ}{\mbox{$\mathop{\mathtt{BHMQ}}\limits$}\xspace}

\newcommand{\PMol}{\mbox{$\mathop{\mathtt{Prompt\text{-}MolOpt}}\limits$}\xspace}

\newcommand{\LlaSMol}{\mbox{$\mathop{\mathtt{LlaSMol}}\limits$}\xspace}
\newcommand{\LlaSMolM}{\mbox{$\mathop{\mathtt{LlaSMol_{Mistral}}}\limits$}\xspace}
\newcommand{\ChemLLM}{\mbox{$\mathop{\mathtt{ChemLLM}}\limits$}\xspace}

\newcommand{\SR}{\mbox{$\mathop{\mathtt{SR}}\nolimits$}}
\newcommand{\Val}{\mbox{$\mathop{\mathtt{Val}}\nolimits$}}
\newcommand{\SAS}{\mbox{$\mathop{\mathtt{SAS}}\nolimits$}}
\newcommand{\Sim}{\mbox{$\mathop{\mathtt{Sim}}\nolimits$}}
\newcommand{\Nov}{\mbox{$\mathop{\mathtt{Nov}}\nolimits$}}

\newcommand{\RI}{\mbox{$\mathop{\mathtt{RI}}\nolimits$}}
\newcommand{\APS}{\mbox{$\mathop{\mathtt{APS}}\nolimits$}}
\newcommand{\APSTr}{\mbox{$\mathop{\mathtt{APS_{train}}}\nolimits$}}

\newcommand{\MPSTr}{\mbox{$\mathop{\mathtt{MPS_{train}}}\nolimits$}}
\newcommand{\deltaTr}{$\Delta_\mathtt{train}$}
\newcommand{\APSTs}{\mbox{$\mathop{\mathtt{APS_{test}}}\nolimits$}}

\newcommand{\ImpT}{\mbox{$\mathop{\mathtt{Impv\text{-}Task}}\limits$}\xspace}
\newcommand{\ImpG}{\mbox{$\mathop{\mathtt{Impv\text{-}Gen}}\limits$}\xspace}

\begin{document}

\maketitle
\def\thefootnote{*}\footnotetext{Equal Contribution}\def\thefootnote{\arabic{footnote}}

\setlength{\fboxsep}{1pt} 

\begin{abstract}
%
Despite recent advancements,
most computational methods for molecule optimization are constrained
to single- or double-property optimization tasks
and
suffer from poor scalability and generalizability to novel optimization tasks.
Meanwhile, Large Language Models (LLMs) demonstrate
remarkable out-of-domain generalizability to novel tasks.
To demonstrate LLMs' potential for molecule optimization,
we introduce \MOptData, 
the first high-quality instruction-tuning dataset specifically
focused on complex multi-property molecule optimization tasks. 
Leveraging \MOptData, we develop {\mollm}s, 
a series of instruction-tuned LLMs
for molecule optimization.
Extensive evaluations across 5 in-domain and 5 out-of-domain
tasks demonstrate that {\mollm}s 
consistently outperform state-of-the-art baselines. 
{\mollm}s also exhibit outstanding zero-shot generalization to unseen tasks, significantly outperforming powerful closed-source LLMs.
Such strong generalizability
demonstrates the tremendous potential of {\mollm}s 
as foundational models for molecule optimization,
thereby tackling novel optimization tasks without resource-intensive retraining. 
\MOptData, models, and code are accessible through 
\url{https://github.com/ninglab/GeLLMO}.

\end{abstract}

\section{Introduction}
\label{sec:intro}

Drug discovery is a costly and
time-consuming process, costing over
\$2 billion and a decade~\cite{sertkaya2024costs}.
One of the most critical stages~\cite{Hughes2011principles}
in this process is
lead optimization~\cite{Sun2022why},
where a molecule with promising bioactivity against a drug target
is optimized into a lead molecule 
%
%
by improving multiple molecular properties simultaneously.
For example, a hit molecule to treat schizophrenia is optimized such that
it can permeate the blood-brain barrier~\cite{Pollak2018} to reach the 
DRD2 target~\cite{Seeman2006} in the brain,
while balancing lipophilicity, solubility
and toxicity. 
Improving all these properties together requires
balancing multiple trade-offs~\cite{Nicolaou2013} and conflicting objectives~\cite{Nicolotti2011}, making
multi-property optimization extremely challenging.

Most computational methods~\cite{gao2022sample} 
for molecule optimization
focus on single- or double-property tasks,
leaving multi-property optimization tasks largely unexplored.
%
Existing approaches~\cite{kim2024genetic,yang2021hit}
use predefined fitness and reward functions~\cite{LUUKKONEN2023}, respectively,
to model property trade-offs.
However, designing such functions for each task 
demands significant effort and domain expertise.
%
In contrast, other methods~\cite{chen2021deep,wu2024leveraging}
obviate the need for such functions,
but depend on scarce task-specific data, limiting their scalability and adaptability.
%
%
%
Additionally, existing methods lack generalization to
unseen tasks, hindering their practical applicability
to emerging therapeutic requirements.






Large language models (LLMs)~\cite{naveed2023comprehensive}
have demonstrated remarkable generalization to unseen tasks 
across diverse domains~\cite{chang2024} recently. 
However, their potential
in challenging, multi-property molecule optimization tasks
remains largely unidentified.
To fully identify LLMs' potential,
we introduce \MOptData, the first 
high-quality instruction-tuning dataset 
specifically focused on complex, multi-property
tasks, each aiming to improve at least 3 properties simultaneously.
This is in stark contrast to
existing instruction-tuning datasets~\cite{ye2025drugassist}
that are limited to single- and double-property tasks.
%

\begin{figure*}[h!]
    \centering
    \includegraphics[width=0.95\textwidth]{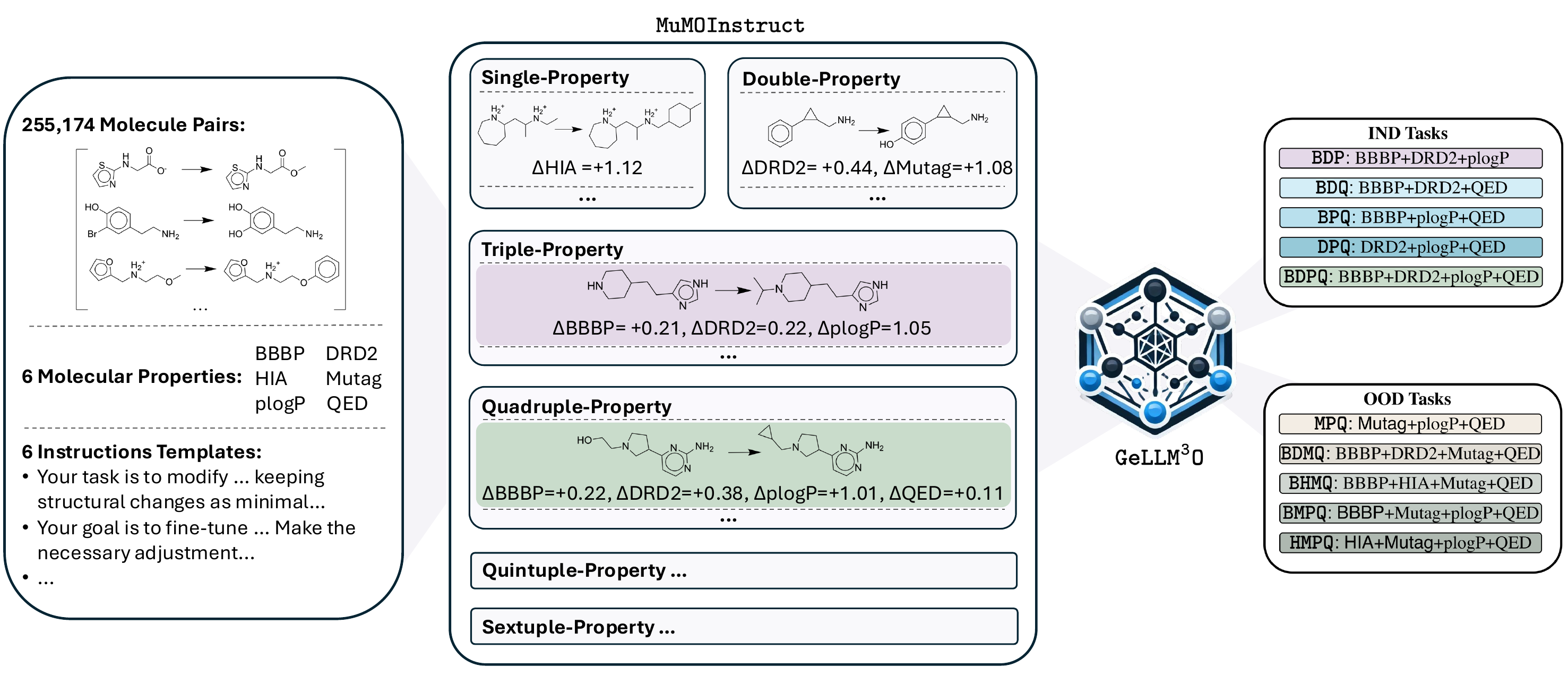}
    \caption{Overview of \MOptData and \mollm}
    \label{fig:Overview}

\vspace{-15pt}
\end{figure*}

Leveraging \MOptData, 
we develop a series of
\underline{Ge}neralizable \underline{LLM}s for 
\underline{M}ulti-property \underline{M}olecule \underline{O}ptimization, denoted 
as {\mollm}s,
by instruction-tuning general-purpose LLMs.
%
Task-specific {\mollm}s are fine-tuned on individual tasks,
learning precise optimization tailored to specific therapeutic contexts.
Generalist {\mollm}s, on the other hand, are fine-tuned on multiple tasks 
which enables them to learn and reason property trade-offs across diverse therapeutic contexts.
%
%
Moreover, fine-tuning using diverse tasks and instructions
enables generalist {\mollm}s to 
effectively handle unseen tasks
and instructions.

All {\mollm} models are extensively
evaluated against strong general-purpose LLMs,
state-of-the-art foundational LLMs for chemistry and
task-specific non-LLMs
across 5 in-domain (IND) and 5 out-of-domain (OOD) tasks.
Our experimental results demonstrate the following key findings:

\textbf{(1)} Both task-specific and generalist {\mollm}s significantly outperform state-of-the-art baselines, 
including powerful closed-source LLMs,
across all IND 
and OOD tasks, 
with significant improvements of up to 186.6\% over the best baselines.

\textbf{(2)} Compared to task-specific {\mollm}s, 
generalist {\mollm}s excel on 3 out of 5 IND tasks and
demonstrate competitive performance on the other 2 tasks,
with remarkable gains of up to 91.3\% 
on more complex tasks, such as \BDPQ.

\textbf{(3)} Generalist \mollmSixGen models show superior
generalization to OOD tasks 
outperforming strong baselines
by as much as 159.9\%.

Figure~\ref{fig:Overview} presents the overall scheme of \mollm.
To the best of our knowledge,
\MOptData is the first large-scale, high-quality instruction-tuning dataset specifically designed
for multi-property molecule optimization.
%
%
Notably, the strong generalization ability of our generalist
{\mollm}s demonstrates their tremendous potential
to accelerate drug discovery by tackling novel optimization tasks
without resource-intensive retraining.
%
Dataset, models, and code are accessible
through \url{https://github.com/ninglab/GeLLMO}.

    
    

\section{Related Work}
\label{sec:rel}
%
Various computational approaches have been developed for molecule optimization~\cite{you2018graph, blaschke2020reinvent, xie2021mars, bung2022silico,sun2022molsearch}.
For example, Modof~\cite{chen2021deep}, MIMOSA~\cite{fu2021mimosa},
and f-RAG~\cite{lee2024molecule} 
perform substructure modifications over molecular graphs.
%
Chemformer~\cite{irwin2022chemformer} and \PMol~\cite{wu2024leveraging} treat optimization as a translation over SMILES~\cite{Weininger1988smiles} strings, 
and learn the required modification from molecule pairs.
%
GraphGA~\cite{jensen2019graph} and MolLeo~\cite{wang2024efficient} leverage 
genetic algorithms
to evolve molecules via genetic algorithm. 
%
These methods~\cite{kim2024genetic,yang2021hit} often require designing non-trivial fitness or reward functions
to capture nuanced trade-offs among multiple properties~\cite{niu2024tradingoff}. 
%
%
Moreover, such methods 
tends to generate
molecules with entirely new scaffolds, 
limiting their applicability \emph{in vitro} optimization.

Recently, LLMs~\cite{chang2024} have emerged as a promising option for molecule optimization. 
For example,
ChatDrug~\cite{liu2023chatgpt} and Re3DF~\cite{le2024utilizing} leverage 
LLMs to optimize a molecule iteratively
through multi-turn dialogues.
%
DrugAssist~\cite{ye2025drugassist}
instruction-tuned Llama2-7B-Chat~\cite{touvron2023llama} 
on each optimization task. 
%
While these approaches offer flexible 
task formulation through natural language, they still face
several limitations.
ChatDrug incurs high costs due to multiple API calls, 
and
instruction-tuning in DrugAssist relies on task-specific data,
limiting scalability and adaptability to more complex multi-property tasks.
\section{\MOptData Dataset}
\label{sec:data}

\begin{table}[t!]
\centering
\caption{Comparison among instruction-tuning datasets for molecular optimization}
\label{tbl:relwork_data}
\vspace{-5pt}
\begin{threeparttable}
\begin{small}
\begin{tabular}{
    @{\hspace{1pt}}p{2.2cm}@{\hspace{0pt}}
    @{\hspace{0pt}}c@{\hspace{1pt}}
    @{\hspace{0pt}}c@{\hspace{1pt}}
    @{\hspace{2pt}}c@{\hspace{1pt}}
}
\toprule
\multirow{2}{*}{Comparison} 
& \scriptsize{\DrugMOpt}
& \scriptsize{\OpenMolIns}
& \scriptsize{\MOptData} \\
& \scriptsize{\cite{ye2025drugassist}} 
& \scriptsize{\cite{li2025tomg}} & \scriptsize{(ours)} \\
\midrule

Realistic tasks & \xmark & \xmark & \cmark \\
{Multi-prop opt} & \cmark & \xmark & \cmark
\\
\# properties & 5 & 3 & 6 \\
\# molecule pairs & 1,029,949 & 1,200,000 & 255,174 \\
\# Total tasks & 8 & 3 & 63 \\
~~\# Train $\ge3$ prop & 0 & 0 &  42 \\
~~\# Eval $\ge3$ prop & 0 & 0 & 10 \\
~~~~~\# IND & 0 & 0 & 5 \\
~~~~~\# OOD & 0 & 0 & 5\\

\bottomrule
\end{tabular}

\end{small}
\end{threeparttable}
\vspace{-10pt}
\end{table}

\paragraph{Comparison among \DrugMOpt and \MOptData:}
To address these gaps, we introduce \MOptData, the first
instruction-tuning dataset specifically focusing on
realistic multi-property optimization tasks.
%
%
Different from \DrugMOpt~\cite{ye2025drugassist},
which focuses on single- and double-property tasks, 
\MOptData emphasizes
on tasks with at least 3 properties for evaluating LLMs
in in-domain and out-of-domain settings.
Table~\ref{tbl:relwork_data} highlights the notable
differences between the two datasets.
%

\textbf{Problem Definition:} 
A molecule optimization task is to transform a hit $M_x$ --  a molecule exhibiting initial bio-activity
against a therapeutic target into a lead molecule $M_y$ – an improved molecule for drug development, 
through structural modification over $M_x$, such that 
\textbf{(1)} $M_y$ is structurally similar to $M_x$ (similarity constraint), and 
\textbf{(2)} $M_y$ is better than $M_x$ in terms of all desired properties of interest 
(property constraint). 
The desirability of a property is determined by the therapeutic goal, where
improved properties indicate more suitable toward a successful drug candidate. 
For example, for drug candidates targeting the central nervous system (CNS), 
higher blood-brain barrier permeability (BBBP) is desired to allow the molecules 
to reach the brain or spinal cord, 
whereas for those targeting the peripheral nervous system (PNS), 
lower BBBP is desired instead to prevent the drugs from damaging the CNS.
Under the property constraint, the molecule pair $(M_x$, $M_y)$ 
is represented as $(M_x \prec_{\Delta_p} M_y)_{\forall p \in P}$, 
indicating that $M_y$ is better than $M_x$ on each property $p$ of all the desired 
properties $P$ with a property-specific difference $\Delta_p$. 
In this paper, we introduce \MOptData,
the first high-quality
instruction-tuning dataset to 
evaluate
models in molecule optimization tasks.
Our design is based on four key principles:
%
\textbf{(1) Pairwise optimization:} 
\MOptData contains a comprehensive set of molecule pairs
satisfying the similarity constraint (Tanimoto similarity > 0.6) and property constraint over 
multiple desired properties.
Such molecule pairs enable opportunities for molecule optimization models 
to learn the association between the structural differences 
and the property improvement among the pairing molecules, 
and apply such associations for new lead optimization.
%
%
%
%
\textbf{(2) Comprehensive coverage:}
\MOptData covers more molecular properties,
and extends beyond single- and double-property tasks in
existing molecular optimization benchmarks. 
It introduces multi-property optimization
tasks that require simultaneous improvement of at least 3 properties,
thereby representing complex
pharmacological trade-offs in lead optimization.
%
%
%
\textbf{(3) Real-world relevance:}
The tasks in \MOptData
are carefully constructed to
represent 
realistic challenges in lead optimization
by combining molecular properties key to drug development.
For instance, one of the tasks aims to improve
intestinal adsorption,
toxicity and BBBP -- key
properties for optimizing orally bioavailable CNS drugs.
%
%
%
\textbf{(4) Diverse instructions:}
\MOptData provides diverse natural language instructions,
each describing the optimization task using
different phrasings.
This prevents LLMs instruction-tuned on \MOptData
from overfitting to a specific phrasing
and thus, enables them to generalize to unseen instructions, 
which is crucial in practice to allow different descriptions on optimization tasks.
%


\begin{table*}[h!]
\centering
\caption{Summary of \MOptData Tasks for Evaluation 
}
\label{tbl:task_summary}
\vspace{-5pt}
\begin{small}
\begin{threeparttable}
\begin{tabular}{
    @{\hspace{2pt}}l@{\hspace{5pt}}
    @{\hspace{2pt}}l@{\hspace{2pt}}
    @{\hspace{2pt}}c@{\hspace{2pt}}
    @{\hspace{2pt}}c@{\hspace{2pt}}
    @{\hspace{2pt}}c@{\hspace{2pt}}
    @{\hspace{2pt}}c@{\hspace{2pt}}
    @{\hspace{2pt}}c@{\hspace{2pt}}
    @{\hspace{2pt}}c@{\hspace{2pt}}
    @{\hspace{2pt}}r@{\hspace{2pt}}
    @{\hspace{2pt}}r@{\hspace{2pt}}
    @{\hspace{2pt}}r@{\hspace{2pt}}
    @{\hspace{2pt}}r@{\hspace{4pt}}
    @{\hspace{2pt}}c@{\hspace{2pt}}
}
\toprule
\multirow{2}{*}{Type}
& \multirow{2}{*}{Task} 
& \multicolumn{6}{c}{Properties $(\Delta_p)$} 
& \multirow{2}{*}{\#Train}
& \multirow{2}{*}{\#Val}
& \multirow{2}{*}{\#Test}
& \multirow{2}{*}{\#Mols}
& \multirow{2}{*}{Cat}
\\
\cmidrule(){3-8}
& & BBBP{$^\uparrow$}(0.2) & DRD2{$^\uparrow$}(0.2) & HIA{$^\uparrow$} (0.1) & Mutag$^\downarrow$(0.1) & plogP$^\uparrow$(1.0) & QED$^\uparrow$(0.1) 
\\
\midrule
\multirow{5}{*}{IND}
& \BDP 
& \cmark & \cmark & - & - & \cmark & -
& {2,064} & {230} & {500} & {2,449} & \CSO
\\
& \BDQ 
& \cmark & \cmark & - & - & - & \cmark
& {4,472} & {497} & {500} & {4,614}  & \CSO
\\
& \BPQ 
& \cmark & - & - & - & \cmark & \cmark
& {4,048} & {450} & {500} & {6,953}& \CSO
\\
& \DPQ 
& - & \cmark & - & - & \cmark & \cmark
& {2,114} & {235} & {500} & {2,589} & \CSO
\\
& \BDPQ 
& \cmark & \cmark & - & - & \cmark & \cmark
& {624} & {70} & {500} & {802} & \CSO
\\
\hline
\multirow{5}{*}{OOD}
& \MPQ 
& - & - & - & \cmark & \cmark & \cmark
& {3,132} & {349} & {500} & {5,384} & \GTO
\\
& \BDMQ 
& \cmark & \cmark & - & \cmark & - & \cmark
& {601} & {67} & {500} & {791} & \CSO
\\
& \BHMQ 
& \cmark & - & \cmark & \cmark & - & \cmark
& {191} & {22} & {118} & {333} & \CSO
\\
& \BMPQ 
& \cmark & - & - & \cmark & \cmark & \cmark
& {373} & {42} & {191} & {690} & \CSO
\\
& \HMPQ 
& - & - & \cmark & \cmark & \cmark & \cmark
& {234} & {26} & {96} & {417} & \GTO
\\

\bottomrule
\end{tabular}

\begin{tablenotes}[normal,flushleft]
\footnotesize
\item ``\#Train", ``\#Val", ``\#Test", ``\#Mols" denote the number of training pairs, validation pairs, test samples, and unique molecules in each task, respectively.
``Type" indicates task types, including IND and OOD tasks.
``Cat" indicates task category.
\end{tablenotes}

\end{threeparttable}
\end{small}
\vspace{-10pt}
\end{table*}

\subsection{Overview of \MOptData Tasks}
\label{sec:data:overview}


{\MOptData comprises 63 tasks, 
with 42 tasks aiming to improve at least 3 properties simultaneously, 
out of which 10 tasks are further divided into IND and OOD tasks (Section~\ref{sec:data:ind_ood}).
}
%
All tasks in \MOptData are systematically designed
by considering combinations from 6
molecular properties:
%
\textbf{(i) Penalized LogP (plogP)} representing lipophilicity,
balancing permeability, 
solubility,
and metabolic stability
-- higher plogP is generally desired
in drug development;
\textbf{(ii) Quantitative Estimate of Drug-Likeness (QED)}
assessing overall drug-likeness,
incorporating multiple molecular attributes
related to molecular weight, lipophilicity, and solubility
-- higher QED indicates better drug-likeliness;
\textbf{(iii) Blood-Brain Barrier Permeability (BBBP)} which
refers to the ability of a drug to cross the blood-brain barrier
-- higher BBBP is desired for CNS drug candidates; 
%
\textbf{(iv) Mutagenicity (Mutag)} indicating the likelihood of a drug
causing genetic mutations --
lower Mutag scores are desired to reduce toxicity;
\textbf{(v) Human Intestinal Absorption (HIA)}
which reflects a drug's ability to be absorbed
through the gastrointestinal tract --
higher HIA is desired for orally administered drugs;
and
\textbf{(vi) Dopamine Receptor D2 (DRD2) binding affinity}
representing the ability of drugs to target dopaminergic pathways
-- higher DRD2 scores are desired for drugs
targeting the DRD2 receptor.
%

We focus on these properties because they 
are critical in influencing the
pharmacokinetics and safety profiles of molecules,
thereby contributing to the development of successful drug candidates.
Additionally, these properties are well-studied in the literature
and are used in existing benchmarks~\cite{gao2022sample}.
%
%
%
{\MOptData provides 10 evaluation tasks 
%
which are 
summarized in Table~\ref{tbl:task_summary}
with details in Appendix~\ref{sec:app:task}.}
These tasks can be categorized into two groups based on
their significance:
\textbf{(1) General Drug-Likeness and Toxicity Optimization (\GTO):}
Focuses on widely studied molecular properties related to 
drug-likeness, absorption, and toxicity,
that are general to any successful drug candidates.
%
\textbf{(2) Context-specific Optimization (\CSO):}
%
Includes properties 
relevant to specific disease contexts and therapeutic requirements.
\subsection{Creation of Task-Specific Training Pairs}
\label{sec:data:pairs:collect}


%
We construct task-specific training pairs,
where each pair 
$(M_x, M_y)$ is sourced
from the dataset provided by \citet{chen2021deep},
which consists of 
255K molecule pairs dervied from 331K molecules.
%
Each pair
differs at only one disconnection site,
meaning $M_x$ can be transformed to $M_y$ by modifying exactly one fragment.
%
%
%
{Among these molecule pairs,
we select those
that satisfy all $P$
property constraints for a given task optimizing $P$ properties
(i.e., $(M_x \prec_{\Delta_p} M_y)_{\forall p \in P}$).
This ensures that
the hit molecule $M_x$ in each pair
requires substantial optimizations, making
the selected pairs
suitable to model realistic optimization tasks.
}
\subsection{Creation of Task-Specific Test Set}
\label{sec:data:test}

We construct a test set 
%
by randomly sampling an initial pool of 
250K molecules from the ZINC database~\cite{sterling2015zinc}
-- a collection of commercially available drug-like molecules --
that are not included in the training set.
Out of this pool,
we select a molecule into the test set of a task
which has a property worse than 
the median among
all $M_x$ in the training pairs 
(i.e., median property scores of $M_x$ denoted as \MPSTr)
for each desired property.
%
This 
provides a task-specific, data-driven selection criteria
that is robust to outliers.
Additional criteria to exclude outliers are detailed in Appendix~\ref{sec:app:filter}.
After applying these steps to the initial pool of 250K molecules,
we randomly select at most 500 molecules into the test set for each task, with possible overlap across tasks.
Table~\ref{tbl:task_prop} presents the task-specific data set characteristics.

%
%

%
%

%

\subsection{Quality Control}
We implement multiple quality control measures as detailed in
Appendix~\ref{sec:app:quality}.
We remove duplicate molecules based on
canonicalized SMILES strings.
For each molecule, we compute
empirical property scores using
well-established tools:
ADMET-AI~\cite{swanson2024admet}
and the official implementation provided by \citet{you2018graph}.
Additionally, we provide 6 distinctly phrased (i.e., diverse) instructions for each task (Appendix~\ref{sec:app:instr}).
To evaluate LLMs' instruction understanding and generalizability
to unseen instructions, 
we hold out one instruction for each task during training.

\subsection{IND and OOD Tasks}
\label{sec:data:ind_ood}
To distinctly assess the capabilities of instruction-tuned LLMs 
on both familiar
and novel optimization tasks,
we categorize our tasks into two groups:

\paragraph{In-Distribution (IND) tasks:}
IND tasks only have property combinations that
are used during training.
Success in these tasks provides insights
on the models' ability to handle optimizations
they are specifically trained on.
%

\paragraph{Out-of-Distribution (OOD) tasks:}
OOD tasks 
include novel property combinations
that are not used during training.
Note that OOD is defined in terms of property combinations
and not individual properties,
that is, each property
is included
as part of training tasks.
OOD tasks evaluate the models' ability to generalize
to novel optimization tasks without task-specific retraining.
This generalizability
is crucial in real-world lead optimization, 
where new therapeutic requirements frequently arise.
%
%

\section{\mollm Models}
\label{sec:model}

We introduce \mollm, a series of general-purpose LLMs
instruction-tuned over \MOptData.
%
%
%
Through instruction tuning, 
\mollm implicitly learns chemical semantics,
structure-property relationships (SPR)~\cite{hansch1969quantitative}
and associations between 
structural differences expressed in molecule pairs and
the desired property improvement
expressed via natural language instruction.
\mollm applies this knowledge
to perform structural modifications on a given molecule
and generate
better molecules with improved
properties. 
{
Notably, unlike existing methods that rely on carefully designed 
reward or fitness functions to balance trade-offs among
multiple properties~\cite{niu2024tradingoff},
\mollm captures such trade-offs implicitly by learning from
multiple paired molecules across diverse multi-property objectives.
}

%
%

We develop both task-specific and generalist {\mollm}s. 
%
Task-specific models
are trained on a single optimization task, 
and thus benefit from dedicated training tailored to that specific task. 
%
In contrast, generalist models are trained across 
multiple optimization tasks simultaneously.
This multi-task training enables cross-task knowledge transfer,
allowing the generalist \mollm to leverage shared
chemical knowledge on SPR
and multi-property trade-offs across all possible property combinations.
Thus, the generalist \mollm
represents a step toward a foundational model for molecule optimization,
capable of handling diverse tasks without task-specific retraining.
%

%

We develop a series of generalist {\mollm}s
trained on the power sets of 3, 4, and 6 properties,
denoted as \mollmTripleGen, \mollmQuadGen, and \mollmSixGen,
respectively.
{Importantly, the training data for {\mollmSixGen} excludes 
all property combinations that appear in the OOD tasks.
This ensures that OOD tasks represent unseen optimization scenarios during inference.}
To train these models, we fine-tune 2 general-purpose LLMs: 
Mistral-7B-Instruct-v0.3~\cite{mistral2023mistral} and Llama3.1-8B-Instruct~\cite{grattafiori2024llama3herdmodels}
by applying LoRA~\cite{hu2022lora}
adapters to all projection layers and the language modeling head. 
We perform 0-shot evaluations
(i.e., without in-context examples) for all {\mollm}s 
in all tasks.
For each test molecule, we generate 20 molecules 
via beam search decoding, with the number of beams set to 20.
Training details are provided
in Appendix~\ref{sec:app:reproducibility}.

\section{Experimental Setup}
\label{sec:expt}
\subsection{Baselines}
\label{sec:expt:baselines}

We compare {\mollm}s against 3 categories of baseline models: 
\textbf{(1)} general-purpose LLMs: Mistral-7B Instruct-v0.3~\cite{mistral2023mistral}, 
Llama-3.1 8B-Instruct~\cite{touvron2023llama}, 
Claude-3.5~\cite{anthropic2024claude}, {and GPT-4o~\cite{openai2024gpt4technicalreport}};
\textbf{(2)} foundational LLMs for chemistry: 
{an Intern-LM2 with 7B parameters tuned on chemistry and molecular tasks, 
denoted as \ChemLLM~\cite{zhang2024chemllm},}
and
\LlaSMol tuned on Mistral-7B, denoted as \LlaSMolM~\cite{yu2024llasmol},
and \textbf{(3)} task-specific non-LLMs: \PMol~\cite{wu2024leveraging}.
%
Similarly to {\mollm}s, we generate 20 molecules for each input molecule for all baselines. 
For LLM baselines that support beam search decoding, 
we use the same generation strategy
as for {\mollm}s.
Experimental setups are detailed in Appendix~\ref{sec:app:expts_setup:baselines}.
Prompt templates for LLMs are in Appendix~\ref{sec:app:prompt}.
Discussions on \PMol and DeepSeek-R1 are in Appendix~\ref{sec:app:baselines} and \ref{sec:app:deepseek}, respectively.

\subsection{Evaluation Metrics}
\label{sec:expt:eval}

We employ multiple evaluation metrics 
(detailed in Appendix~\ref{sec:app:eval})
for a holistic comparison.
For brevity and clarity,
we present the results only in terms of:
\textbf{(1) Success Rate} ({\SR})
which is the proportion of input molecules that
are successfully optimized with improvement in all
desired properties;
\textbf{(2) Similarity with input} ({\Sim})
which denotes the average Tanimoto similarity~\cite{Bajusz2015why} between the
optimized and the corresponding input molecule;
and 
\textbf{(3) Relative Improvement} ({\RI})
representing the average change in each desired property
relative to its initial value in the input molecule.
{
Formally, for a task improving $\mathcal{P}$ properties, {\RI} 
is computed as the average of
relative change (\RI$_p$) in each property $p \in \mathcal{P}$ as:
\begin{equation*}
 \RI = \frac{\sum_{\scriptsize{p \in \mathcal{P}}} \RI_p}{|\mathcal{P}|},
\end{equation*}
where \RI$_p$ is computed as:
\begin{equation*}
    \RI_p = \frac{\mathbb{D}[p](p(M_y)-p(M_x))}{p(M_x)},
\end{equation*}
where 
$\mathbb{D}[p]$ is an indicator function denoting whether higher scores of $p$ is desirable; and
$p(M_x)$ and $p(M_y)$ denote the score of property $p$ in 
the input molecule $M_x$ and generated molecule $M_y$, respectively.
}
%
Higher {\SR}, {\Sim}, and {\RI} are desirable,
indicating more successful optimizations.
%

\begin{table*}[h!]
\centering
\setlength{\tabcolsep}{0pt}%
\caption{Overall Performance in IND Tasks}
\label{tbl:main_ind}
\vspace{-5pt}
\begin{small}
\begin{threeparttable}
\begin{tabular}{
   @{\hspace{0pt}}l@{\hspace{5pt}}
   @{\hspace{2pt}}r@{\hspace{2pt}}
   @{\hspace{2pt}}r@{\hspace{2pt}}
   @{\hspace{2pt}}r@{\hspace{2pt}}
   @{\hspace{5pt}}c@{\hspace{5pt}} 
   @{\hspace{0pt}}r@{\hspace{2pt}}
   @{\hspace{2pt}}r@{\hspace{2pt}}
   @{\hspace{2pt}}r@{\hspace{2pt}}
   @{\hspace{5pt}}c@{\hspace{5pt}} 
   @{\hspace{0pt}}r@{\hspace{2pt}}
   @{\hspace{2pt}}r@{\hspace{2pt}}
   @{\hspace{2pt}}r@{\hspace{2pt}}
   @{\hspace{5pt}}c@{\hspace{5pt}} 
   @{\hspace{0pt}}r@{\hspace{2pt}}
   @{\hspace{2pt}}r@{\hspace{2pt}}
   @{\hspace{2pt}}r@{\hspace{2pt}}
   @{\hspace{5pt}}c@{\hspace{5pt}} 
   @{\hspace{0pt}}r@{\hspace{2pt}}
   @{\hspace{2pt}}r@{\hspace{2pt}}
   @{\hspace{2pt}}r@{\hspace{0pt}}
}
\toprule
\multirow{2}{*}{Model} & \multicolumn{3}{c}{\BDP} && \multicolumn{3}{c}{\BDQ} && \multicolumn{3}{c}{\BPQ} && \multicolumn{3}{c}{\DPQ} && \multicolumn{3}{c}{\BDPQ} \\
\cmidrule(){2-4} \cmidrule(){6-8} \cmidrule(){10-12} \cmidrule(){14-16} \cmidrule(){18-20}
& \SR$^{\uparrow}$ & \Sim$^{\uparrow}$ & \RI$^{\uparrow}$ & 
& \SR$^{\uparrow}$ & \Sim$^{\uparrow}$ & \RI$^{\uparrow}$ &
& \SR$^{\uparrow}$ & \Sim$^{\uparrow}$ & \RI$^{\uparrow}$ &
& \SR$^{\uparrow}$ & \Sim$^{\uparrow}$ & \RI$^{\uparrow}$ &
& \SR$^{\uparrow}$ & \Sim$^{\uparrow}$ & \RI$^{\uparrow}$ \\
\midrule

\rowcolor{lightgray}
\multicolumn{20}{c}{\textbf{General-purpose LLMs}} 
\\

Mistral (0-shot) & 6.60 & \textbf{\underline{0.81}} & 0.68 &  & 3.00 & \textbf{\underline{0.76}} & 0.53 &  & 15.80 & {0.73} & 0.51 &  & 2.20 & \textbf{\underline{0.65}} & 0.41 &  & 3.20 & 0.77 & 0.87 \\
Llama (0-shot) & 22.00 & 0.73 & 0.74 &  & 2.20 & 0.64 & 0.53 &  & 28.40 & 0.64 & 0.72 &  & 2.60 & 0.62 & 0.32 &  & 5.20 & \textbf{\underline{0.80}} & 0.62 \\
Claude-3.5 (0-shot) & 19.60 & 0.66 & 1.05 &  & 13.00 & 0.62 & 1.14 &  & 56.00 & 0.62 & 0.86 &  & 11.00 & 0.54 & 0.51 &  & 8.00 & 0.60 & 1.34 \\
GPT-4o (0-shot) & 7.80 & 0.69 & 0.90 &  & 2.00 & 0.69 & 0.62 &  & 36.40 & 0.73 & 0.42 &  & 2.80 & 0.57 & 0.50 &  & 1.80 & 0.71 & 0.39 \\
Mistral (5-shot) & 35.20 & 0.64 & 2.10 &  & 17.00 & 0.60 & 2.32 &  & 68.60 & 0.63 & 0.79 &  & 10.40 & 0.54 & 1.10 &  & 11.00 & 0.69 & 0.96 \\
Llama (5-shot) & 35.40 & 0.57 & 2.71 &  & 16.60 & 0.43 & \textbf{\underline{5.70}} &  & 34.60 & 0.70 & 0.64 &  & 8.20 & 0.44 & 3.02 &  & 9.60 & 0.54 & 3.45 \\
\cellcolor{yellow!20}Claude-3.5 (5-shot) & 35.40 & 0.50 & 2.43 &  & 29.40 & 0.43 & 3.80 &  & 76.80 & 0.53 & 1.24 &  & \underline{\cellcolor{yellow!20}29.20} & \cellcolor{yellow!20}0.37 & \cellcolor{yellow!20}2.87 &  & \underline{\cellcolor{yellow!20}20.80} & \cellcolor{yellow!20}0.35 & \cellcolor{yellow!20}3.53 \\
GPT-4o (1-shot) & 9.40 & 0.69 & 0.79 &  & 7.60 & 0.66 & 0.61 &  & 40.00 & \textbf{\underline{0.75}} & 0.41 &  & 7.00 & 0.62 & 0.44 &  & 3.40 & 0.70 & 0.61 
\\

\rowcolor{lightgray}
\multicolumn{20}{c}{\textbf{Foundational LLMs for Chemistry}}
\\

\ChemLLM & 0.20 & 0.17 & 1.20 &  & 1.00 & 0.55 & 0.82 &  & 4.80 & 0.29 & 0.96 &  & 0.60 & 0.28 & 0.42 &  & 0.00 & n/a & n/a
\\
\cellcolor{yellow!20}\LlaSMolM & \underline{\cellcolor{yellow!20}43.60} & \cellcolor{yellow!20}0.62 & \cellcolor{yellow!20}1.09 &  & \underline{\cellcolor{yellow!20}31.40} & \cellcolor{yellow!20}0.66 & \cellcolor{yellow!20}0.93 &  & \underline{\cellcolor{yellow!20}86.00} & \cellcolor{yellow!20}0.58 & \cellcolor{yellow!20}0.84 &  & 24.00 & 0.57 & 0.61 &  & 14.00 & 0.62 & 1.03 \\

\rowcolor{lightgray}
\multicolumn{20}{c}{\textbf{Task-specific non-LLMs}}
\\
\PMol & 12.20 & 0.12 & \textbf{7.46} &  & 23.20 & 0.10 & 5.40 &  & 15.80 & 0.10 & \underline{1.50} &  & 23.60 & 0.10 & \textbf{5.46} &  & 6.60 & 0.11 & \textbf{5.36} \\

\rowcolor{lightgray}
\multicolumn{20}{c}{\textbf{Task-specific LLMs}}
\\

\cellcolor{green!10}\mollmTripleTaskM 
& 84.80 & 0.47 & 4.30 &  
& 87.00 & 0.47 & 5.61 &
& 93.00 & 0.46 & 1.49 &
& \cellcolor{green!10}62.80 & \cellcolor{green!10}0.37 & \cellcolor{green!10}3.87 &
& - & - & -
\\

\cellcolor{green!10}\mollmTripleTaskL
& \textbf{\cellcolor{green!10}86.80} & \cellcolor{green!10}0.48 & \cellcolor{green!10}4.38 & 
& \textbf{\cellcolor{green!10}90.00} & \cellcolor{green!10}0.46 & \cellcolor{green!10}5.66 &
& \cellcolor{green!10}94.00 & \cellcolor{green!10}0.50 & \cellcolor{green!10}1.38 &
& 60.60 & 0.44 & 3.76 &
& - & - & -
\\

\cellcolor{green!10}\mollmQuadTaskM & 71.60 & 0.49 & 3.27 &  & 57.40 & 0.55 & 2.56 &  & 90.20 & 0.46 & 1.41 &  & 54.00 & 0.44 & 3.02 &  & \cellcolor{green!10}30.00 & \cellcolor{green!10}0.48 & \cellcolor{green!10}3.44 \\

\mollmQuadTaskL & 53.60 & 0.63 & 1.94 &  & 48.60 & 0.59 & 1.29 &  & 93.40 & 0.59 & 1.12 &  & 39.60 & 0.57 & 1.32 &  & 28.00 & 0.66 & 1.02 \\

\hline

\ImpT (\%) 
& 99.1 & -22.6 & 301.8 &  & 186.6 & -30.3 & 508.6 &  & 9.3 & -13.8 & 64.3 &  & 115.1 & 0.0 & 34.8 &  & 44.2 & 37.1 & -2.5
\\

\rowcolor{lightgray}
\multicolumn{20}{c}{\textbf{Generalist LLMs}}
\\

\mollmTripleGenM 
& 75.60 & 0.56 & 3.31 &
& 79.40 & 0.53 & 4.52 &
& 93.20 & 0.55 & 1.23 &
& 57.20 & 0.50 & 2.22 & 
& - & - & -
\\

\mollmTripleGenL
& 77.40 & 0.51 & 3.16 &
& 76.40 & 0.57 & 4.41 &
& 95.40 & 0.50 & 1.46 & 
& 63.40 & 0.49 & 2.46 &
& - & - & -
\\

\cellcolor{blue!10}\mollmQuadGenM & 81.40 & 0.55 & 3.95 &  & 82.60 & 0.56 & 5.24 &  & 96.20 & 0.52 & 1.52 &  & \textbf{\cellcolor{blue!10}66.60} & \cellcolor{blue!10}0.53 & \cellcolor{blue!10}2.41 &  & \textbf{\cellcolor{blue!10}57.40} & \cellcolor{blue!10}0.52 & \cellcolor{blue!10}3.04 \\

\mollmQuadGenL & 80.40 & 0.54 & 3.60 &  & 81.40 & 0.56 & 4.81 &  & 93.80 & 0.47 & 1.64 &  & 61.40 & 0.50 & 2.02 &  & 49.80 & 0.48 & 3.26 \\

\cellcolor{blue!10}\mollmSixGenM & \cellcolor{blue!10}83.00 & \cellcolor{blue!10}0.57 & \cellcolor{blue!10}3.60 &  & \cellcolor{blue!10}85.80 & \cellcolor{blue!10}0.59 & \cellcolor{blue!10}4.78 &  & \textbf{\cellcolor{blue!10}96.80} & \cellcolor{blue!10}0.53 & \cellcolor{blue!10}1.48 &  & 60.80 & 0.54 & 2.16 &  & 54.00 & 0.54 & 3.09 \\

\mollmSixGenL & 77.00 & 0.53 & 3.73 &  & 79.60 & 0.56 & 5.05 &  & 95.00 & 0.47 & \textbf{1.66} &  & 57.00 & 0.49 & 2.50 &  & 52.20 & 0.49 & 3.48 \\

\hline
\ImpG (\%)
& 90.4 & -8.1 & 230.3 &  & 173.2 & -10.6 & 414.0 &  & 12.6 & -8.6 & 76.2 &  & 128.1 & 43.2 & -16.0 &  & 176.0 & 48.6 & -13.9
\\

\bottomrule
\end{tabular}

\begin{tablenotes}[normal,flushleft]
\footnotesize
\item $^\uparrow$ and $^\downarrow$ denote whether a higher or lower value of the metric is desirable, respectively.
For each task,
the best baseline performance is \underline{underlined} 
and the best overall performance is in \textbf{bold}
for each metric.
{\ImpT} and {\ImpG} denote the percentage improvement from the \colorbox{green!10}{best task-specific LLM} 
and \colorbox{blue!10}{best generalist LLM} over the 
\colorbox{yellow!20}{best baseline},
respectively,
where the best models are selected based on {\SR} for each task.
`-' indicates cases where models are trained on 3 properties but the task has additional properties 
not included in the model training.
`n/a' indicates undefined metrics due to no optimized cases (i.e., 0\% \SR).
\end{tablenotes}

\end{threeparttable}
\end{small}
\vspace{-15pt}
\end{table*}

\section{Experimental Results}
\paragraph{Main Findings:} 
Our experiments reveal the following findings:
\textbf{(1)} Both task-specific and generalist 
{\mollm}s consistently outperform 
general-purpose LLMs, foundational LLMs for chemistry, 
and task-specific non-LLMs across all IND (Section~\ref{sec:results:ind}) and OOD tasks (Section~\ref{sec:results:ood}), 
significantly improving {\SR} by as much as 186.6\% over the best baseline.
%
%
%
\textbf{(2)} Compared to task-specific {\mollm}s, 
generalist {\mollm}s excel on 3 out of 5 IND tasks and
demonstrate competitive performance on the other 2 tasks,
with remarkable gains of 91.3\% in
{\SR} on more complex tasks such as \BDPQ 
(Section~\ref{sec:results:ind}).
{\textbf{(3)} 
Generalist \mollmSixGen models
exhibit strong 0-shot generalization to unseen (OOD) tasks and unseen instructions, 
significantly outperforming powerful general-purpose LLMs (Section~\ref{sec:results:ood}).}
\textbf{(4)} All {\mollm}s substantially outperform
the best general-purpose LLM, Claude-3.5 (5-shot) and foundational
LLM for chemistry, \LlaSMolM,
across all IND and OOD tasks.
\subsection{IND Evaluation}
\label{sec:results:ind}

Table~\ref{tbl:main_ind} shows the overall performance
of {\mollm}s and baselines across all 5 IND tasks.
Detailed results for each task are in Appendix~\ref{sec:app:results:ind}.

\paragraph{Overall Comparison:} 
Both task-specific and generalist {\mollm}s
significantly outperform all baselines across all IND tasks.
Specifically, the generalist {\mollm}s,
\mollmQuadGenM and \mollmSixGenM,
achieve an average {\SR} of 76.8\% and 76.1\%, respectively, across all 5 tasks --
outperforming the best baseline by 113.2\% and 108.8\% on average.
This is due to the ability of generalist {\mollm}s 
to leverage knowledge synergistically by optimizing different property combinations, 
thereby effectively capturing shared chemical principles and property trade-offs.
%
%
On the most challenging task, \BDPQ with 4 properties,
the generalist {\mollm} outperforms all baselines, task-specific {\mollm}s
and non-LLMs
by as much as 176\% in terms of {\SR}, 
showcasing its strong ability to tackle complex tasks 
with limited training data.
Detailed comparison between {\mollm}s and task-specific non-LLMs
are provided in Appendix~\ref{sec:app:results:ind}.

%
%
%

%

\paragraph{Comparison between task-specific and generalist {\mollm}s:} 
As shown in Table~\ref{tbl:main_ind},
generalist {\mollm}s 
outperform task-specific {\mollm}s on 3 out of 5 IND tasks,
particularly with remarkable gains on more complex task \BDPQ.
On \BDPQ 
the generalist {\mollmQuadGenM} achieves a substantial
improvement of 91.3\% in {\SR} over the best task-specific 
{\mollmQuadTaskM}.
This is likely due to {\BDPQ}'s complexity in balancing four properties.
By leveraging data from other tasks, the generalist {\mollm}s
can better capture the nuanced property trade-offs
that task-specific {\mollm}s 
can not with limited task-specific training data.

%
On the other hand, task-specific {\mollm}s outperform
generalist ones on 2 IND tasks
(\BDP and \BDQ),
where
the properties BBBP and DRD2 are positively correlated
(Pearson correlation of 0.6),
and there is sufficient training data (2,064 pairs in \BDP and 4,472 in \BDQ).
Notably, DRD2 targets the inhibition of a specific receptor, representing a more specialized therapeutic objective than fundamental molecular properties like pLogP or QED.
Therefore, by focusing exclusively on such highly correlated
properties and specific therapeutic requirements,
task-specific {\mollm}s learn
more targeted structural modifications for each task.
Nonetheless, this only leads to modest improvements of 4.6\% and 4.9\%
over the generalist {\mollmSixGenM}.
These results collectively highlight 
the complementary strengths of generalist and task-specific {\mollm}s, 
with the generalist \mollmSixGenM offering a more scalable and foundational model for diverse optimization tasks.

\paragraph{Comparison between {\mollm}s and general-purpose LLMs:} 

All {\mollm}s substantially
outperform general-purpose LLMs by a large margin.
For example, across 5 IND tasks,
{\mollmQuadGenM} and {\mollmSixGenM}
achieve a significant average improvement of 
128.1\% and 124.0\% in {{\SR}}
over the best general-purpose LLM baseline, Claude-3.5 (5-shot).
This remarkable performance can be attributed to
the instruction tuning of \mollm on molecule pairs, 
enabling it to learn
modification strategies that general-purpose LLMs fail to acquire through in-context learning.
General-purpose LLMs,
particularly with 0-shot prompting, 
exhibit extremely low {\SR} and relatively high {\Sim},
{meaning} that the very few optimized
molecules are highly similar to the input.
In contrast, {\mollm}s achieve substantially higher {\SR} and {\RI},
demonstrating more successful optimizations
while maintaining reasonable similarity 
(e.g., {\Sim} in [0.5, 0.6]).

\paragraph{Comparison between {\mollm}s and foundational LLM for chemistry:} 
All {\mollm}s demonstrate significant improvement over
the state-of-the-art foundational LLM for chemistry, \LlaSMolM, on all IND tasks.
For example, \mollmSixGenM outperforms \LlaSMolM significantly by
186.6\% on \BDQ, 
and 99.1\% on \BDP,
with up to a 5-fold relative improvement 
(e.g., {\RI} of 508.6 on \BDQ)
in desired properties.
Note that \LlaSMol was not instruction-tuned on 
molecule optimization tasks.
Thus, the performance gap 
suggests that the pre-trained chemistry knowledge in foundational LLMs, such as \LlaSMol,
is not sufficient to solve tasks as specialized as molecule optimization.
This highlights the importance of instruction tuning on \MOptData 
for acquiring more specialized knowledge.

%
%
\begin{table*}[h!]
\centering
\setlength{\tabcolsep}{0pt}%
\caption{Overall Performance in OOD Tasks}
\label{tbl:main_ood}
\vspace{-5pt}
\begin{small}
\begin{threeparttable}
\begin{tabular}{
   @{\hspace{0pt}}l@{\hspace{5pt}}
   @{\hspace{2pt}}r@{\hspace{2pt}}
   @{\hspace{2pt}}r@{\hspace{2pt}}
   @{\hspace{2pt}}r@{\hspace{2pt}}
   @{\hspace{5pt}}c@{\hspace{5pt}} 
   @{\hspace{0pt}}r@{\hspace{2pt}}
   @{\hspace{2pt}}r@{\hspace{2pt}}
   @{\hspace{2pt}}r@{\hspace{2pt}}
   @{\hspace{5pt}}c@{\hspace{5pt}} 
   @{\hspace{0pt}}r@{\hspace{2pt}}
   @{\hspace{2pt}}r@{\hspace{2pt}}
   @{\hspace{2pt}}r@{\hspace{2pt}}
   @{\hspace{5pt}}c@{\hspace{5pt}} 
   @{\hspace{0pt}}r@{\hspace{2pt}}
   @{\hspace{2pt}}r@{\hspace{2pt}}
   @{\hspace{2pt}}r@{\hspace{2pt}}
   @{\hspace{5pt}}c@{\hspace{5pt}} 
   @{\hspace{0pt}}r@{\hspace{2pt}}
   @{\hspace{2pt}}r@{\hspace{2pt}}
   @{\hspace{2pt}}r@{\hspace{0pt}}
}
\toprule
\multirow{2}{*}{Model} & \multicolumn{3}{c}{\MPQ} && \multicolumn{3}{c}{\BDMQ} && \multicolumn{3}{c}{\BHMQ} && \multicolumn{3}{c}{\BMPQ} && \multicolumn{3}{c}{\HMPQ} \\
\cmidrule(){2-4} \cmidrule(){6-8} \cmidrule(){10-12} \cmidrule(){14-16} \cmidrule(){18-20}
& \SR$^{\uparrow}$ & \Sim$^{\uparrow}$ & \RI$^{\uparrow}$ & 
& \SR$^{\uparrow}$ & \Sim$^{\uparrow}$ & \RI$^{\uparrow}$ &
& \SR$^{\uparrow}$ & \Sim$^{\uparrow}$ & \RI$^{\uparrow}$ &
& \SR$^{\uparrow}$ & \Sim$^{\uparrow}$ & \RI$^{\uparrow}$ &
& \SR$^{\uparrow}$ & \Sim$^{\uparrow}$ & \RI$^{\uparrow}$ \\
\midrule

\rowcolor{lightgray}
\multicolumn{20}{c}{\textbf{General-purpose LLMs}} 
\\
Mistral (0-shot) & 11.20 & {0.57} & 0.48 &  & 1.20 & 0.68 & 0.37 &  & 12.71 & 0.73 & 1.90 &  & 12.57 & 0.61 & 0.54 &  & 21.88 & \textbf{\underline{0.72}} & 0.72 \\
Llama (0-shot) & 25.80 & 0.44 & 0.61 &  & 1.20 & \textbf{\underline{0.76}} & 0.30 &  & 11.02 & {0.74} & 0.68 &  & 16.75 & 0.51 & 0.57 &  & 15.62 & 0.47 & 0.60 \\
Claude-3.5 (0-shot) & 17.40 & 0.49 & 0.52 &  & 15.00 & 0.57 & 0.87 &  & 38.98 & 0.51 & 2.35 &  & 44.50 & 0.55 & 0.85 &  & 38.54 & 0.54 & 1.01 \\
GPT-4o (0-shot) & 19.40 & \textbf{\underline{0.61}} & 0.35 &  & 1.60 & 0.67 & 0.18 &  & 17.80 & 0.66 & 0.65 &  & 25.13 & 0.67 & 0.35 &  & 20.83 & 0.67 & 0.37 \\
Mistral (5-shot) & 59.60 & 0.54 & 0.57 &  & 20.40 & 0.59 & 1.65 &  & 34.75 & 0.70 & 1.31 &  & 49.21 & 0.62 & 0.73 &  & 46.88 & 0.66 & 0.91 \\
Llama (5-shot) & 34.80 & {0.57} & 0.53 &  & 16.80 & 0.39 & \underline{3.22} &  & 36.44 & 0.67 & 1.13 &  & 31.94 & {0.66} & 0.60 &  & 33.33 & 0.68 & 0.61 \\
\cellcolor{yellow!20}Claude-3.5 (5-shot) & 50.60 & 0.49 & \underline{0.71} &  & \underline{\cellcolor{yellow!20}30.40} & \cellcolor{yellow!20}0.49 & \cellcolor{yellow!20}2.32 &  & 52.54 & 0.48 & \underline{2.52} &  & 52.36 & 0.46 & \underline{1.08} &  & \underline{\cellcolor{yellow!20}65.62} & \cellcolor{yellow!20}0.48 & \underline{\cellcolor{yellow!20}1.32} \\

GPT-4o (1-shot) & 21.40 & 0.60 & 0.48 &  & 6.20 & 0.70 & 0.75 &  & 14.41 & \textbf{\underline{0.75}} & 0.67 &  & 24.08 & \textbf{\underline{0.68}} & 0.45 &  & 25.00 & 0.70 & 0.54 \\

\rowcolor{lightgray}
\multicolumn{20}{c}{\textbf{Foundational LLMs for Chemistry}}
\\

\ChemLLM & 6.20 & 0.31 & 0.61 &  & 0.00 & n/a & n/a &  & 1.69 & 0.53 & 0.74 &  & 5.24 & 0.24 & 0.72 &  & 3.12 & 0.58 & 0.51 \\

\cellcolor{yellow!20}\LlaSMolM & \underline{\cellcolor{yellow!20}76.40} & \cellcolor{yellow!20}0.55 & \cellcolor{yellow!20}0.53 &  & 28.20 & 0.66 & 0.52 &  & \underline{\cellcolor{yellow!20}53.39} & \cellcolor{yellow!20}0.62 & \cellcolor{yellow!20}1.14 &  & \underline{\cellcolor{yellow!20}64.92} & \cellcolor{yellow!20}0.58 & \cellcolor{yellow!20}0.57 &  & 53.12 & 0.62 & 0.70 \\

\rowcolor{lightgray}
\multicolumn{20}{c}{\textbf{Generalist LLMs}}
\\

\cellcolor{blue!10}\mollmSixGenM & \textbf{\cellcolor{blue!10}95.20} & \cellcolor{blue!10}0.53 & \cellcolor{blue!10}0.85 &  & \textbf{\cellcolor{blue!10}79.00} & \cellcolor{blue!10}0.56 & \cellcolor{blue!10}3.10 &  & 86.44 & 0.54 & 2.58 &  & 91.10 & 0.53 & 1.06 &  & 91.67 & 0.55 & 1.42 \\

\cellcolor{blue!10}\mollmSixGenL & 93.60 & 0.48 & \textbf{0.91} &  & 74.20 & 0.55 & \textbf{3.25} &  & \textbf{\cellcolor{blue!10}93.22} & \cellcolor{blue!10}0.49 & \textbf{\cellcolor{blue!10}3.57} &  & \textbf{\cellcolor{blue!10}95.29} & \cellcolor{blue!10}0.49 & \textbf{\cellcolor{blue!10}1.20} &  & \textbf{\cellcolor{blue!10}97.92} & \cellcolor{blue!10}0.46 & \textbf{\cellcolor{blue!10}1.76} \\

\hline
\ImpG (\%)
& 24.6 & -3.6 & 60.4 &  & 159.9 & 14.3 & 33.6 &  & 74.6 & -21.0 & 213.2 &  & 46.8 & -15.5 & 110.5 &  & 49.2 & -4.2 & 33.3 \\

\bottomrule
\end{tabular}

\begin{tablenotes}[normal,flushleft]
\footnotesize
\item The metrics, notations and formatting have the same meanings as those
in Table~\ref{tbl:main_ind}.
\par
\end{tablenotes}

\end{threeparttable}
\end{small}
\end{table*}

\paragraph{Comparison between \mollm and \PMol:} 

{\mollm}s consistently outperform \PMol across all IND tasks,
improving {\SR} by as much as 770.0\% on \BDPQ task.
This gain likely arises due to instruction-tuning which enables {\mollm}s 
to effectively learn modification strategies
while leveraging their pre-trained general-purpose and chemical knowledge.
In contrast, \PMol learns both
chemical knowledge and task-specific knowledge from scratch (Appendix~\ref{sec:app:pmol:discuss}),
making it heavily reliant on limited task-specific training data
and thus resulting in extremely low {\SR}.
Notably, the very few optimized cases from \PMol exhibit high {\RI} but low {\Sim}, indicating substantial property improvements with drastic structural changes. 
This suggests that \PMol tends to generate entirely new molecules, thus failing to retain the core scaffold -- a key requirement in lead optimization.

\paragraph{Comparison between \mollmTripleTask and \mollmQuadTask:} 
\mollmTripleTask is consistently better than \mollmQuadTask in terms of {{\SR}} across all 4 IND tasks with 3 properties. 
This performance gap can be attributed to \mollmQuadTask's more constrained training setup, 
with fewer training pairs (e.g., 624 in \BDPQ) and the added complexity of an additional property constraint. 
As a result, when evaluated on tasks with 3 properties,
\mollmQuadTask must adapt its knowledge learned
from improving 4 properties, which can limit its effectiveness.
In contrast, 
\mollmTripleTask benefits from more focused task-specific training with larger datasets (e.g., 4,472 in \BDQ),
enabling better performance.
%

\subsection{OOD Evaluation}
\label{sec:results:ood}

Table~\ref{tbl:main_ood} presents the overall performance of {\mollm}s
and baselines
across all 5 OOD tasks
(with detailed results in Appendix~\ref{sec:app:results:ood}).
Note that OOD tasks involve novel property combinations excluded from training, making task-specific models and comparisons with
\mollmTripleGen and \mollmQuadGen
inapplicable.
Generalist {\mollm}s demonstrate robust 0-shot generalization to OOD tasks, 
significantly outperforming all baselines. 
For instance, both \mollmSixGenM and \mollmSixGenL
achieve very high {\SR} of 88.7\% and 90.8\%, respectively,
on average across all 5 tasks --
outperforming strong baselines such as Claude-3.5 (5-shot) and \LlaSMolM
by as much as 159.9\% on task \BDMQ.
%
%
%
%
By learning optimization strategies and property trade-offs
across diverse tasks during training, 
generalist {\mollm}s develop a flexible understanding of modification strategies that can generalize to novel optimization tasks.
This generalizability is crucial in practice, where 
the dynamic nature of therapeutic requirements requires one unified foundational model capable of handling novel and diverse optimization tasks
without task-specific retraining.

\subsection{Generalizability to Unseen Instructions}
\label{sec:results:uninst}

\begin{table*}[h!]
\centering
\setlength{\tabcolsep}{0pt}%
\caption{Overall Performance with Unseen Instructions in IND Tasks}
\label{tbl:main_uninst}
\vspace{-5pt}
\begin{small}
\begin{threeparttable}
\begin{tabular}{
   @{\hspace{2pt}}l@{\hspace{2pt}}
   @{\hspace{2pt}}l@{\hspace{10pt}}
   @{\hspace{2pt}}r@{\hspace{2pt}}
   @{\hspace{2pt}}r@{\hspace{2pt}}
   @{\hspace{2pt}}r@{\hspace{2pt}}
   @{\hspace{5pt}}c@{\hspace{5pt}}
   @{\hspace{0pt}}r@{\hspace{2pt}}
   @{\hspace{2pt}}r@{\hspace{2pt}}
   @{\hspace{2pt}}r@{\hspace{2pt}}
   @{\hspace{5pt}}c@{\hspace{5pt}}
   @{\hspace{2pt}}r@{\hspace{2pt}}
   @{\hspace{2pt}}r@{\hspace{2pt}}
   @{\hspace{2pt}}r@{\hspace{2pt}}
   @{\hspace{5pt}}c@{\hspace{5pt}}
   @{\hspace{0pt}}r@{\hspace{2pt}}
   @{\hspace{2pt}}r@{\hspace{2pt}}
   @{\hspace{2pt}}r@{\hspace{2pt}}
   @{\hspace{5pt}}c@{\hspace{5pt}}
   @{\hspace{0pt}}r@{\hspace{2pt}}
   @{\hspace{2pt}}r@{\hspace{2pt}}
   @{\hspace{2pt}}r@{\hspace{0pt}}
}
\toprule
\multirow{2}{*}{Model} & 
\multirow{2}{*}{Instr} & 
\multicolumn{3}{c}{\BDP} && \multicolumn{3}{c}{\BDQ} && \multicolumn{3}{c}{\BPQ} && \multicolumn{3}{c}{\DPQ} && \multicolumn{3}{c}{\BDPQ} \\
\cmidrule(){3-5} \cmidrule(){7-9} \cmidrule(){11-13} \cmidrule(){15-17} \cmidrule(){19-21}
\mollm &  & \SR$^{\uparrow}$ & \Sim$^{\uparrow}$ & \RI$^{\uparrow}$ &
& \SR$^{\uparrow}$ & \Sim$^{\uparrow}$ & \RI$^{\uparrow}$ &
& \SR$^{\uparrow}$ & \Sim$^{\uparrow}$ & \RI$^{\uparrow}$ &
& \SR$^{\uparrow}$ & \Sim$^{\uparrow}$ & \RI$^{\uparrow}$ &
& \SR$^{\uparrow}$ & \Sim$^{\uparrow}$ & \RI$^{\uparrow}$ \\
\midrule

\rowcolor{lightgray}
\multicolumn{21}{c}{\textbf{Task-specific LLMs}}
\\
\multirow{2}{*}{Mistral} 
& seen & 84.80 & 0.47 & 4.30 &
& 87.00 & \textbf{0.47} & 5.61 &
& 93.00 & 0.46 & 1.49 &
& 62.80 & \textbf{0.37} & 3.87 &
& 30.00 & \textbf{0.48} & 3.44 
\\
& unseen & \textbf{89.60} & 0.45 & \textbf{5.11} &
& 87.40 & 0.44 & \textbf{6.29} &
& 93.00 & 0.45 & 1.48 &
& 64.20 & 0.35 & 3.95 &
& \textbf{32.80} & 0.45 & \textbf{3.62}
\\
\hline

\multirow{2}{*}{Llama} 
& seen & 86.80 & \textbf{0.48} & 4.38 &
& 90.00 & 0.46 & 5.66 &
& 94.00 & 0.50 & 1.38 & 
& 60.60 & \textbf{0.44} & 3.76 &
& \textbf{28.00} & 0.66 & 1.02
\\

& unseen & 85.40 & 0.44 & \textbf{4.69} &
& 90.40 & 0.46 & 5.68 &
& 93.80 & 0.49 & 1.42 &
& 63.60 & 0.39 & \textbf{4.36} &
& 24.20 & 0.64 & \textbf{1.29}
\\

\rowcolor{lightgray}
\multicolumn{21}{c}{\textbf{Generalist LLMs}}
\\

\multirow{2}{*}{{\mbox{$\mathop{\mathtt{\text{-}P(6)_{Mistral}}}\limits$}\xspace}} 
& seen & \textbf{83.00} & 0.57 & \textbf{3.60} &  & \textbf{85.80} & 0.59 & \textbf{4.78} &  & 96.80 & 0.53 & 1.48 &  & \textbf{60.80} & 0.54 & \textbf{2.16} &  & \textbf{54.00} & 0.54 & \textbf{3.09} \\
 & unseen & 75.80 & 0.59 & 3.15 &  & 80.40 & 0.59 & 4.54 &  & 96.20 & 0.54 & 1.42 &  & 54.60 & 0.55 & 1.99 &  & 49.80 & \textbf{0.57} & 2.81 \\
 
\hline
\multirow{2}{*}{{\mbox{$\mathop{\mathtt{\text{-}P(6)_{Llama}}}\limits$}\xspace}} 
& seen & \textbf{77.00} & 0.53 & \textbf{3.73} &  & \textbf{79.60} & 0.56 & \textbf{5.05} &  & 95.00 & 0.47 & 1.66 &  & \textbf{57.00} & 0.49 & \textbf{2.50} &  & \textbf{52.20} & 0.49 & 3.48 \\
 & unseen & 64.60 & 0.53 & 3.06 &  & 73.40 & 0.57 & 4.56 &  & 95.60 & 0.47 & 1.66 &  & 53.60 & 0.50 & 2.15 &  & 46.40 & 0.48 & 3.52
\\
\bottomrule
\end{tabular}

\begin{tablenotes}[normal,flushleft]
\footnotesize
\item 
``seen" and ``unseen" indicate whether the {\mollm}s 
are evaluated with seen and unseen instructions, respectively.
$^\uparrow$ and $^\downarrow$ denote whether a higher or lower value of the metric is desirable, respectively.
The best-performing {\mollm} in each row block is in \textbf{bold} 
if the performance difference between the models evaluated with seen and unseen instructions exceeds 5\%. 
\par
\end{tablenotes}

\end{threeparttable}
\end{small}
\vspace{-10pt}
\end{table*}

Table~\ref{tbl:main_uninst} presents the performance of task-specific {\mollm}s
and generalist model, {\mollmSixGen}, when prompted
with a hold-out instruction and unseen property names (Appendix~\ref{sec:app:instr}).
Overall, task-specific {\mollm}s retain their performance across all tasks,
while, generalist {\mollm}s exhibit a slight
drop of 7\% in {\SR} on average.
This minor drop is expected, 
since generalist {\mollm}s trained on more property combinations, 
encounter the same property names more frequently during instruction-tuning.
This may lead to subtle overfitting to specific names.
Importantly, even with this minor performance drop, 
generalist {\mollm}s still outperform all baselines by a large margin, 
(Section~\ref{sec:results:ind}), 
highlighting their overall superiority.
%
Detailed results are provided in Appendix~\ref{sec:app:results:uninst}.

\subsection{Case Studies}
\label{sec:results:case}

\begin{figure}[htbp]
    \centering
    
    \begin{subfigure}[t]{\columnwidth}
        \centering
        \begin{minipage}{0.43\linewidth}
            \centering
            \tiny $M_x$
            \includegraphics[width=1.0\linewidth]{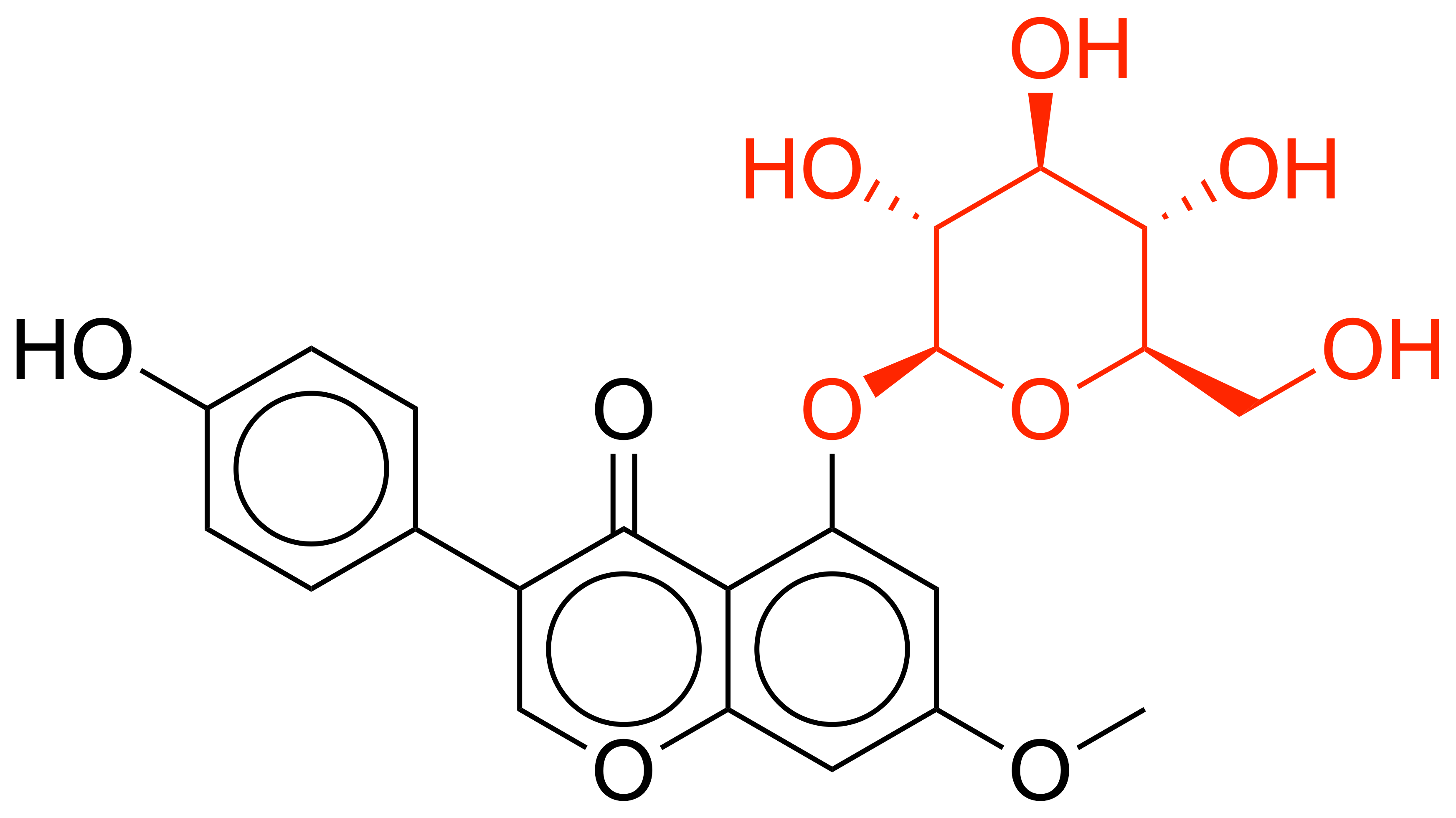}
            \par\vspace{2pt}
            \tiny BBBP=0.08, HIA=0.42, \\
            \tiny Mutag=0.60, QED=0.37
        \end{minipage}
        \hfill
        \begin{minipage}{0.1\linewidth}
            \centering \Large$\Rightarrow$\\
            \raggedright \tiny \mollm
            \raggedright \tiny \Sim=0.61
        \end{minipage}
        \hfill
        \begin{minipage}{0.43\linewidth}
            \centering
            \tiny $M_y$
            \includegraphics[width=1.0\linewidth]{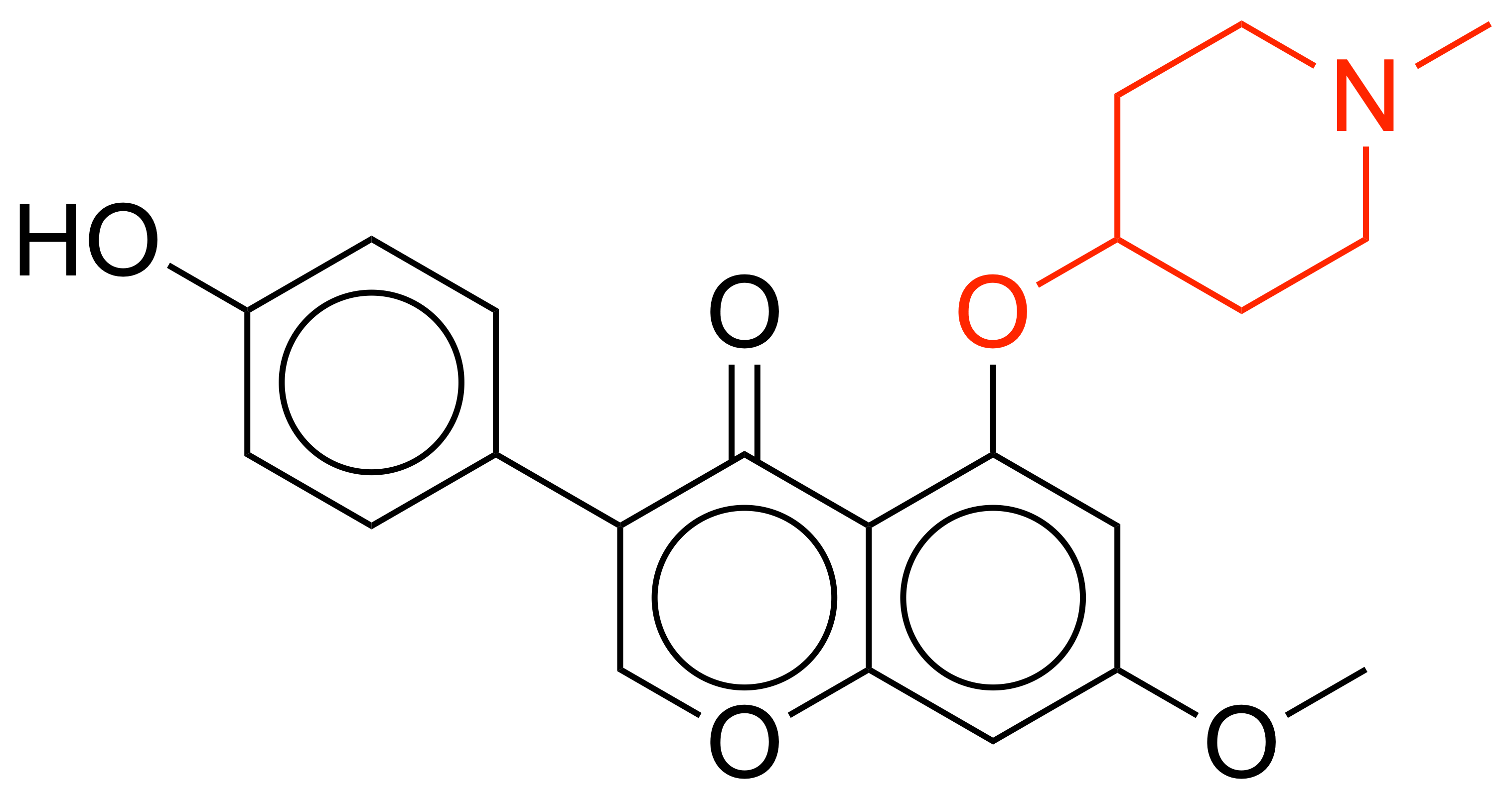}
            \par\vspace{2pt}
            \tiny BBBP=0.82 (+0.74), HIA=0.98 (+0.56),\\ 
            \tiny Mutag=0.30 (-0.30), QED=0.74 (+0.37)
        \end{minipage}
        \caption{{\mollmSixGenM}}
        \label{fig:BHMQ_mollm_case}
    \end{subfigure}
    
    \vspace{1em}
    
    \begin{subfigure}[t]{\columnwidth}
        \centering
        \begin{minipage}{0.43\linewidth}
            \centering
            \tiny $M_x$
            \includegraphics[width=1.0\linewidth]{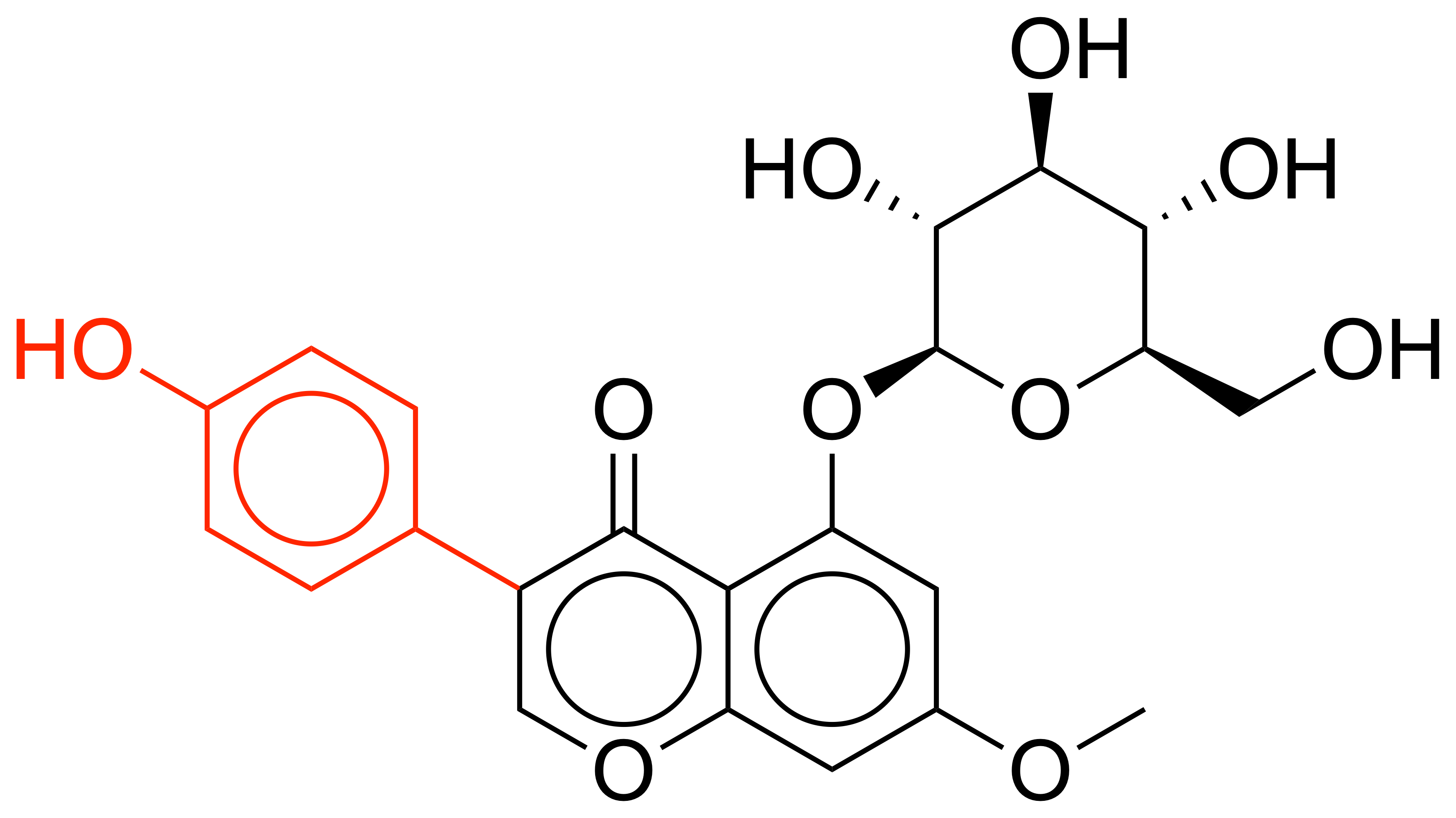}
            \par\vspace{2pt}
            \tiny BBBP=0.08, HIA=0.42, \\
            \tiny Mutag=0.60, QED=0.37
        \end{minipage}
        \hfill
        \begin{minipage}{0.1\linewidth}
            \centering \Large$\Rightarrow$\\
            \raggedright \tiny \LlaSMol
            \raggedright \tiny \Sim=0.66
        \end{minipage}
        \hfill
        \begin{minipage}{0.43\linewidth}
            \centering
            \tiny $M_y$\\
            \includegraphics[width=0.65\linewidth]{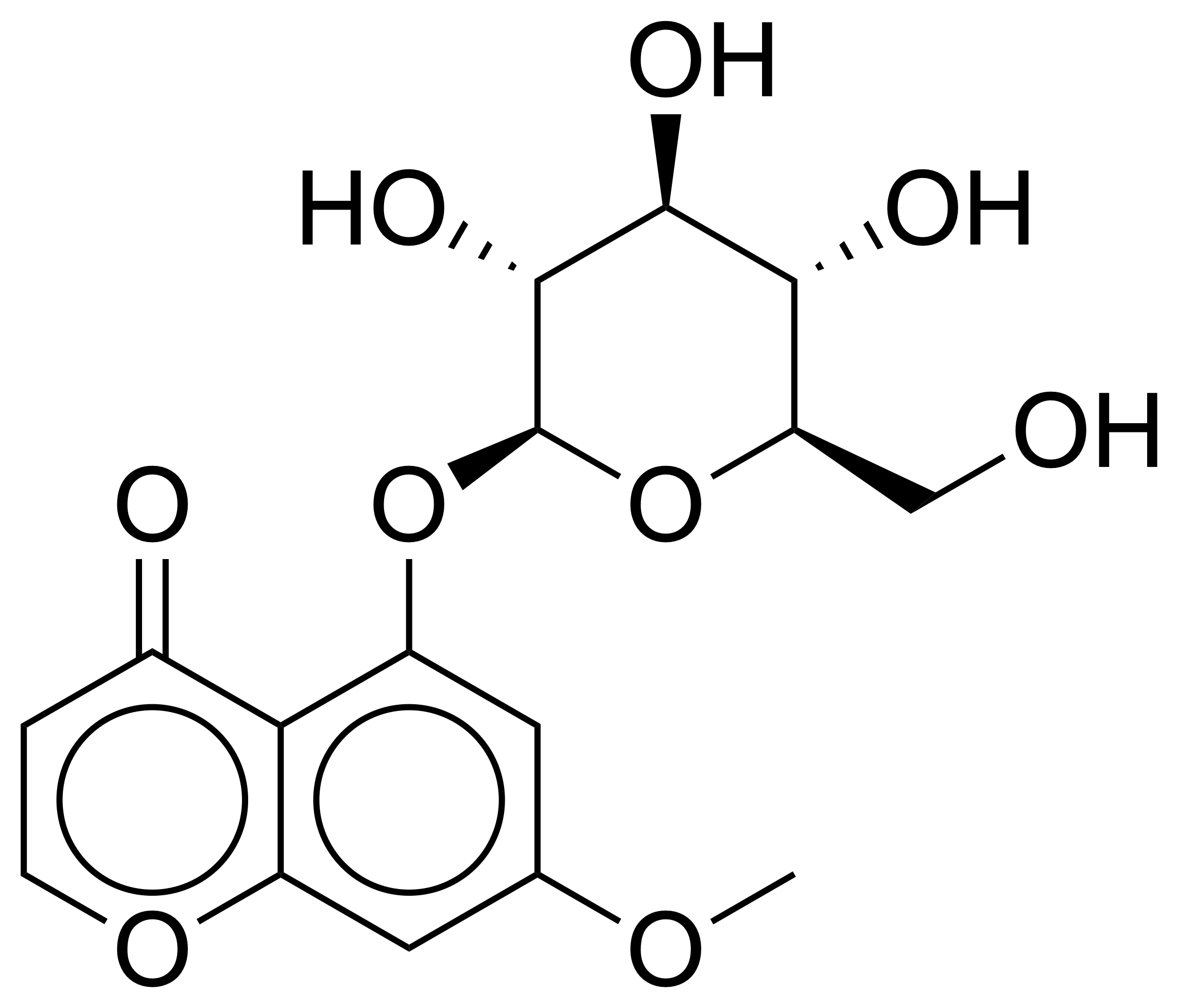}
            \par\vspace{1pt}
            \tiny BBBP=0.45 (+0.37), HIA=0.63 (+0.21),\\ 
            \tiny Mutag=0.47 (-0.13), QED=0.54 (+0.17)
        \end{minipage}
        \caption{{\LlaSMolM}}
        \label{fig:BHMQ_llasmol_case}
    \end{subfigure}
    
    \caption{An optimization case on \BHMQ. Modifications are highlighted in red.}
    \label{fig:BHMQ_cases}
\end{figure}

%
Figure~\ref{fig:BHMQ_mollm_case} shows a
successful optimization for the OOD task \BHMQ,
where \mollmSixGenM improves all desired properties by 
replacing the sugar moiety in $M_x$ 
with a nitrogen-containing heterocycle in $M_y$ (highlighted fragments).
%
The sugar moiety, with multiple hydroxyl (-OH) groups,
increases polarity and hydrogen bonding, 
limiting passive permeability and leading to low BBBP and HIA~\cite{goetz2017relationship, mullard2018re}.
Replacing this fragment with a nitrogen heterocycle
reduces polarity and hydrogen bonding,
leading to improved BBBP (+0.74) and HIA (+0.56).
Moreover, hydroxyl-rich sugars in $M_x$ are 
prone to oxidation and glycation,
compromising stability and bioavailability~\cite{twarda2022advanced, chen2024advanced}. 
In contrast, the nitrogen heterocycle in $M_y$ 
is a well-known motif for improving metabolic stability and bioavailability~\cite{kerru2020review, ebenezer2022overview},
leading to significant improvements in mutagenicity (-0.30) and QED (+0.37).
In contrast, \LlaSMolM retains the sugar moiety and
instead removes a phenol group (Figure~\ref{fig:BHMQ_llasmol_case}), leaving polarity and hydrogen bonding largely unaffected, 
resulting in limited improvements.
Additional cases are in Appendix~\ref{sec:app:results:additional_cases}.

\section{Conclusion}
\label{sec:conclusion}

In this work, we introduced \MOptData, the first high-quality 
instruction-tuning dataset specifically focused on challenging
multi-property optimization tasks.
Leveraging \MOptData, 
{\mollm}s achieve state-of-the-art performance across all IND and OOD tasks,
notably outperforming strong general-purpose LLMs and foundational LLMs for chemistry.
Generalist {\mollm}s demonstrated remarkable generalization to unseen tasks with an average {\SR} of 90.9\%, making them
promising candidates for foundational models in molecule optimization.
This highlights the potential of {\mollm}s
to adapt to diverse optimization tasks mirroring dynamic
therapeutic requirements.

\clearpage

\section{Limitations}

Despite the strong performance of {\mollm}s as demonstrated in our work,
we acknowledge several limitations.
\textbf{(1)}
We did not explore scenarios 
where users specify precise property-specific 
improvement thresholds during inference, 
which could enhance the applicability of {\mollm}s 
for highly customized therapeutic needs.
\textbf{(2)}
Our evaluations are limited to single-step optimization.
We did not explore iterative refinement of generated molecules that could yield even better lead molecules over multiple steps.
\textbf{(3)}
{Our construction of \MOptData and evaluation
leverages well-established empirical property predictors} that
are not experimentally validated,
which may impact the accuracy of optimization outcomes.
\textbf{(4)}
\MOptData encompasses 6 molecular properties 
that play a critical role in successful drug design. 
However, real-world lead optimization often involves additional,
more specialized properties and complex trade-offs
depending on specific therapeutic requirements.
Addressing these limitations in future work could enhance
{\mollm}s' applicability in practice.
{
\textbf{(5)}
While we evaluate generated molecules on synthetic accessibility via SAS, 
we do not validate the synthetic routes or retrosynthetic feasibility of generated molecules using wet-lab experiments.
We acknowledge that such validation would provide stronger evidence, but such efforts are beyond the scope of this work. 
}

\section{Impact Statement}

Our work introduces the first large-scale, high-quality
instruction-tuning dataset, \MOptData, specifically focused on
molecule optimization tasks improving at least 3 properties simultaneously.
By leveraging \MOptData, 
we developed a series of instruction-tuned LLMs ({\mollm}s).
These models
significantly outperform strong closed-source LLMs 
such as Claude-3.5 as well as foundational LLMs for chemistry
on complex multi-property optimization tasks.
To the best of our knowledge, our work is the first to 
introduce a generalist model training framework and a foundational model for molecule optimization.
Notably, the robust zero-shot performance of our generalist {\mollm}s demonstrates their potential
as foundational models for molecule optimization, 
offering scalability and adaptability to diverse optimization scenarios. 
%

\paragraph{Broader Impacts:}
The introduction of foundational models 
capable of handling diverse optimization tasks holds
tremendous potential to accelerate drug discovery pipelines.
These models offer unparalleled flexibility and scalability, enabling practitioners
to adapt them to a wide range of therapeutic requirements
without requiring resource-intensive training.
By relying solely on an efficient inference process,
such models democratize access to advanced optimization capabilities to a broader range of practitioners. 
This advancement could streamline the identification of 
novel drug candidates, 
significantly reducing the cost and time required to develop a new drug. 
%

\section{Ethics Statement}
\label{sec:ethics}

While \MOptData has been carefully curated
to include drug-like, commercially accessible molecules, 
we cannot guarantee that the dataset is entirely free from inaccuracies
or harmful content.
We also cannot eliminate
the potential of our tuned {\mollm}s to generate undesirable or harmful content (e.g., lethal drugs).
We should emphasize that our models are specifically tuned to 
improve widely used molecule properties aligned with 
general drug discovery goals,
and are not intended for generating toxic or lethal molecules. 
%

%
The only property in \MOptData that is related to toxicity
is mutagenicity, which measures the risk of DNA mutations. 
Importantly, our models are tuned explicitly to reduce mutagenicity,
and not to increase it.
%
Furthermore, \mollm models are tuned exclusively on drug-like molecules and optimization objectives aimed at reducing mutagenicity.
As a result, they are unlikely to generate molecules with increased
toxicity or molecules that can be lethal under a normal dosage.

However, if such molecules can be generated with adversarial prompts, 
this could potentially arise from the pretrained knowledge of the base models, 
which includes broader chemical information outside the scope of \MOptData and our instruction-tuning.
To mitigate such risks, safeguards such as usage monitoring, and integration with toxicity prediction pipelines should be implemented when deploying these models. 
Users of our dataset and models are expected to uphold the 
highest ethical standards and
incorporate robust validation pipelines
to prevent misuse. 

\bibliography{paper}

\clearpage

\appendix

\renewcommand{\thefigure}{A\arabic{figure}} 
\setcounter{figure}{0} 

\renewcommand{\thetable}{A\arabic{table}} 
\setcounter{table}{0} 

\section{Details on \MOptData}
\label{sec:app:task}



\begin{table*}[h!]
\centering
\caption{Overview of Properties in \MOptData Tasks for Evaluation}
\label{tbl:task_prop}
\begin{scriptsize}
\begin{threeparttable}
\begin{tabular}{
    @{\hspace{3pt}}l@{\hspace{8pt}}
    @{\hspace{3pt}}r@{\hspace{3pt}}
    @{\hspace{3pt}}r@{\hspace{3pt}}
    @{\hspace{3pt}}r@{\hspace{3pt}}
    @{\hspace{3pt}}r@{\hspace{3pt}}
    @{\hspace{3pt}}r@{\hspace{3pt}}
    @{\hspace{3pt}}r@{\hspace{3pt}}
    @{\hspace{3pt}}r@{\hspace{3pt}}
    @{\hspace{8pt}}r@{\hspace{8pt}}
    @{\hspace{3pt}}r@{\hspace{3pt}}
    @{\hspace{3pt}}r@{\hspace{3pt}}
    @{\hspace{3pt}}r@{\hspace{3pt}}
    @{\hspace{3pt}}r@{\hspace{3pt}}
    @{\hspace{3pt}}r@{\hspace{3pt}}
    @{\hspace{3pt}}r@{\hspace{3pt}}
}
\toprule
\multirow{2}{*}{Task ID} 
& \multicolumn{6}{c}{\MPSTr (\deltaTr)} &
& \multicolumn{6}{c}{\APSTs} 
\\
\cmidrule(){2-7} \cmidrule(){9-14} 
& BBBP$^\uparrow$ & DRD2$^\uparrow$ & HIA$^\uparrow$ & Mutag$^\downarrow$ & plogP$^\uparrow$ & QED$^\uparrow$ 
&
& BBBP$^\uparrow$ & DRD2$^\uparrow$ & HIA$^\uparrow$ & Mutag$^\downarrow$ & plogP$^\uparrow$ & QED$^\uparrow$ 
\\
\midrule
{\mbox{$\mathop{\mathtt{BDP}}\limits$}\xspace} & 0.51 (0.32) & 0.04 (0.45) & - & - & -0.23 (1.98) & - && 0.34 & 0.01 & - & - & -2.33 & - \\
{\mbox{$\mathop{\mathtt{BDQ}}\limits$}\xspace} & 0.55 (0.32) & 0.04 (0.44) & - & - & - & 0.35 (0.24) && 0.37 & 0.01 & - & - & - & 0.21 \\
{\mbox{$\mathop{\mathtt{BPQ}}\limits$}\xspace} & 0.52 (0.36) & - & - & - & -1.51 (2.23) & 0.70 (0.17) && 0.31 & - & - & - & -2.87 & 0.41 \\
{\mbox{$\mathop{\mathtt{DPQ}}\limits$}\xspace} & - & 0.06 (0.48) & - & - & -0.84 (2.67) & 0.48 (0.21) && - & 0.01 & - & - & -3.32 & 0.36 \\
{\mbox{$\mathop{\mathtt{BDPQ}}\limits$}\xspace} & 0.51 (0.35) & 0.04 (0.51) &  & - & -1.15 (2.53) & 0.37 (0.25) && 0.26 & 0.02 & - & - & -4.92 & 0.24 \\

\hline

{\mbox{$\mathop{\mathtt{MPQ}}\limits$}\xspace} & - & - & - & 0.50 (-0.25) & -0.44 (1.86) & 0.72 (0.17) && - & - & - & 0.71 & -1.61 & 0.52 \\
{\mbox{$\mathop{\mathtt{BDMQ}}\limits$}\xspace} & 0.54 (0.31) & 0.04 (0.42) &  & 0.45 (-0.20) & - & 0.35 (0.23) && 0.34 & 0.01 & - & 0.58 & - & 0.19 \\
{\mbox{$\mathop{\mathtt{BHMQ}}\limits$}\xspace} & 0.43 (0.37) & - & 0.74 (0.28) & 0.46 (-0.19) & - & 0.70 (0.20) && 0.18 & - & 0.34 & 0.54 & - & 0.23 \\
{\mbox{$\mathop{\mathtt{BMPQ}}\limits$}\xspace} & 0.49 (0.33) & - & - & 0.47 (-0.22) & -0.76 (2.05) & 0.69 (0.19) && 0.33 & - & - & 0.65 & -2.27 & 0.38 \\
{\mbox{$\mathop{\mathtt{HMPQ}}\limits$}\xspace} & - & - & 0.71 (0.29) & 0.50 (-0.23) & -2.04 (2.30) & 0.62 (0.19) && - & - & 0.39 & 0.65 & -3.23 & 0.30 \\

\bottomrule
\end{tabular}

\begin{tablenotes}[normal,flushleft]
\footnotesize
\item 
{\MPSTr} and {\APSTs} denote the median and average property scores of the 
hit molecule $M_x$ in the training and test set, respectively.
{\deltaTr} denotes the average property difference across all training pairs.
\end{tablenotes}

\end{threeparttable}
\end{scriptsize}
\end{table*}

\begin{table*}[h!]
\centering
\caption{Overview of all 63 tasks in \MOptData}
\label{tbl:task_prop_all}
\begin{scriptsize}
\begin{threeparttable}
\begin{tabular}{
    @{\hspace{7pt}}l@{\hspace{7pt}}
    @{\hspace{7pt}}r@{\hspace{7pt}}
    @{\hspace{7pt}}r@{\hspace{10pt}}
    @{\hspace{7pt}}r@{\hspace{7pt}}
    @{\hspace{7pt}}r@{\hspace{7pt}}
    @{\hspace{7pt}}r@{\hspace{7pt}}
    @{\hspace{7pt}}r@{\hspace{7pt}}
    @{\hspace{7pt}}r@{\hspace{7pt}}
    @{\hspace{7pt}}r@{\hspace{7pt}}
}
\toprule
\multirow{2}{*}{Task ID} 
& \multirow{2}{*}{\#Pairs}
& \multirow{2}{*}{\#Mols} 
& \multicolumn{6}{c}{\APSTr (\deltaTr)} 
\\
\cmidrule(lr){4-9} 
& & & BBBP$^\uparrow$ & DRD2$^\uparrow$ & HIA$^\uparrow$ & Mutag$^\downarrow$ & plogP$^\uparrow$ & QED$^\uparrow$ 
\\
\midrule
{\mbox{$\mathop{\mathtt{B}}\limits$}\xspace} & 42,296 & 54,398 & 0.51 (0.32) & - & - & - & - & - \\
{\mbox{$\mathop{\mathtt{D}}\limits$}\xspace} & 116,755 & 60,651 & - & 0.14 (0.40) & - & - & - & - \\
{\mbox{$\mathop{\mathtt{H}}\limits$}\xspace} & 13,929 & 21,388 & - & - & 0.58 (0.34) & - & - & - \\
{\mbox{$\mathop{\mathtt{M}}\limits$}\xspace} & 110,586 & 116,604 & - & - & - & 0.47 (-0.21) & - & - \\
{\mbox{$\mathop{\mathtt{P}}\limits$}\xspace} & 105,949 & 149,774 & - & - & - & - & -0.96 (1.88) & - \\
{\mbox{$\mathop{\mathtt{Q}}\limits$}\xspace} & 157,863 & 167,041 & - & - & - & - & - & 0.61 (0.18) \\
{\mbox{$\mathop{\mathtt{BD}}\limits$}\xspace} & 11,703 & 9,804 & 0.50 (0.31) & 0.09 (0.41) & - & - & - & - \\
{\mbox{$\mathop{\mathtt{BH}}\limits$}\xspace} & 6,676 & 10,657 & 0.45 (0.39) & - & 0.53 (0.41) & - & - & - \\
{\mbox{$\mathop{\mathtt{BM}}\limits$}\xspace} & 4,126 & 5,637 & 0.50 (0.32) & - & - & 0.47 (-0.19) & - & - \\
{\mbox{$\mathop{\mathtt{BP}}\limits$}\xspace} & 12,197 & 19,360 & 0.49 (0.35) & - & - & - & -2.01 (2.21) & - \\
{\mbox{$\mathop{\mathtt{BQ}}\limits$}\xspace} & 24,577 & 33,303 & 0.52 (0.32) & - & - & - & - & 0.55 (0.22) \\
{\mbox{$\mathop{\mathtt{DH}}\limits$}\xspace} & 1,924 & 2,151 & - & 0.09 (0.43) & 0.68 (0.27) & - & - & - \\
{\mbox{$\mathop{\mathtt{DM}}\limits$}\xspace} & 22,834 & 19,359 & - & 0.12 (0.41) & - & 0.50 (-0.21) & - & - \\
{\mbox{$\mathop{\mathtt{DP}}\limits$}\xspace} & 19,154 & 16,796 & - & 0.12 (0.45) & - & - & -1.43 (1.90) & - \\
{\mbox{$\mathop{\mathtt{DQ}}\limits$}\xspace} & 26,860 & 19,109 & - & 0.15 (0.43) & - & - & - & 0.46 (0.21) \\
{\mbox{$\mathop{\mathtt{HM}}\limits$}\xspace} & 1,247 & 1,786 & - & - & 0.67 (0.25) & 0.49 (-0.22) & - & - \\
{\mbox{$\mathop{\mathtt{HP}}\limits$}\xspace} & 10,348 & 18,052 & - & - & 0.53 (0.37) & - & -2.83 (2.33) & - \\
{\mbox{$\mathop{\mathtt{HQ}}\limits$}\xspace} & 5,194 & 7,208 & - & - & 0.58 (0.35) & - & - & 0.64 (0.18) \\
{\mbox{$\mathop{\mathtt{MP}}\limits$}\xspace} & 27,554 & 43,224 & - & - & - & 0.48 (-0.22) & -0.50 (1.82) & - \\
{\mbox{$\mathop{\mathtt{MQ}}\limits$}\xspace} & 33,192 & 39,643 & - & - & - & 0.47 (-0.22) & - & 0.64 (0.17) \\
{\mbox{$\mathop{\mathtt{PQ}}\limits$}\xspace} & 15,874 & 22,621 & - & - & - & - & -2.18 (2.09) & 0.67 (0.16) \\
{\mbox{$\mathop{\mathtt{BDH}}\limits$}\xspace} & 829 & 997 & 0.47 (0.36) & 0.08 (0.45) & 0.65 (0.31) & - & - & - \\
{\mbox{$\mathop{\mathtt{BDM}}\limits$}\xspace} & 1,423 & 1,595 & 0.48 (0.30) & 0.07 (0.39) & - & 0.51 (-0.20) & - & - \\
{\mbox{$\mathop{\mathtt{BDP}}\limits$}\xspace} & 2,294 & 2,449 & 0.49 (0.32) & 0.08 (0.45) & - & - & -2.95 (1.98) & - \\
{\mbox{$\mathop{\mathtt{BDQ}}\limits$}\xspace} & 4,969 & 4,614 & 0.52 (0.32) & 0.09 (0.44) & - & - & - & 0.37 (0.24) \\
{\mbox{$\mathop{\mathtt{BHM}}\limits$}\xspace} & 380 & 617 & 0.41 (0.37) & - & 0.64 (0.31) & 0.46 (-0.19) & - & - \\
{\mbox{$\mathop{\mathtt{BHP}}\limits$}\xspace} & 5,259 & 9,141 & 0.45 (0.39) & - & 0.48 (0.45) & - & -3.05 (2.61) & - \\
{\mbox{$\mathop{\mathtt{BHQ}}\limits$}\xspace} & 3,100 & 4,619 & 0.43 (0.41) & - & 0.54 (0.40) & - & - & 0.64 (0.18) \\
{\mbox{$\mathop{\mathtt{BMP}}\limits$}\xspace} & 1,125 & 1,841 & 0.47 (0.32) & - & - & 0.49 (-0.21) & -1.07 (2.06) & - \\
{\mbox{$\mathop{\mathtt{BMQ}}\limits$}\xspace} & 2,414 & 3,388 & 0.52 (0.32) & - & - & 0.47 (-0.20) & - & 0.60 (0.19) \\
{\mbox{$\mathop{\mathtt{BPQ}}\limits$}\xspace} & 4,498 & 6,953 & 0.49 (0.36) & - & - & - & -2.92 (2.23) & 0.64 (0.17) \\
{\mbox{$\mathop{\mathtt{DHM}}\limits$}\xspace} & 304 & 301 & - & 0.07 (0.42) & 0.72 (0.22) & 0.60 (-0.27) & - & - \\
{\mbox{$\mathop{\mathtt{DHP}}\limits$}\xspace} & 967 & 1,246 & - & 0.07 (0.45) & 0.65 (0.28) & - & -3.15 (2.17) & - \\
{\mbox{$\mathop{\mathtt{DHQ}}\limits$}\xspace} & 708 & 790 & - & 0.08 (0.47) & 0.69 (0.27) & - & - & 0.44 (0.27) \\
{\mbox{$\mathop{\mathtt{DMP}}\limits$}\xspace} & 4,805 & 5,427 & - & 0.10 (0.45) & - & 0.50 (-0.23) & -0.54 (1.86) & - \\
{\mbox{$\mathop{\mathtt{DMQ}}\limits$}\xspace} & 6,398 & 6,220 & - & 0.14 (0.43) & - & 0.50 (-0.23) & - & 0.47 (0.21) \\
{\mbox{$\mathop{\mathtt{DPQ}}\limits$}\xspace} & 2,349 & 2,589 & - & 0.11 (0.48) & - & - & -5.53 (2.67) & 0.49 (0.21) \\
{\mbox{$\mathop{\mathtt{HMP}}\limits$}\xspace} & 730 & 1,211 & - & - & 0.62 (0.28) & 0.47 (-0.22) & -2.44 (2.22) & - \\
{\mbox{$\mathop{\mathtt{HMQ}}\limits$}\xspace} & 534 & 749 & - & - & 0.69 (0.25) & 0.52 (-0.22) & - & 0.59 (0.19) \\
{\mbox{$\mathop{\mathtt{HPQ}}\limits$}\xspace} & 3,142 & 5,289 & - & - & 0.49 (0.42) & - & -3.38 (2.39) & 0.65 (0.17) \\
{\mbox{$\mathop{\mathtt{MPQ}}\limits$}\xspace} & 3,481 & 5,384 & - & - & - & 0.51 (-0.25) & -1.44 (1.86) & 0.65 (0.17) \\
{\mbox{$\mathop{\mathtt{BDHM}}\limits$}\xspace} & 83 & 116 & 0.38 (0.32) & 0.09 (0.43) & 0.67 (0.28) & 0.56 (-0.23) & - & - \\
{\mbox{$\mathop{\mathtt{BDHP}}\limits$}\xspace} & 415 & 543 & 0.46 (0.35) & 0.06 (0.46) & 0.64 (0.32) & - & -5.63 (2.45) & - \\
{\mbox{$\mathop{\mathtt{BDHQ}}\limits$}\xspace} & 410 & 493 & 0.45 (0.38) & 0.07 (0.50) & 0.67 (0.29) & - & - & 0.39 (0.30) \\
{\mbox{$\mathop{\mathtt{BDMP}}\limits$}\xspace} & 321 & 438 & 0.43 (0.31) & 0.05 (0.40) & - & 0.59 (-0.23) & -0.40 (1.75) & - \\
{\mbox{$\mathop{\mathtt{BDMQ}}\limits$}\xspace} & 668 & 791 & 0.50 (0.31) & 0.10 (0.42) & - & 0.47 (-0.20) & - & 0.37 (0.23) \\
{\mbox{$\mathop{\mathtt{BDPQ}}\limits$}\xspace} & 694 & 802 & 0.49 (0.35) & 0.10 (0.51) & - & - & -8.34 (2.53) & 0.41 (0.25) \\
{\mbox{$\mathop{\mathtt{BHMP}}\limits$}\xspace} & 243 & 421 & 0.42 (0.35) & - & 0.60 (0.34) & 0.46 (-0.20) & -3.05 (2.83) & - \\
{\mbox{$\mathop{\mathtt{BHMQ}}\limits$}\xspace} & 213 & 333 & 0.42 (0.37) & - & 0.67 (0.28) & 0.49 (-0.19) & - & 0.60 (0.20) \\
{\mbox{$\mathop{\mathtt{BHPQ}}\limits$}\xspace} & 2,102 & 3,560 & 0.43 (0.41) & - & 0.45 (0.47) & - & -3.66 (2.58) & 0.64 (0.17) \\
{\mbox{$\mathop{\mathtt{BMPQ}}\limits$}\xspace} & 415 & 690 & 0.48 (0.33) & - & - & 0.50 (-0.22) & -1.48 (2.05) & 0.61 (0.19) \\
{\mbox{$\mathop{\mathtt{DHMP}}\limits$}\xspace} & 162 & 170 & - & 0.05 (0.45) & 0.71 (0.22) & 0.64 (-0.30) & -0.89 (1.77) & - \\
{\mbox{$\mathop{\mathtt{DHMQ}}\limits$}\xspace} & 130 & 136 & - & 0.07 (0.43) & 0.70 (0.23) & 0.65 (-0.31) & - & 0.43 (0.24) \\
{\mbox{$\mathop{\mathtt{DHPQ}}\limits$}\xspace} & 302 & 363 & - & 0.06 (0.48) & 0.64 (0.30) & - & -7.22 (2.45) & 0.47 (0.26) \\
{\mbox{$\mathop{\mathtt{DMPQ}}\limits$}\xspace} & 577 & 797 & - & 0.10 (0.46) & - & 0.51 (-0.26) & -1.97 (2.09) & 0.49 (0.22) \\
{\mbox{$\mathop{\mathtt{HMPQ}}\limits$}\xspace} & 260 & 417 & - & - & 0.63 (0.29) & 0.52 (-0.23) & -2.87 (2.30) & 0.60 (0.19) \\
{\mbox{$\mathop{\mathtt{BDHMP}}\limits$}\xspace} & 42 & 59 & 0.33 (0.30) & 0.06 (0.45) & 0.70 (0.25) & 0.63 (-0.26) & -1.89 (2.05) & - \\
{\mbox{$\mathop{\mathtt{BDHMQ}}\limits$}\xspace} & 50 & 72 & 0.41 (0.30) & 0.09 (0.47) & 0.68 (0.27) & 0.55 (-0.24) & - & 0.39 (0.28) \\
{\mbox{$\mathop{\mathtt{BDHPQ}}\limits$}\xspace} & 186 & 232 & 0.41 (0.40) & 0.06 (0.50) & 0.66 (0.30) & - & -10.79 (2.74) & 0.42 (0.30) \\
{\mbox{$\mathop{\mathtt{BDMPQ}}\limits$}\xspace} & 73 & 119 & 0.43 (0.31) & 0.10 (0.47) & - & 0.52 (-0.25) & -1.06 (1.61) & 0.43 (0.25) \\
{\mbox{$\mathop{\mathtt{BHMPQ}}\limits$}\xspace} & 120 & 199 & 0.43 (0.35) & - & 0.63 (0.31) & 0.51 (-0.21) & -3.48 (2.93) & 0.61 (0.20) \\
{\mbox{$\mathop{\mathtt{DHMPQ}}\limits$}\xspace} & 61 & 64 & - & 0.06 (0.46) & 0.67 (0.26) & 0.70 (-0.35) & -1.50 (1.71) & 0.46 (0.24) \\
{\mbox{$\mathop{\mathtt{BDHMPQ}}\limits$}\xspace} & 23 & 33 & 0.36 (0.29) & 0.07 (0.49) & 0.68 (0.27) & 0.64 (-0.29) & -2.88 (1.93) & 0.45 (0.26) \\

\bottomrule
\end{tabular}

\begin{tablenotes}[normal,flushleft]
\footnotesize
\item 
\#Pairs and \#Mols denote the unique number of training pairs and molecules in each task.
{\APSTr} and {\deltaTr} denote the average property score of $M_x$ and average property change across all training pairs in each task.
\end{tablenotes}

\end{threeparttable}
\end{scriptsize}
\end{table*}

\subsection{Details on Evaluation Tasks}
\label{sec:app:task:ind}

In this section, we provide descriptions of 
10 tasks in \MOptData used for evaluation.

\subsubsection{IND tasks}
\label{sec:app:task:ind}

Below are the 5 IND tasks:
\begin{enumerate}[leftmargin=*]

\item 
\BDP:
This task optimizes molecules to improve BBBP, DRD2 receptor inhibition, and lipophilicity (plogP). 
These properties are critical for central nervous system (CNS) drugs, where molecules must penetrate the blood-brain barrier, bind effectively to the DRD2 receptor (a common target for neurological disorders), and maintain sufficient lipophilicity for stability and membrane permeability.

\item \BDQ:
This task optimizes molecules to increase BBBP, DRD2 binding affinity, and improve QED. By balancing brain permeability, receptor activity, and drug-likeness, this task captures realistic trade-offs required in CNS drug development.

\item \BPQ:
This task aims to improve BBBP, plogP, and QED, prioritizing brain permeability and appropriate lipophilicity while ensuring the optimized molecules retain favorable drug-like properties.

\item \DPQ:
This task focuses on improving DRD2, plogP, and QED. It targets receptor binding potency while optimizing lipophilicity and maintaining overall drug-likeness, representing key requirements for receptor-specific drug design.

\item \BDPQ:
This task jointly optimizes BBBP, DRD2 activity, plogP, and QED, reflecting a challenging and comprehensive scenario for developing CNS drug candidates with high permeability, receptor activity, and drug-like characteristics.

\end{enumerate}

\subsubsection{OOD tasks}
\label{sec:app:task:ood}

Below are the 5 tasks used for evaluating out-of-domain generalizability:

\begin{enumerate}[leftmargin=*]
\item {\MPQ:}
This task focuses on reducing mutagenicity, improving plogP, and enhancing drug-likeness (QED). 
This task represents an early-stage lead optimization scenario
to reduce genotoxic risks while ensuring adequate lipophilicity and drug-like properties.

\item {\BDMQ:}
This task optimizes BBBP, DRD2 inhibition, mutagenicity, and QED. 
It reflects CNS drug development by balancing domapine receptor activity, brain permeability, and safety while ensuring overall drug-likeness.

\item {\BHMQ:}
This task focuses on increasing BBBP and HIA, reducing mutagenicity, and improving QED. It is particularly relevant for orally administered CNS drugs, where both brain and intestinal absorption are critical.

\item {\BMPQ:}
This task optimizes BBBP, mutagenicity, plogP, and QED. It reflects CNS drug design by balancing adequate lipophilicity, reduced toxicity, and favorable drug-like properties, simulating realistic requirements for CNS-active drugs.

\item {\HMPQ:}
This task enhances HIA, reduces mutagenicity, and improves plogP and QED. It represents optimization for orally administered drugs, focusing on absorption, genotoxic risk reduction, and overall drug-like quality.

\end{enumerate}

\subsection{Additional Filtering in Test Set}
\label{sec:app:filter}
Out of the initial pool of 250K molecules sampled from ZINC,
we select a molecule into the test set of a task
which has a property worse than 
the median \MPSTr.
Additionally, for properties with highly skewed distributions, we exclude
molecules 
falling below the 10th percentile of properties in training hit molecules,
thereby eliminating extreme cases (e.g., a molecule with a plogP of -30) that are rarely encountered as hits.
After applying these steps to the initial pool of 250K molecules,
we randomly select at most 500 molecules into the test set for each task, with possible overlap across tasks.
Table~\ref{tbl:task_prop} presents the property characteristics
of training pairs and test molecules
in all 10 evaluation tasks.
{Table~\ref{tbl:task_prop_all} presents the property characteristics of all 63 tasks in {\MOptData}.}

\subsection{Quality Control}
\label{sec:app:quality}

We implement multiple quality control measures to ensure
dataset integrity.
%
%
In \MOptData, molecules are represented as 
Simplified Molecular Input Line Entry System (SMILES)~\cite{Weininger1988smiles} strings
that are canonicalized and deduplicated.
%
For each molecule,
empirical property scores are computed using
well-established tools:
ADMET-AI~\cite{swanson2024admet}
for BBBP, HIA, Mutag and QED, 
and the official implementation provided by \citet{you2018graph} for DRD2 and plogP.
While these property scores are not experimentally validated, 
they provide reliable and computationally efficient estimates, 
making them well-suited for a large-scale dataset construction like ours.
%

{
We further emphasize that the computational property predictors used in this work (e.g., ADMET-AI) are among the top-performing models in the TDC benchmark~\cite{Catacutan2024} and have been widely adopted in recent literature~\cite{WEI2024,Thomas2024,Wahnou2024}. 
These predictors demonstrate strong alignment with experimentally measured values and serve as reliable proxies in practice, enabling scalable and consistent evaluation across large benchmarks.
Moreover, it is worth noting that experimentally validated data are extremely limited for many pharmacological properties. 
For example, BBBP measurements are only available for about 2k drug-like molecules in existing public datasets
-- far fewer than what is required to train deep learning models or LLMs at scale.
Meanwhile, correlating predicted improvements in generated molecules with experimental assays would strengthen real-world validation, but doing so would require wet-lab synthesis and testing at scale, which remains cost-prohibitive and logistically infeasible. 
As a result, most AI research in molecular optimization leverages predicted properties, making our approach consistent with prevailing practices.
}

We also ensure instruction diversity
to enhance the generalizability of instruction-tuned LLMs~\cite{xu2024wizardlm}.
We provide a manually written, clear and concise
seed instruction
into GPT-4~\cite{openai2024gpt4technicalreport}
to construct multiple distinctly phrased (i.e., diverse) instructions.
We select into \MOptData 5 diverse instructions 
synonymous
with the seed instruction.
To evaluate LLMs' instruction understanding and generalizability
to unseen instructions, 
we hold out one instruction for each task during training. 
Thus, each task in \MOptData has 5 diverse instructions
for instruction tuning, 
and 1 unseen instruction for testing.
All instructions are presented in Appendix~\ref{sec:app:instr}.


\subsection{Diverse Instructions}
\label{sec:app:instr}

Figure~\ref{fig:instr} presents the prompt
template used for instruction-tuning.
\begin{figure*}[h!]
\begin{tcolorbox}[
  colback=lightergray, 
  colframe=black, 
  sharp corners, 
  boxrule=0.5pt, 
  width=\textwidth,
  left=1mm, 
  right=1mm, 
  top=1mm, 
  bottom=1mm 
]
\begin{lstlisting}[
  language=, 
  basicstyle=\ttfamily\footnotesize, 
  breaklines=true, 
  breakindent=0pt, % Disable indentation for wrapped lines
  showstringspaces=false, % Remove visible spaces
  xleftmargin=0pt, % Remove left margin
  xrightmargin=0pt, % Remove right margin
  aboveskip=0pt,belowskip=0pt
]
[INST]
{instruction}

%%% Input : <SMILES> {source-smiles} </SMILES>
%%% Adjust: {change_i} {property_i}, ..., {change_k} {property_k}
[/INST]

%%% Response: {target-smiles}
\end{lstlisting}
\end{tcolorbox}
\caption{Prompt template used for instruction-tuning {\mollm}s}
\label{fig:instr}
\end{figure*}

The `\{instruction\}' will be replaced with one of 6
diverse instructions.
5 diverse instructions are used in training,
and 1 is held out for testing in the unseen instruction setting.
Below are the six diverse instructions, where the first one
is manually written, and the rest are generated by GPT-4o.
The last one is the hold-out instruction.

\begin{enumerate}[leftmargin=*]

\item 
``Your task is to modify the given molecule to adjust specific molecular properties while keeping structural changes as minimal as possible. Your response should only contain a valid SMILES representation of the modified molecule enclosed with <SMILES> </SMILES> tag."
    
    \item    ``Modify the given molecule to adjust the specified molecular properties by substituting functional groups while keeping changes to the core structure minimal. Output only the SMILES of the modified molecule, wrapped in <SMILES> </SMILES> tags."
    
    \item  ``Your goal is to fine-tune the specified molecular properties of the given compound with minimal structural changes. Make the necessary adjustments and return the modified molecule in a SMILES format enclosed in <SMILES> </SMILES> tags."

    \item
    ``Adjust the structure of the given molecule to target the specified adjustments in molecular properties. Retain the core structure as much as possible. Respond with only the SMILES of the modified molecule enclosed in <SMILES> </SMILES> tags."

    \item 
    ``Alter the given molecule to meet the desired property changes with the least structural alteration possible. Output only the adjusted molecule in SMILES format, using <SMILES> </SMILES> tags."
        
    \item
    ``Modify the given molecular structure to target specific property changes, aiming to keep structural adjustments minimal. Respond solely with the SMILES notation for the adjusted molecule, enclosed within <SMILES> </SMILES> tags."
    
\end{enumerate}

\paragraph{Property Names:}

We used the following names for each property 
where the former is used during instruction-tuning
and the latter is used for evaluation in
the unseen instruction setting.
For other evaluation settings, we used the same
property name as used in tuning.

\begin{enumerate}[leftmargin=*]
    \item BBBP: ``BBB permeability", ``Blood-brain barrier permeability (BBBP)"
    \item DRD2: ``DRD2 inhibition", ``inhibition probability of Dopamine receptor D2"
    \item HIA: ``Intestinal adsorption", ``human intestinal adsorption ability"
    \item Mutag: ``Mutagenicity",  ``probability to induce genetic alterations (mutagenicity)"
    \item plogP: ``Penalized octanol-water partition coefficient (penalized logP)", 
    ``Penalized logP which is logP penalized by synthetic accessibility score and number of large rings"
    \item QED: ``QED", ``drug-likeness quantified by QED score"
\end{enumerate}

\section{Details on Experimental Setup}
\label{sec:app:expts_setup}

\subsection{{\mollm}s}
\label{sec:app:reproducibility}
We develop a series of generalist {\mollm}s
which are trained on the power sets of 3, 4, and 6 properties,
denoted as \mollmTripleGen, \mollmQuadGen, and \mollmSixGen,
respectively.
To train these models, we fine-tune 2 general-purpose LLMs: 
Mistral-7B-Instruct-v0.3~\cite{mistral2023mistral} and Llama3.1-8B-Instruct~\cite{grattafiori2024llama3herdmodels}
using LoRA~\cite{hu2022lora},
leveraging the Huggingface Transformers library~\cite{wolf2020transformers}. 
We fine-tune all models with a learning rate of 1e-4 and a batch size of 128, using a cosine learning rate scheduler with a 5\% warm-up period. 
We fine-tune task-specific {\mollm}s 
and generalist {\mollm}s for 10 and 3 epochs, respectively,
to balance efficiency and overfitting. 
We set LoRA parameters with $\alpha=16$, dropout of 0.05, and a rank of 16,
and apply LoRA adapters to all projection layers and the language modeling head. 
We perform 0-shot evaluations
(i.e., without in-context examples) for all \mollm models 
in all tasks.
For each test molecule, we generate 20 molecules 
using beam search decoding, with the number of beams set to 20.

The number of trainable parameters varies from 42 million for Mistral-7B-Instruct-v0.3
to 44 million for Llama3.1-8B-Instruct.
Task-specific {\mollm}s need up to 1 hour on average on a NVIDIA H100 (Hopper) GPU for 10 epochs.
Generalist {\mollm}s take from 8 to 24 hours on average on the same GPU for 3 epochs,
depending on the number of tasks (property combinations).
In total, we spent about 120 GPU hours on an NVIDIA H100 GPU with 96 GB HBM2e memory.

\subsection{Baselines}
\label{sec:app:expts_setup:baselines}
In this section, we present the baselines considered and selected for our comparison. Table~\ref{tbl:baseline_LLMs} details the licenses and sources for both the datasets and models (i.e., artifacts).
We ensured that all artifacts used in this work were employed in a manner consistent with their intended use as specified by the original authors or licensors. 
For the models we developed, we identified ethical considerations
which are discussed in Section~\ref{sec:ethics}.

\paragraph{General-purpose LLMs:}

We evaluate 4 general-purpose LLMs:
2 open-weights LLMs, namely,
Mistral-7B Instruct-v0.3~\cite{mistral2023mistral}, 
Llama-3.1 8B-Instruct~\cite{touvron2023llama},  
and 2 closed-weights LLMs, namely,
Claude-3.5 and GPT-4o to assess the performance of such LLMs
in molecule optimization.
%
%
For Mistral-7B Instruct-v0.3 and Llama-3.1 8B-Instruct,
we use the officially released checkpoints provided in Huggingface.
For Claude-3.5,
we access the Sonnet checkpoints using the official API. 
We conduct both 0-shot and 5-shot inference (i.e., with 0 and 5 in-context examples, respectively) 
on all general-purpose LLMs
using the same prompt templates (Appendix~\ref{sec:app:prompt:glm}).
{
Due to the high cost of GPT-4o APIs, we adopt 1-shot prompting as a practical trade-off, as increasing to 5-shot yields minimal performance gains while significantly increasing inference cost.}

\paragraph{Foundational LLMs for Chemistry:}
We use \LlaSMolM (i.e., \LlaSMol tuned over the base model Mistral-7B) 
and \ChemLLM
as two foundational LLMs for chemistry since
they demonstrated state-of-the-art performance over
others, such as MolInst~\cite{fang2024molinstructions},
on a wide range
of molecular tasks.
%
We conduct only 0-shot inference
since we did not observe
any improvement with in-context examples in our preliminary experiments. 
We use a simpler prompt template 
(Appendix~\ref{sec:app:prompt:llasmol}) for inference
since \LlaSMol struggles to follow instructions in more
detailed prompts.


\paragraph{Task-specific non-LLM:}
%
%

%
We use \PMol as our task-specific non-LLM baseline since 
it demonstrated better performance over other methods such as Modof on single- and double-property optimization tasks. 
%
%
During inference, we leverage its 
embedding generator and
transformer modules
which are separately trained for each task. 
%
We discuss the training details and 
limitations of \PMol in
Appendix~\ref{sec:app:pmol} and \ref{sec:app:pmol:discuss}, respectively.

\begin{table*}[h]
\footnotesize
\centering
\caption{Licenses and Sources of Artifacts}
\label{tbl:baseline_LLMs}
\setlength{\tabcolsep}{0pt}%

\begin{small}
\begin{threeparttable}

\begin{tabular}{
    @{\hspace{2pt}}l@{\hspace{2pt}}
    @{\hspace{2pt}}p{0.45\textwidth}@{\hspace{2pt}}
    @{\hspace{2pt}}p{0.2\textwidth}@{\hspace{2pt}}
    @{\hspace{2pt}}l@{\hspace{2pt}}
}

\toprule
\textbf{Artifact} & \textbf{Source} & \textbf{License Type} & \textbf{Accessibility} \\
\midrule

\multirow{2}{*}{Modof} & \url{https://github.com/ziqi92/Modof} & {PolyForm Noncommercial License 1.0.0} & \multirow{2}{*}{Open Source}\\
\PMol & \url{https://github.com/wzxxxx/Prompt-MolOpt} & MIT License & {Open Source}\\

\multirow{2}{*}{\ChemLLM} & \url{https://huggingface.co/AI4Chem/ChemLLM-7B-Chat} & {Apache license 2.0} & \multirow{2}{*}{Checkpoint}\\

\multirow{2}{*}{\LlaSMolM} &
\url{https://huggingface.co/datasets/osunlp/SMolInstruct}
 & {Creative Commons Attribution 4.0} & \multirow{2}{*}{Checkpoint}\\

\multirow{2}{*}{Claude 3.5 (Sonnet)} & \url{https://docs.anthropic.com/claude/reference/getting-started-with-the-api} & 
\multirow{2}{*}{Proprietary} 
& \multirow{2}{*}{API} \\

GPT-4o & \url{https://openai.com/api/} & 
Proprietary
& API \\

\multirow{2}{*}{Llama-3.1 8B-Instruct} & \url{https://huggingface.co/meta-llama/Llama-3.1-8B-Instruct} & \multirow{2}{*}{Llama 3.1} & \multirow{2}{*}{Checkpoint} \\

\multirow{2}{*}{Mistral-7B-Instruct-v0.3} & \url{https://huggingface.co/mistralai/Mistral-7B-Instruct-v0.3} & \multirow{2}{*}{Apache license 2.0} & \multirow{2}{*}{Checkpoint} \\
\bottomrule
\end{tabular}

\end{threeparttable}
\end{small}

\end{table*}
\subsection{Evaluation Metrics}
\label{sec:app:eval}

We use the following evaluation metrics for a holistic comparison.

\begin{enumerate}[leftmargin=*]

\item \textbf{Success rate ({\SR}):} 
Success rate is the proportion
of test molecules for which at least one of 20 generated molecules has improvements in all desired properties.
If multiple generated molecules have improved properties, the one achieving the highest improvement across all properties is selected for evaluation.
Higher {\SR} demonstrates the models' ability to 
successfully optimize most hit molecules.

\item \textbf{Validity ({\Val}):}
Validity is the proportion of test hit molecules for which
at least one of 20 generated molecules is chemically valid.
A molecule is considered valid if it can be successfully
parsed by RDKit.
Higher validity indicates more test cases have valid generations.

\item \textbf{Similarity {({\Sim})}:}
{{\Sim}} denotes the average Tanimoto similarity
between successfully optimized molecules and the
corresponding test molecules.
The Tanimoto similarity is computed using binary Morgan fingerprints with a dimension of 2,048 and a radius of 2.
Higher {\Sim} indicates minimal structural modifications, 
which is desirable for retaining the core scaffold
-- a key requirement in lead optimization.

\item \textbf{Novelty ({\Nov}):} Novelty is defined as the 
percentage of optimized molecules that are unseen during training.
Higher {\Nov} indicates the models' ability to generate novel molecules, which is important for
ensuring chemical diversity and finding new lead molecules.

\item \textbf{Synthetic Accessibility Score ({\SAS}):}
SAS estimates how easily a molecule can be synthesized based on its structural complexity and the presence of uncommon fragments.
SAS generally ranges from 1 (easy to synthesize)
to 10 (challenging to synthesize~\cite{Ertl2009}.

\item \textbf{Relative Improvement ({\RI}):}
RI measures the average improvement in each property relative
to its initial score in the input hit molecule. 
Higher {\RI} implies significant improvements in desired properties on average.

\item \textbf{Average Property Score ({\APS}):}
{\APS} represents the average property score
of the generated optimized molecules in each property.
Higher {\APS} indicates that the model consistently generates molecules with better properties.

\end{enumerate}

\section{Prompt Templates}
\label{sec:app:prompt}

The following prompts are used to evaluate general-purpose LLMs
and \LlaSMol.

\subsection{Prompt Template for General-purpose LLMs}
\label{sec:app:prompt:glm}

We use a detailed prompt template
which is clearly structured with a system prompt,
task instruction, the input hit molecule, and task-specific optimization goals.
Figure~\ref{fig:prompt_glm} shows the prompt template with an example
task.

\begin{figure*}[!t]
\begin{tcolorbox}[
  colback=lightergray, 
  colframe=black, 
  sharp corners, 
  boxrule=0.5pt, 
  width=\textwidth,
  left=1mm, 
  right=1mm, 
  top=1mm, 
  bottom=1mm 
]
\begin{lstlisting}[
  language=, 
  basicstyle=\ttfamily\footnotesize, 
  breaklines=true, 
  breakindent=0pt, % Disable indentation for wrapped lines
  showstringspaces=false, % Remove visible spaces
  xleftmargin=0pt, % Remove left margin
  xrightmargin=0pt, % Remove right margin
  aboveskip=0pt,belowskip=0pt
]
<<SYS>>
You are an expert medicinal chemist specializing in molecular optimization. You understand how structural modifications affect key ADMET properties and inhibitions of common receptor targets like DRD2.
<</SYS>>

[INST]
Your task is to modify the given molecule to adjust specific molecular properties while keeping structural changes as minimal as possible. Use the examples (if provided) as a guide. Your response should only contain a valid  SMILES representation of the modified molecule enclosed with <SMILES> </SMILES> tag.

Examples:
%%% Input : <SMILES> COCC(=O)Nc1ccc(C(N)=O)cc1 </SMILES>
%%% Adjust: decrease Mutagenicity, increase Penalized octanol-water partition coefficient (penalized logP) and increase QED
%%% Response: <SMILES> COCC(=O)Nc1ccc(Br)cc1 </SMILES>

Task:
%%% Input : <SMILES> COC1COCCN(C(=O)c2ccno2)C1 </SMILES>
%%% Adjust: decrease Mutagenicity, increase Penalized octanol-water partition coefficient (penalized logP) and increase QED
[/INST]

%%% Response:
\end{lstlisting}
\end{tcolorbox}
\vspace{-10pt}
\caption{An example of a prompt used for general-purpose LLMs}
\label{fig:prompt_glm}
\end{figure*}

\subsection{Prompt Template for \LlaSMol}
\label{sec:app:prompt:llasmol}

Unlike general-purpose LLMs, \LlaSMol was instruction-tuned 
on multiple chemistry tasks
with a specific prompt template. 
We observed that \LlaSMol struggled with following instructions 
using the prompt template for general-purpose LLMs,
resulting in poor performance. 
Hence, we used a simpler prompt template without a system prompt
and without a clear separation of task instruction, input, and response.
Moreover, we only conduct 0-shot inference for \LlaSMol.
Figure~\ref{fig:prompt_llasmol} 
shows the prompt template with the same task
used above but without in-context examples:
\begin{figure*}
\begin{tcolorbox}[
  colback=lightergray, 
  colframe=black, 
  sharp corners, 
  boxrule=0.5pt, 
  width=\textwidth,
  left=1mm, 
  right=1mm, 
  top=1mm, 
  bottom=1mm 
]
\begin{lstlisting}[
  language=, 
  basicstyle=\ttfamily\footnotesize, 
  breaklines=true, 
  breakindent=0pt, % Disable indentation for wrapped lines
  showstringspaces=false, % Remove visible spaces
  xleftmargin=0pt, % Remove left margin
  xrightmargin=0pt, % Remove right margin
  aboveskip=0pt,belowskip=0pt
]
Modify the molecule <SMILES> COC1COCCN(C(=O)c2ccno2)C1 </SMILES> to decrease its Mutagenicity, increase its Penalized octanol-water partition coefficient (penalized logP) value, and increase its QED value. Keep the modifications to the molecule structure as minimal as possible.
%%% Response:
\end{lstlisting}
\end{tcolorbox}
\vspace{-10pt}
\caption{An example of a prompt used for \LlaSMol}
\label{fig:prompt_llasmol}
\end{figure*}

\section{Complete Experimental Results}
\label{sec:app:results}

\subsection{IND Evaluation}
\label{sec:app:results:ind}

Tables~\ref{tbl:bdp_ind}, \ref{tbl:bdq_ind}, \ref{tbl:bpq_ind}, \ref{tbl:dpq_ind} and \ref{tbl:bdpq_ind} present the performance comparison of {\mollm}s
with baselines on all 5 IND tasks.

\begin{table*}[h!]
\centering
\caption{Overall Performance on \BDP}
\setlength{\tabcolsep}{0pt}%
\label{tbl:bdp_ind}
\begin{small}
\begin{threeparttable}

\begin{tabular}{
    @{\hspace{9pt}}l@{\hspace{9pt}}
    @{\hspace{9pt}}r@{\hspace{9pt}}
    @{\hspace{9pt}}r@{\hspace{9pt}}
    @{\hspace{9pt}}r@{\hspace{9pt}}
    @{\hspace{9pt}}r@{\hspace{9pt}}
    @{\hspace{9pt}}r@{\hspace{9pt}}
    @{\hspace{9pt}}r@{\hspace{9pt}}
    @{\hspace{4pt}}r@{\hspace{4pt}}
    @{\hspace{4pt}}r@{\hspace{4pt}}
    @{\hspace{4pt}}r@{\hspace{4pt}}
}
\toprule
\multirow{2}{*}{Model} 
& \multirow{2}{*}{\SR$^{\uparrow}$}
& \multirow{2}{*}{\Val$^{\uparrow}$} 
& \multirow{2}{*}{\Sim$^{\uparrow}$} 
& \multirow{2}{*}{\Nov$^{\uparrow}$}
& \multirow{2}{*}{\SAS$^{\downarrow}$}
& \multirow{2}{*}{\RI$^{\uparrow}$}
& \multicolumn{3}{c}{\APS}
\\
\cmidrule(){8-10} 
& & & & & & & BBBP$^\uparrow$ & DRD2$^\uparrow$ & plogP$^\uparrow$ 
\\
\midrule

\rowcolor{lightgray}
\multicolumn{10}{c}{\textbf{General-purpose LLMs}} 
\\
Mistral (0-shot) & 6.60 & 75.00 & \textbf{\underline{0.81}} & \textbf{\underline{100.00}} & 3.72 & 0.68 & 0.47 & 0.03 & -0.91 \\
Llama (0-shot) & 22.00 & 85.60 & 0.73 & \textbf{\underline{100.00}} & 3.95 & 0.74 & 0.58 & 0.03 & -1.94 \\
Claude-3.5 (0-shot) & 19.60 & 94.40 & 0.66 & \textbf{\underline{100.00}} & 3.53 & 1.05 & 0.65 & 0.05 & -1.49 \\
GPT-4o (0-shot) & 7.80 & 87.20 & 0.69 & \textbf{\underline{100.00}} & 3.36 & 0.90 & 0.61 & 0.04 & -1.10 \\
Mistral (5-shot) & 35.20 & 95.20 & 0.64 & 96.59 & 3.42 & 2.10 & 0.66 & 0.11 & -0.87 \\
Llama (5-shot) & 35.40 & 96.80 & 0.57 & 79.10 & 3.50 & 2.71 & 0.64 & 0.17 & -0.83 \\
Claude-3.5 (5-shot) & 35.40 & 95.40 & 0.50 & \textbf{\underline{100.00}} & 3.18 & 2.43 & 0.77 & 0.10 & -0.45 \\
GPT-4o (1-shot) & 9.40 & 88.60 & 0.69 & \textbf{\underline{100.00}} & 3.59 & 0.79 & 0.62 & 0.04 & -0.91 \\

\rowcolor{lightgray}
\multicolumn{10}{c}{\textbf{Foundational LLMs for Chemistry}} 
\\
\ChemLLM & 0.20 & 12.80 & 0.17 & \textbf{\underline{100.00}} & \textbf{\underline{2.48}} & 1.20 & \textbf{\underline{0.97}} & 0.03 & 0.81 \\

\cellcolor{yellow!20}\LlaSMolM & \underline{\cellcolor{yellow!20}43.60} & \underline{\cellcolor{yellow!20}98.40} & \cellcolor{yellow!20}0.62 & \cellcolor{yellow!20}99.54 & \cellcolor{yellow!20}3.38 & \cellcolor{yellow!20}1.09 & \cellcolor{yellow!20}0.59 & \cellcolor{yellow!20}0.05 & \cellcolor{yellow!20}-1.09 \\

\rowcolor{lightgray}
\multicolumn{10}{c}{\textbf{Task-specific non-LLMs}} 
\\

\PMol & 12.20 & 20.80 & 0.12 & 96.72 & {2.66} & \textbf{\underline{7.46}} & {0.96} & \textbf{\underline{0.45}} & \textbf{\underline{1.59}} \\

\rowcolor{lightgray}
\multicolumn{10}{c}{\textbf{Task-specific LLMs}}
\\
\mollmTripleTaskM & 84.80 & 96.80 & 0.47 & \textbf{100.00} & 3.06 & 4.30 & 0.77 & 0.23 & 0.46 \\
\cellcolor{green!10}\mollmTripleTaskL & \textbf{\cellcolor{green!10}86.80} & \cellcolor{green!10}99.00 & \cellcolor{green!10}0.48 & \cellcolor{green!10}99.31 & \cellcolor{green!10}3.01 & \cellcolor{green!10}4.38 & \cellcolor{green!10}0.76 & \cellcolor{green!10}0.25 & \cellcolor{green!10}0.42 \\

\mollmQuadTaskM & 71.60 & 97.40 & 0.49 & 95.25 & 2.92 & 3.27 & 0.74 & 0.18 & 0.17 \\
\mollmQuadTaskL & 53.60 & 98.80 & 0.63 & 99.25 & 3.19 & 1.94 & 0.68 & 0.09 & -0.57 \\

\hline
\ImpT & 99.1 & 0.6 & -22.6 & -0.2 & 10.9 & 301.8 & 28.8 & 400.0 & 138.5 \\

\rowcolor{lightgray}
\multicolumn{10}{c}{\textbf{Generalist LLMs}} 
\\

\mollmTripleGenM & 75.60 & 98.20 & 0.56 & \textbf{100.00} & 3.18 & 3.31 & 0.68 & 0.16 & 0.07 \\
\mollmTripleGenL & 77.40 & 99.00 & 0.51 & 99.74 & 3.10 & 3.16 & 0.74 & 0.16 & 0.04 \\

\mollmQuadGenM & 81.40 & 98.80 & 0.55 & 99.75 & 3.07 & 3.95 & 0.73 & 0.19 & 0.12 \\

\mollmQuadGenL & 80.40 & \textbf{99.40} & 0.54 & 99.75 & 3.01 & 3.60 & 0.73 & 0.18 & 0.02 \\

\cellcolor{blue!10}\mollmSixGenM  & \cellcolor{blue!10}83.00 & \cellcolor{blue!10}98.80 & \cellcolor{blue!10}0.57 & \cellcolor{blue!10}99.76 & \cellcolor{blue!10}3.17 & \cellcolor{blue!10}3.60 & \cellcolor{blue!10}0.72 & \cellcolor{blue!10}0.17 & \cellcolor{blue!10}0.12 \\

\mollmSixGenL & 77.00 & \textbf{99.40} & 0.53 & 99.48 & 3.00 & 3.73 & 0.74 & 0.18 & 0.25 \\

\hline
\ImpG & 90.4 & 0.4 & -8.1 & 0.2 & 6.2 & 230.3 & 22.0 & 240.0 & 111.0 \\

\bottomrule
\end{tabular}

\begin{tablenotes}[normal,flushleft]
\footnotesize
\item $^\uparrow$ and $^\downarrow$ denote whether a higher or lower value of the metric is desirable, respectively.
For each task,
the best baseline performance is \underline{underlined} 
and the best overall performance is in \textbf{bold}
for each metric.
{\ImpT} and {\ImpG} denote the percentage improvement from the \colorbox{green!10}{best task-specific LLM} 
and \colorbox{blue!10}{best generalist LLM} over the 
\colorbox{yellow!20}{best baseline},
respectively,
where the best models are selected based on {\SR} for each task.
`-' indicates cases where models are trained on 3 properties but the task has additional properties 
not included in the model training.
\end{tablenotes}

\end{threeparttable}
\end{small}
\end{table*}
\begin{table*}[h!]
\centering
\caption{Overall Performance on \BDQ}
\setlength{\tabcolsep}{0pt}%
\label{tbl:bdq_ind}
\begin{small}
\begin{threeparttable}

\begin{tabular}{
    @{\hspace{9pt}}l@{\hspace{9pt}}
    @{\hspace{9pt}}r@{\hspace{9pt}}
    @{\hspace{9pt}}r@{\hspace{9pt}}
    @{\hspace{9pt}}r@{\hspace{9pt}}
    @{\hspace{9pt}}r@{\hspace{9pt}}
    @{\hspace{9pt}}r@{\hspace{9pt}}
    @{\hspace{9pt}}r@{\hspace{9pt}}
    @{\hspace{5pt}}r@{\hspace{5pt}}
    @{\hspace{5pt}}r@{\hspace{5pt}}
    @{\hspace{5pt}}r@{\hspace{5pt}}
}
\toprule
\multirow{2}{*}{Model} 
& \multirow{2}{*}{\SR$^{\uparrow}$}
& \multirow{2}{*}{\Val$^{\uparrow}$} 
& \multirow{2}{*}{\Sim$^{\uparrow}$} 
& \multirow{2}{*}{\Nov$^{\uparrow}$}
& \multirow{2}{*}{\SAS$^{\downarrow}$}
& \multirow{2}{*}{\RI$^{\uparrow}$}
& \multicolumn{3}{c}{\APS}
\\
\cmidrule(){8-10} 
& & & & & & & BBBP$^\uparrow$ & DRD2$^\uparrow$ & QED$^\uparrow$ 
\\
\midrule

\rowcolor{lightgray}
\multicolumn{10}{c}{\textbf{General-purpose LLMs}} 
\\
Mistral (0-shot) & 3.00 & 86.80 & \textbf{\underline{0.76}} & \textbf{\underline{100.00}} & 3.91 & 0.53 & 0.56 & 0.02 & 0.23 \\
Llama (0-shot) & 2.20 & 92.00 & 0.64 & \textbf{\underline{100.00}} & 4.23 & 0.53 & 0.49 & 0.02 & 0.22 \\
Claude-3.5 (0-shot) & 13.00 & 98.00 & 0.62 & \textbf{\underline{100.00}} & 2.96 & 1.14 & 0.61 & 0.04 & 0.35 \\
GPT-4o (0-shot) & 2.00 & 91.20 & 0.69 & \textbf{\underline{100.00}} & 2.81 & 0.62 & 0.63 & 0.03 & 0.33 \\
Mistral (5-shot) & 17.00 & 97.60 & 0.60 & 94.12 & 3.26 & 2.32 & 0.62 & 0.14 & 0.36 \\
Llama (5-shot) & 16.60 & 97.80 & 0.43 & 63.86 & 3.47 & \textbf{\underline{5.70}} & 0.70 & \textbf{\underline{0.26}} & 0.43 \\
Claude-3.5 (5-shot) & 29.40 & 94.20 & 0.43 & \textbf{\underline{100.00}} & \textbf{\underline{2.70}} & 3.80 & 0.79 & 0.17 & 0.51 \\
GPT-4o (1-shot) & 7.60 & 92.40 & 0.66 & \textbf{\underline{100.00}} & 2.83 & 0.61 & 0.64 & 0.04 & 0.31 \\

\rowcolor{lightgray}
\multicolumn{10}{c}{\textbf{Foundational LLMs for Chemistry}} 
\\
\ChemLLM & 1.00 & 21.20 & 0.55 & \textbf{\underline{100.00}} & 4.12 & 0.82 & 0.56 & 0.02 & 0.31 \\
\cellcolor{yellow!20}\LlaSMolM & \underline{\cellcolor{yellow!20}31.40} & \underline{\cellcolor{yellow!20}98.80} & \cellcolor{yellow!20}0.66 & \textbf{\underline{\cellcolor{yellow!20}100.00}} & \cellcolor{yellow!20}2.97 & \cellcolor{yellow!20}0.93 & \cellcolor{yellow!20}0.58 & \cellcolor{yellow!20}0.04 & \cellcolor{yellow!20}0.31 \\

\rowcolor{lightgray}
\multicolumn{10}{c}{\textbf{Task-specific non-LLMs}} 
\\

\PMol & 23.20 & 36.40 & 0.10 & \textbf{\underline{100.00}} & 3.51 & 5.40 & \textbf{\underline{0.95}} & 0.22 & \textbf{\underline{0.73}} \\

\rowcolor{lightgray}
\multicolumn{10}{c}{\textbf{Task-specific LLMs}}
\\
\mollmTripleTaskM & 87.00 & 98.80 & 0.47 & \textbf{100.00} & 2.90 & 5.61 & 0.79 & 0.22 & 0.49 \\
\cellcolor{green!10}\mollmTripleTaskL & \textbf{\cellcolor{green!10}90.00} & \cellcolor{green!10}99.40 & \cellcolor{green!10}0.46 & \cellcolor{green!10}99.78 & \cellcolor{green!10}2.93 & \cellcolor{green!10}5.66 & \cellcolor{green!10}0.81 & \cellcolor{green!10}0.23 & \cellcolor{green!10}0.51 \\

\mollmQuadTaskM & 57.40 & 99.20 & 0.55 & 99.65 & 3.05 & 2.56 & 0.70 & 0.11 & 0.43 \\
\mollmQuadTaskL & 48.60 & 99.60 & 0.59 & \textbf{100.00} & 3.36 & 1.29 & 0.64 & 0.06 & 0.37 \\

\hline
\ImpT & 186.6 & 0.6 & -30.3 & -0.2 & 1.3 & 508.6 & 39.7 & 475.0 & 64.5 \\

\rowcolor{lightgray}
\multicolumn{10}{c}{\textbf{Generalist LLMs}} 
\\

\mollmTripleGenM & 79.40 & 99.40 & 0.53 & 99.75 & 2.92 & 4.52 & 0.76 & 0.18 & 0.44 \\
\mollmTripleGenL & 76.40 & 99.60 & 0.57 & 99.74 & 2.90 & 4.41 & 0.74 & 0.17 & 0.41 \\

\mollmQuadGenM & 82.60 & \textbf{99.80} & 0.56 & 99.76 & 2.90 & 5.24 & 0.76 & 0.22 & 0.42 \\
\mollmQuadGenL & 81.40 & 99.60 & 0.56 & \textbf{100.00} & 2.87 & 4.81 & 0.76 & 0.19 & 0.42 \\

\rowcolor{blue!10}
\cellcolor{blue!10}\mollmSixGenM  & \cellcolor{blue!10}85.80 & \cellcolor{blue!10}99.60 & \cellcolor{blue!10}0.59 & \textbf{\cellcolor{blue!10}100.00} & \cellcolor{blue!10}2.94 & \cellcolor{blue!10}4.78 & \cellcolor{blue!10}0.75 & \cellcolor{blue!10}0.19 & \cellcolor{blue!10}0.40 \\

\mollmSixGenL & 79.60 & \textbf{99.80} & 0.56 & \textbf{100.00} & 2.87 & 5.05 & 0.76 & 0.20 & 0.42 \\

\hline
\ImpG & 173.2 & 0.8 & -10.6 & 0.0 & 1.0 & 414.0 & 29.3 & 375.0 & 29.0 \\

\bottomrule
\end{tabular}

\begin{tablenotes}[normal,flushleft]
\footnotesize
\item $^\uparrow$ and $^\downarrow$ denote whether a higher or lower value of the metric is desirable, respectively.
For each task,
the best baseline performance is \underline{underlined} 
and the best overall performance is in \textbf{bold}
for each metric.
{\ImpT} and {\ImpG} denote the percentage improvement from the \colorbox{green!10}{best task-specific LLM} 
and \colorbox{blue!10}{best generalist LLM} over the 
\colorbox{yellow!20}{best baseline},
respectively,
where the best models are selected based on {\SR} for each task.
\end{tablenotes}

\end{threeparttable}
\end{small}
\end{table*}
\begin{table*}[h!]
\centering
\caption{Overall Performance on \BPQ}
\setlength{\tabcolsep}{0pt}%
\label{tbl:bpq_ind}
\begin{small}
\begin{threeparttable}

\begin{tabular}{
    @{\hspace{9pt}}l@{\hspace{9pt}}
    @{\hspace{9pt}}r@{\hspace{9pt}}
    @{\hspace{9pt}}r@{\hspace{9pt}}
    @{\hspace{9pt}}r@{\hspace{9pt}}
    @{\hspace{9pt}}r@{\hspace{9pt}}
    @{\hspace{9pt}}r@{\hspace{9pt}}
    @{\hspace{9pt}}r@{\hspace{9pt}}
    @{\hspace{5pt}}r@{\hspace{5pt}}
    @{\hspace{5pt}}r@{\hspace{5pt}}
    @{\hspace{5pt}}r@{\hspace{5pt}}
}
\toprule
\multirow{2}{*}{Model} 
& \multirow{2}{*}{\SR$^{\uparrow}$}
& \multirow{2}{*}{\Val$^{\uparrow}$} 
& \multirow{2}{*}{\Sim$^{\uparrow}$} 
& \multirow{2}{*}{\Nov$^{\uparrow}$}
& \multirow{2}{*}{\SAS$^{\downarrow}$}
& \multirow{2}{*}{\RI$^{\uparrow}$}
& \multicolumn{3}{c}{\APS}
\\
\cmidrule(){8-10} 
& & & & & & & BBBP$^\uparrow$ & plogP$^\uparrow$ & QED$^\uparrow$ 
\\
\midrule

\rowcolor{lightgray}
\multicolumn{10}{c}{\textbf{General-purpose LLMs}} 
\\
Mistral (0-shot) & 15.80 & 75.20 & {0.73} & \textbf{\underline{100.00}} & 3.75 & 0.51 & 0.47 & -1.75 & 0.47 \\
Llama (0-shot) & 28.40 & 89.40 & 0.64 & \textbf{\underline{100.00}} & 3.91 & 0.72 & 0.59 & -1.99 & 0.51 \\
Claude-3.5 (0-shot) & 56.00 & 93.00 & 0.62 & \textbf{\underline{100.00}} & 3.68 & 0.86 & 0.58 & -1.35 & 0.56 \\
GPT-4o (0-shot) & 36.40 & 89.00 & 0.73 & \textbf{\underline{100.00}} & 3.62 & 0.42 & 0.50 & -1.90 & 0.55 \\
Mistral (5-shot) & 68.60 & 89.80 & 0.63 & 98.54 & 3.44 & 0.79 & 0.66 & -1.35 & 0.59 \\
Llama (5-shot) & 34.60 & 94.00 & 0.70 & 98.84 & 3.66 & 0.64 & 0.58 & -1.64 & 0.53 \\
Claude-3.5 (5-shot) & 76.80 & 95.40 & 0.53 & 99.74 & 3.30 & 1.24 & 0.69 & -0.45 & 0.64 \\
GPT-4o (1-shot) & 40.00 & 90.00 & \textbf{\underline{0.75}} & \textbf{\underline{100.00}} & 3.79 & 0.41 & 0.47 & -1.95 & 0.50 \\

\rowcolor{lightgray}
\multicolumn{10}{c}{\textbf{Foundational LLMs for Chemistry}} 
\\
\ChemLLM & 4.80 & 19.80 & 0.29 & \textbf{\underline{100.00}} & 3.12 & 0.96 & 0.84 & -1.04 & 0.62 \\
\cellcolor{yellow!20}\LlaSMolM & \underline{\cellcolor{yellow!20}86.00} & \underline{\cellcolor{yellow!20}96.00} & \cellcolor{yellow!20}0.58 & \cellcolor{yellow!20}98.37 & \cellcolor{yellow!20}3.37 & \cellcolor{yellow!20}0.84 & \cellcolor{yellow!20}0.62 & \cellcolor{yellow!20}-1.17 & \cellcolor{yellow!20}0.62 \\

\rowcolor{lightgray}
\multicolumn{10}{c}{\textbf{Task-specific non-LLMs}} 
\\

\PMol & 15.80 & 16.60 & 0.10 & 98.73 & \textbf{\underline{2.64}} & \underline{1.50} & \textbf{\underline{0.96}} & \textbf{\underline{1.26}} & \textbf{\underline{0.76}} \\

\rowcolor{lightgray}
\multicolumn{10}{c}{\textbf{Task-specific LLMs}}
\\
\mollmTripleTaskM & 93.00 & 97.40 & 0.46 & 99.57 & 3.14 & 1.49 & 0.77 & -0.16 & 0.69 \\
\cellcolor{green!10}\mollmTripleTaskL & \cellcolor{green!10}94.00 & \cellcolor{green!10}98.20 & \cellcolor{green!10}0.50 & \cellcolor{green!10}98.94 & \cellcolor{green!10}3.13 & \cellcolor{green!10}1.38 & \cellcolor{green!10}0.77 & \cellcolor{green!10}-0.22 & \cellcolor{green!10}0.67 \\

\mollmQuadTaskM & 90.20 & 96.40 & 0.46 & 97.78 & 2.97 & 1.41 & 0.77 & 0.01 & 0.69 \\
\mollmQuadTaskL & 93.40 & 97.40 & 0.59 & 99.36 & 3.28 & 1.12 & 0.71 & -0.69 & 0.63 \\

\hline
\ImpT & 9.3 & 2.3 & -13.8 & 0.6 & 7.1 & 64.3 & 24.2 & 81.2 & 8.1 \\

\rowcolor{lightgray}
\multicolumn{10}{c}{\textbf{Generalist LLMs}} 
\\

\mollmTripleGenM & 93.20 & 97.40 & 0.55 & 99.79 & 3.30 & 1.23 & 0.72 & -0.51 & 0.65 \\
\mollmTripleGenL & 95.40 & \textbf{99.60} & 0.50 & 99.58 & 3.14 & 1.46 & 0.77 & -0.17 & 0.68 \\

\mollmQuadGenM & 96.20 & 98.60 & 0.52 & 99.58 & 3.11 & 1.52 & 0.78 & -0.18 & 0.69 \\
\mollmQuadGenL & 93.80 & 99.20 & 0.47 & 98.72 & 3.03 & 1.64 & 0.82 & -0.04 & 0.70 \\

\cellcolor{blue!10}\mollmSixGenM  & \textbf{\cellcolor{blue!10}96.80} & \cellcolor{blue!10}99.20 & \cellcolor{blue!10}0.53 & \cellcolor{blue!10}99.38 & \cellcolor{blue!10}3.23 & \cellcolor{blue!10}1.48 & \cellcolor{blue!10}0.78 & \cellcolor{blue!10}-0.34 & \cellcolor{blue!10}0.68 \\

\mollmSixGenL & 95.00 & 98.60 & 0.47 & 99.58 & 2.98 & \textbf{1.66} & 0.81 & 0.16 & 0.71 \\

\hline
\ImpG & 12.6 & 3.3 & -8.6 & 1.0 & 4.2 & 76.2 & 25.8 & 70.9 & 9.7 \\

\bottomrule
\end{tabular}

\begin{tablenotes}[normal,flushleft]
\footnotesize
\item $^\uparrow$ and $^\downarrow$ denote whether a higher or lower value of the metric is desirable, respectively.
For each task,
the best baseline performance is \underline{underlined} 
and the best overall performance is in \textbf{bold}
for each metric.
{\ImpT} and {\ImpG} denote the percentage improvement from the \colorbox{green!10}{best task-specific LLM} 
and \colorbox{blue!10}{best generalist LLM} over the 
\colorbox{yellow!20}{best baseline},
respectively,
where the best models are selected based on {\SR} for each task.
\end{tablenotes}

\end{threeparttable}
\end{small}
\end{table*}
\begin{table*}[h!]
\centering
\caption{Overall Performance on \DPQ}
\setlength{\tabcolsep}{0pt}%
\label{tbl:dpq_ind}
\begin{small}
\begin{threeparttable}

\begin{tabular}{
    @{\hspace{9pt}}l@{\hspace{9pt}}
    @{\hspace{9pt}}r@{\hspace{9pt}}
    @{\hspace{9pt}}r@{\hspace{9pt}}
    @{\hspace{9pt}}r@{\hspace{9pt}}
    @{\hspace{9pt}}r@{\hspace{9pt}}
    @{\hspace{9pt}}r@{\hspace{9pt}}
    @{\hspace{9pt}}r@{\hspace{9pt}}
    @{\hspace{5pt}}r@{\hspace{5pt}}
    @{\hspace{5pt}}r@{\hspace{5pt}}
    @{\hspace{5pt}}r@{\hspace{5pt}}
}
\toprule
\multirow{2}{*}{Model} 
& \multirow{2}{*}{\SR$^{\uparrow}$}
& \multirow{2}{*}{\Val$^{\uparrow}$} 
& \multirow{2}{*}{\Sim$^{\uparrow}$} 
& \multirow{2}{*}{\Nov$^{\uparrow}$}
& \multirow{2}{*}{\SAS$^{\downarrow}$}
& \multirow{2}{*}{\RI$^{\uparrow}$}
& \multicolumn{3}{c}{\APS}
\\
\cmidrule(){8-10} 
& & & & & & & DRD2$^\uparrow$ & plogP$^\uparrow$ & QED$^\uparrow$ 
\\
\midrule

\rowcolor{lightgray}
\multicolumn{10}{c}{\textbf{General-purpose LLMs}} 
\\
Mistral (0-shot) & 2.20 & 75.20 & \textbf{\underline{0.65}} & \textbf{\underline{100.00}} & 4.15 & 0.41 & 0.03 & -2.79 & 0.44 \\
Llama (0-shot) & 2.60 & 87.60 & 0.62 & \textbf{\underline{100.00}} & 4.01 & 0.32 & 0.03 & -1.79 & 0.49 \\
Claude-3.5 (0-shot) & 11.00 & 95.80 & 0.54 & \textbf{\underline{100.00}} & 3.70 & 0.51 & 0.04 & -1.30 & 0.54 \\
GPT-4o (0-shot) & 2.80 & 86.00 & 0.57 & \textbf{\underline{100.00}} & 3.55 & 0.50 & 0.03 & -1.43 & 0.50 \\
Mistral (5-shot) & 10.40 & 92.60 & 0.54 & 92.31 & 3.49 & 1.10 & 0.13 & -1.33 & 0.54 \\
Llama (5-shot) & 8.20 & 96.00 & 0.44 & 60.98 & 3.51 & 3.02 & 0.24 & -0.83 & 0.59 \\
\cellcolor{yellow!20}Claude-3.5 (5-shot) & \underline{\cellcolor{yellow!20}29.20} & \cellcolor{yellow!20}92.60 & \cellcolor{yellow!20}0.37 & \cellcolor{yellow!20}98.63 & \cellcolor{yellow!20}3.02 & \cellcolor{yellow!20}2.87 & \cellcolor{yellow!20}0.16 & \cellcolor{yellow!20}0.08 & \cellcolor{yellow!20}0.64 \\
GPT-4o (1-shot) & 7.00 & 86.60 & 0.62 & \textbf{\underline{100.00}} & 3.81 & 0.44 & 0.03 & -1.58 & 0.47 \\

\rowcolor{lightgray}
\multicolumn{10}{c}{\textbf{Foundational LLMs for Chemistry}} 
\\

\ChemLLM & 0.60 & 18.20 & 0.28 & \textbf{\underline{100.00}} & 3.31 & 0.42 & 0.03 & -2.01 & 0.53 \\
\LlaSMolM & 24.00 & \underline{97.60} & 0.57 & \textbf{\underline{100.00}} & 3.53 & 0.61 & 0.04 & -1.64 & 0.51 \\

\rowcolor{lightgray}
\multicolumn{10}{c}{\textbf{Task-specific non-LLMs}} 
\\

\PMol & 23.60 & 31.60 & 0.10 & 96.61 & \textbf{\underline{2.58}} & \textbf{\underline{5.46}} & \textbf{\underline{0.46}} & \textbf{\underline{1.08}} & \textbf{\underline{0.80}} \\

\rowcolor{lightgray}
\multicolumn{10}{c}{\textbf{Task-specific LLMs}}
\\
\cellcolor{green!10}\mollmTripleTaskM & \cellcolor{green!10}62.80 & \cellcolor{green!10}98.20 & \cellcolor{green!10}0.37 & \cellcolor{green!10}97.45 & \cellcolor{green!10}3.04 & \cellcolor{green!10}3.87 & \cellcolor{green!10}0.24 & \cellcolor{green!10}-0.08 & \cellcolor{green!10}0.64 \\

\mollmTripleTaskL & 60.60 & 99.00 & 0.44 & 97.03 & 3.08 & 3.76 & 0.24 & -0.12 & 0.59 \\

\mollmQuadTaskM & 54.00 & 96.60 & 0.44 & 94.07 & 3.01 & 3.02 & 0.21 & -0.54 & 0.58 \\
\mollmQuadTaskL & 39.60 & 98.40 & 0.57 & 98.99 & 3.36 & 1.32 & 0.08 & -1.20 & 0.54 \\

\hline
\ImpT & 115.1 & 6.0 & 0.0 & -1.2 & -0.7 & 34.8 & 50.0 & -200.0 & 0.0 \\

\rowcolor{lightgray}
\multicolumn{10}{c}{\textbf{Generalist LLMs}} 
\\

\mollmTripleGenM & 57.20 & 98.20 & 0.50 & 99.65 & 3.26 & 2.22 & 0.13 & -0.57 & 0.58 \\
\mollmTripleGenL & 63.40 & \textbf{99.80} & 0.49 & \textbf{100.00} & 3.17 & 2.46 & 0.14 & -0.45 & 0.59 \\

\cellcolor{blue!10}\mollmQuadGenM & \textbf{\cellcolor{blue!10}66.60} & \cellcolor{blue!10}99.20 & \cellcolor{blue!10}0.53 & \cellcolor{blue!10}99.40 & \cellcolor{blue!10}3.26 & \cellcolor{blue!10}2.41 & \cellcolor{blue!10}0.13 & \cellcolor{blue!10}-0.69 & \cellcolor{blue!10}0.55 \\

\mollmQuadGenL & 61.40 & 99.00 & 0.50 & \textbf{100.00} & 3.16 & 2.02 & 0.12 & -0.40 & 0.58 \\

\mollmSixGenM  & 60.80 & 99.40 & 0.54 & 99.67 & 3.31 & 2.16 & 0.12 & -0.57 & 0.57 \\

\mollmSixGenL & 57.00 & 99.00 & 0.49 & 99.65 & 3.14 & 2.50 & 0.14 & -0.36 & 0.58 \\

\hline
\ImpG & 128.1 & 7.1 & 43.2 & 0.8 & -7.9 & -16.0 & -18.8 & -962.5 & -14.1 \\

\bottomrule
\end{tabular}

\begin{tablenotes}[normal,flushleft]
\footnotesize
\item $^\uparrow$ and $^\downarrow$ denote whether a higher or lower value of the metric is desirable, respectively.
For each task,
the best baseline performance is \underline{underlined} 
and the best overall performance is in \textbf{bold}
for each metric.
{\ImpT} and {\ImpG} denote the percentage improvement from the \colorbox{green!10}{best task-specific LLM} 
and \colorbox{blue!10}{best generalist LLM} over the 
\colorbox{yellow!20}{best baseline},
respectively,
where the best models are selected based on {\SR} for each task.
\end{tablenotes}

\end{threeparttable}
\end{small}
\end{table*}
\begin{table*}[h!]
\centering
\caption{Overall Performance on \BDPQ}
\setlength{\tabcolsep}{0pt}%
\label{tbl:bdpq_ind}
\begin{small}
\begin{threeparttable}

\begin{tabular}{
    @{\hspace{9pt}}l@{\hspace{9pt}}
    @{\hspace{9pt}}r@{\hspace{9pt}}
    @{\hspace{9pt}}r@{\hspace{9pt}}
    @{\hspace{9pt}}r@{\hspace{9pt}}
    @{\hspace{9pt}}r@{\hspace{9pt}}
    @{\hspace{9pt}}r@{\hspace{9pt}}
    @{\hspace{9pt}}r@{\hspace{9pt}}
    @{\hspace{5pt}}r@{\hspace{5pt}}
    @{\hspace{5pt}}r@{\hspace{5pt}}
    @{\hspace{5pt}}r@{\hspace{5pt}}
    @{\hspace{5pt}}r@{\hspace{5pt}}
}
\toprule
\multirow{2}{*}{Model} 
& \multirow{2}{*}{\SR$^{\uparrow}$}
& \multirow{2}{*}{\Val$^{\uparrow}$} 
& \multirow{2}{*}{\Sim$^{\uparrow}$} 
& \multirow{2}{*}{\Nov$^{\uparrow}$}
& \multirow{2}{*}{\SAS$^{\downarrow}$}
& \multirow{2}{*}{\RI$^{\uparrow}$}
& \multicolumn{3}{c}{\APS}
\\
\cmidrule(){8-11} 
& & & & & & & BBBP$^\uparrow$ & DRD2$^\uparrow$ & plogP$^\uparrow$ & QED$^\uparrow$ 
\\
\midrule

\rowcolor{lightgray}
\multicolumn{11}{c}{\textbf{General-purpose LLMs}} 
\\
Mistral (0-shot) & 3.20 & 67.00 & 0.77 & \textbf{\underline{100.00}} & 4.26 & 0.87 & 0.53 & 0.03 & -1.85 & 0.32 \\
Llama (0-shot) & 5.20 & 83.40 & \textbf{\underline{0.80}} & \textbf{\underline{100.00}} & 4.52 & 0.62 & 0.46 & 0.02 & -2.73 & 0.23 \\
Claude-3.5 (0-shot) & 8.00 & \underline{94.80} & 0.60 & \textbf{\underline{100.00}} & 3.77 & 1.34 & 0.49 & 0.06 & -3.29 & 0.40 \\
GPT-4o (0-shot) & 1.80 & 87.00 & 0.71 & \textbf{\underline{100.00}} & 4.08 & 0.39 & 0.41 & 0.02 & -1.64 & 0.33 \\
Mistral (5-shot) & 11.00 & 79.00 & 0.69 & 98.18 & 3.71 & 0.96 & 0.57 & 0.06 & -3.25 & 0.41 \\
Llama (5-shot) & 9.60 & 89.20 & 0.54 & 72.92 & 3.75 & 3.45 & 0.57 & 0.15 & -2.04 & 0.40 \\
\cellcolor{yellow!20}Claude-3.5 (5-shot) & \underline{\cellcolor{yellow!20}20.80} & \cellcolor{yellow!20}93.00 & \cellcolor{yellow!20}0.35 & \cellcolor{yellow!20}98.08 & \cellcolor{yellow!20}3.04 & \cellcolor{yellow!20}3.53 & \cellcolor{yellow!20}0.77 & \cellcolor{yellow!20}0.15 & \cellcolor{yellow!20}-0.58 & \cellcolor{yellow!20}0.61 \\
GPT-4o (1-shot) & 3.40 & 88.20 & 0.70 & \textbf{\underline{100.00}} & 3.66 & 0.61 & 0.46 & 0.03 & -0.84 & 0.39 \\

\rowcolor{lightgray}
\multicolumn{11}{c}{\textbf{Foundational LLMs for Chemistry}} 
\\
\ChemLLM & 0.00 & 10.60 & n/a & n/a & n/a & n/a & n/a & n/a & n/a & n/a  \\
\LlaSMolM  & 14.00 & 90.20 & 0.62 & 98.57 & 3.48 & 1.03 & 0.50 & 0.06 & -1.97 & 0.44 \\

\rowcolor{lightgray}
\multicolumn{11}{c}{\textbf{Task-specific non-LLMs}} 
\\

\PMol  & 6.60 & 21.80 & 0.11 & \textbf{\underline{100.00}} & \textbf{\underline{2.70}} & \textbf{\underline{5.36}} & \textbf{\underline{0.92}} & \textbf{\underline{0.39}} & \textbf{\underline{0.51}} & \textbf{\underline{0.77}} \\

\rowcolor{lightgray}
\multicolumn{11}{c}{\textbf{Task-specific LLMs}}
\\

\cellcolor{green!10}\mollmQuadTaskM & \cellcolor{green!10}30.00 & \cellcolor{green!10}93.00 & \cellcolor{green!10}0.48 & \cellcolor{green!10}95.33 & \cellcolor{green!10}3.02 & \cellcolor{green!10}3.44 & \cellcolor{green!10}0.65 & \cellcolor{green!10}0.17 & \cellcolor{green!10}-1.55 & \cellcolor{green!10}0.53 \\

\mollmQuadTaskL & 28.00 & 94.00 & 0.66 & 98.57 & 3.57 & 1.02 & 0.56 & 0.05 & -2.68 & 0.42 \\

\hline
\ImpT & 44.2 & 0.0 & 37.1 & -2.8 & 0.7 & -2.5 & -15.6 & 13.3 & -167.2 & -13.1 \\

\rowcolor{lightgray}
\multicolumn{11}{c}{\textbf{Generalist LLMs}} 
\\

\cellcolor{blue!10}\mollmQuadGenM  & \textbf{\cellcolor{blue!10}57.40} & \textbf{\cellcolor{blue!10}97.60} & \cellcolor{blue!10}0.52 & \cellcolor{blue!10}99.65 & \cellcolor{blue!10}3.29 & \cellcolor{blue!10}3.04 & \cellcolor{blue!10}0.65 & \cellcolor{blue!10}0.15 & \cellcolor{blue!10}-0.88 & \cellcolor{blue!10}0.49 \\

\mollmQuadGenL & 49.80 & 97.40 & 0.48 & \textbf{100.00} & 3.18 & 3.26 & 0.68 & 0.16 & -0.69 & 0.52 \\

\mollmSixGenM  & 54.00 & 97.40 & 0.54 & 99.26 & 3.34 & 3.09 & 0.65 & 0.16 & -0.93 & 0.48 \\

\mollmSixGenL  & 52.20 & 97.20 & 0.49 & 99.23 & 3.17 & 3.48 & 0.69 & 0.16 & -0.65 & 0.53 \\

\hline
\ImpG & 176.0 & 4.9 & 48.6 & 1.6 & -8.2 & -13.9 & -15.6 & 0.0 & -51.7 & -19.7 \\

\bottomrule
\end{tabular}

\begin{tablenotes}[normal,flushleft]
\footnotesize
\item $^\uparrow$ and $^\downarrow$ denote whether a higher or lower value of the metric is desirable, respectively.
For each task,
the best baseline performance is \underline{underlined} 
and the best overall performance is in \textbf{bold}
for each metric.
{\ImpT} and {\ImpG} denote the percentage improvement from the \colorbox{green!10}{best task-specific LLM} 
and \colorbox{blue!10}{best generalist LLM} over the 
\colorbox{yellow!20}{best baseline},
respectively,
where the best models are selected based on {\SR} for each task.
`n/a' indicates undefined metrics due to no optimized cases (i.e., 0\% \SR).
\end{tablenotes}

\end{threeparttable}
\end{small}
\end{table*}

\subsection{OOD Evaluation}
\label{sec:app:results:ood}

Tables~\ref{tbl:mpq_ood}, \ref{tbl:bdmq_ood}, \ref{tbl:bhmq_ood}, \ref{tbl:bmpq_ood} and \ref{tbl:hmpq_ood} present the performance comparison of {\mollm}s
with baselines on all 5 OOD tasks.

Since OOD tasks represent novel property combinations excluded from the training data,
task-specific models are not applicable in this setting.
Additionally, several properties in these tasks are not used in training
generalist models \mollmTripleGen and \mollmQuadGen,
making comparison with these models infeasible.

\begin{table*}[h!]
\centering
\caption{Overall Performance on \MPQ}
\setlength{\tabcolsep}{0pt}%
\label{tbl:mpq_ood}
\begin{small}
\begin{threeparttable}

\begin{tabular}{
    @{\hspace{9pt}}l@{\hspace{9pt}}
    @{\hspace{9pt}}r@{\hspace{9pt}}
    @{\hspace{9pt}}r@{\hspace{9pt}}
    @{\hspace{9pt}}r@{\hspace{9pt}}
    @{\hspace{9pt}}r@{\hspace{9pt}}
    @{\hspace{9pt}}r@{\hspace{9pt}}
    @{\hspace{9pt}}r@{\hspace{9pt}}
    @{\hspace{5pt}}r@{\hspace{5pt}}
    @{\hspace{5pt}}r@{\hspace{5pt}}
    @{\hspace{5pt}}r@{\hspace{5pt}}
}
\toprule
\multirow{2}{*}{Model} 
& \multirow{2}{*}{\SR$^{\uparrow}$}
& \multirow{2}{*}{\Val$^{\uparrow}$} 
& \multirow{2}{*}{\Sim$^{\uparrow}$} 
& \multirow{2}{*}{\Nov$^{\uparrow}$}
& \multirow{2}{*}{\SAS$^{\downarrow}$}
& \multirow{2}{*}{\RI$^{\uparrow}$}
& \multicolumn{3}{c}{\APS}
\\
\cmidrule(){8-10} 
& & & & & & & Mutag$^\downarrow$ & plogP$^\uparrow$ & QED$^\uparrow$ 
\\
\midrule

\rowcolor{lightgray}
\multicolumn{10}{c}{\textbf{General-purpose LLMs}} 
\\
Mistral (0-shot) & 11.20 & 79.40 & 0.57 & \textbf{\underline{100.00}} & 2.84 & 0.48 & 0.49 & -0.33 & 0.61 \\
Llama (0-shot) & 25.80 & 89.20 & 0.44 & 99.22 & 2.89 & 0.61 & \underline{0.37} & -0.41 & \underline{0.68} \\
Claude-3.5 (0-shot) & 17.40 & 95.00 & 0.49 & \textbf{\underline{100.00}} & 3.22 & 0.52 & 0.47 & -0.42 & 0.66 \\
GPT-4o (0-shot) & 19.40 & 85.40 & \textbf{\underline{0.61}} & \textbf{\underline{100.00}} & 3.25 & 0.35 & 0.53 & -0.83 & 0.61 \\
Mistral (5-shot) & 59.60 & 98.40 & 0.54 & 98.66 & 3.07 & 0.57 & 0.39 & -0.38 & 0.66 \\
Llama (5-shot) & 34.80 & 95.20 & 0.57 & 97.13 & 3.18 & 0.53 & 0.48 & -0.41 & 0.65 \\
Claude-3.5 (5-shot) & 50.60 & 93.60 & 0.49 & 99.21 & 3.01 & \underline{0.71} & 0.41 & \underline{0.13} & \underline{0.68} \\
GPT-4o (1-shot) & 21.40 & 86.20 & 0.60 & \textbf{\underline{100.00}} & 3.17 & 0.48 & 0.47 & -0.52 & 0.65 \\

\rowcolor{lightgray}
\multicolumn{10}{c}{\textbf{Foundational LLMs for Chemistry}} 
\\
\ChemLLM & 6.20 & 31.60 & 0.31 & \textbf{\underline{100.00}} & \underline{2.82} & 0.61 & 0.40 & -0.28 & 0.63 \\

\cellcolor{yellow!20}\LlaSMolM & \underline{\cellcolor{yellow!20}76.40} & \textbf{\underline{\cellcolor{yellow!20}100.00}} & \cellcolor{yellow!20}0.55 & \cellcolor{yellow!20}99.74 & \cellcolor{yellow!20}3.07 & \cellcolor{yellow!20}0.53 & \cellcolor{yellow!20}0.42 & \cellcolor{yellow!20}-0.48 & \cellcolor{yellow!20}0.67 \\

\rowcolor{lightgray}
\multicolumn{10}{c}{\textbf{Generalist LLMs}} 
\\

\cellcolor{blue!10}\mollmSixGenM & \textbf{\cellcolor{blue!10}95.20} & \cellcolor{blue!10}99.80 & \cellcolor{blue!10}0.53 & \cellcolor{blue!10}99.79 & \cellcolor{blue!10}2.97 & \cellcolor{blue!10}0.85 & \cellcolor{blue!10}0.37 & \cellcolor{blue!10}0.46 & \textbf{\cellcolor{blue!10}0.70} \\

\mollmSixGenL  & 93.60 & \textbf{100.00} & 0.48 & 99.79 & \textbf{2.80} & \textbf{0.91} & \textbf{0.35} & \textbf{0.68} & \textbf{0.70} \\

\hline
\ImpG & 24.6 & -0.2 & -3.6 & 0.1 & 3.3 & 60.4 & -11.9 & 195.8 & 4.5 \\

\bottomrule
\end{tabular}

\begin{tablenotes}[normal,flushleft]
\footnotesize
\item $^\uparrow$ and $^\downarrow$ denote whether a higher or lower value of the metric is desirable, respectively.
For each task,
the best baseline performance is \underline{underlined} 
and the best overall performance is in \textbf{bold}
for each metric.
{\ImpG} denotes the percentage improvement from the
\colorbox{blue!10}{best generalist LLM} over the 
\colorbox{yellow!20}{best baseline},
where the best models are selected based on {\SR} for each task.
\end{tablenotes}

\end{threeparttable}
\end{small}
\end{table*}
\begin{table*}[h!]
\centering
\caption{Overall Performance on \BDMQ}
\setlength{\tabcolsep}{0pt}%
\label{tbl:bdmq_ood}
\begin{small}
\begin{threeparttable}

\begin{tabular}{
    @{\hspace{9pt}}l@{\hspace{9pt}}
    @{\hspace{9pt}}r@{\hspace{9pt}}
    @{\hspace{9pt}}r@{\hspace{9pt}}
    @{\hspace{9pt}}r@{\hspace{9pt}}
    @{\hspace{9pt}}r@{\hspace{9pt}}
    @{\hspace{9pt}}r@{\hspace{9pt}}
    @{\hspace{9pt}}r@{\hspace{9pt}}
    @{\hspace{5pt}}r@{\hspace{5pt}}
    @{\hspace{5pt}}r@{\hspace{5pt}}
    @{\hspace{5pt}}r@{\hspace{5pt}}
    @{\hspace{5pt}}r@{\hspace{5pt}}
}
\toprule
\multirow{2}{*}{Model} 
& \multirow{2}{*}{\SR$^{\uparrow}$}
& \multirow{2}{*}{\Val$^{\uparrow}$} 
& \multirow{2}{*}{\Sim$^{\uparrow}$} 
& \multirow{2}{*}{\Nov$^{\uparrow}$}
& \multirow{2}{*}{\SAS$^{\downarrow}$}
& \multirow{2}{*}{\RI$^{\uparrow}$}
& \multicolumn{3}{c}{\APS}
\\
\cmidrule(){8-11} 
& & & & & & & BBBP$^\uparrow$ & DRD2$^\uparrow$ & Mutag$^\downarrow$ & QED$^\uparrow$ 
\\
\midrule

\rowcolor{lightgray}
\multicolumn{11}{c}{\textbf{General-purpose LLMs}} 
\\
Mistral (0-shot) & 1.20 & 82.00 & 0.68 & \textbf{\underline{100.00}} & 3.13 & 0.37 & 0.50 & 0.02 & 0.42 & 0.24 \\
Llama (0-shot) & 1.20 & 87.20 & \textbf{\underline{0.76}} & \textbf{\underline{100.00}} & 3.89 & 0.30 & 0.46 & 0.02 & 0.45 & 0.18 \\
Claude-3.5 (0-shot) & 15.00 & 97.00 & 0.57 & \textbf{\underline{100.00}} & 2.84 & 0.87 & 0.58 & 0.06 & 0.32 & 0.34 \\
GPT-4o (0-shot) & 1.60 & 89.40 & 0.67 & \textbf{\underline{100.00}} & 3.08 & 0.18 & 0.45 & 0.02 & 0.36 & 0.32 \\
Mistral (5-shot) & 20.40 & 94.00 & 0.59 & 94.12 & 2.98 & 1.65 & 0.61 & 0.11 & 0.30 & 0.37 \\
Llama (5-shot) & 16.80 & 95.80 & 0.39 & 54.76 & 3.26 & \underline{3.22} & \underline{0.71} & \textbf{\underline{0.29}} & \textbf{\underline{0.24}} & 0.43 \\
Claude-3.5 (5-shot) & \underline{\cellcolor{yellow!20}30.40} & \cellcolor{yellow!20}95.60 & \cellcolor{yellow!20}0.49 & \textbf{\underline{\cellcolor{yellow!20}100.00}} & \textbf{\underline{\cellcolor{yellow!20}2.71}} & \cellcolor{yellow!20}2.32 & \cellcolor{yellow!20}0.68 & \cellcolor{yellow!20}0.12 & \cellcolor{yellow!20}0.31 & \textbf{\underline{\cellcolor{yellow!20}0.45}} \\
GPT-4o (1-shot) & 6.20 & 90.80 & 0.70 & \textbf{\underline{100.00}} & 2.86 & 0.75 & 0.51 & 0.04 & 0.34 & 0.29 \\

\rowcolor{lightgray}
\multicolumn{11}{c}{\textbf{Foundational LLMs for Chemistry}} 
\\
\ChemLLM & 0.00 & 13.20 & n/a & n/a & n/a & n/a & n/a & n/a & n/a & n/a \\
\LlaSMolM & 28.20 & \underline{98.20} & 0.66 & \textbf{\underline{100.00}} & 2.89 & 0.52 & 0.51 & 0.03 & 0.37 & 0.31 \\

\rowcolor{lightgray}
\multicolumn{11}{c}{\textbf{Generalist LLMs}} 
\\

\cellcolor{blue!10}\mollmSixGenM & \textbf{\cellcolor{blue!10}79.00} & \cellcolor{blue!10}99.00 & \cellcolor{blue!10}0.56 & \textbf{\cellcolor{blue!10}100.00} & \cellcolor{blue!10}2.84 & \cellcolor{blue!10}3.10 & \textbf{\cellcolor{blue!10}0.73} & \cellcolor{blue!10}0.16 & \cellcolor{blue!10}0.33 & \cellcolor{blue!10}0.42 \\

\mollmSixGenL & 74.20 & \textbf{99.60} & 0.55 & \textbf{100.00} & 2.74 & \textbf{3.25} & \textbf{0.73} & 0.16 & 0.33 & \textbf{0.45} \\
\hline
\ImpG & 159.9 & 3.6 & 14.3 & 0.0 & -4.8 & 33.6 & 7.4 & 33.3 & 6.5 & -6.7 \\

\bottomrule
\end{tabular}

\begin{tablenotes}[normal,flushleft]
\footnotesize
\item $^\uparrow$ and $^\downarrow$ denote whether a higher or lower value of the metric is desirable, respectively.
For each task,
the best baseline performance is \underline{underlined} 
and the best overall performance is in \textbf{bold}
for each metric.
{\ImpG} denotes the percentage improvement from the \colorbox{blue!10}{best generalist LLM} over the 
\colorbox{yellow!20}{best baseline},
where the best models are selected based on {\SR} for each task.
`n/a' indicates undefined metrics due to no optimized cases (i.e., 0\% \SR).
\end{tablenotes}

\end{threeparttable}
\end{small}
\end{table*}
\begin{table*}[h!]
\centering
\caption{Overall Performance on \BHMQ}
\setlength{\tabcolsep}{0pt}%
\label{tbl:bhmq_ood}
\begin{small}
\begin{threeparttable}

\begin{tabular}{
    @{\hspace{9pt}}l@{\hspace{9pt}}
    @{\hspace{9pt}}r@{\hspace{9pt}}
    @{\hspace{9pt}}r@{\hspace{9pt}}
    @{\hspace{9pt}}r@{\hspace{9pt}}
    @{\hspace{9pt}}r@{\hspace{9pt}}
    @{\hspace{9pt}}r@{\hspace{9pt}}
    @{\hspace{9pt}}r@{\hspace{9pt}}
    @{\hspace{5pt}}r@{\hspace{5pt}}
    @{\hspace{5pt}}r@{\hspace{5pt}}
    @{\hspace{5pt}}r@{\hspace{5pt}}
    @{\hspace{5pt}}r@{\hspace{5pt}}
}
\toprule
\multirow{2}{*}{Model} 
& \multirow{2}{*}{\SR$^{\uparrow}$}
& \multirow{2}{*}{\Val$^{\uparrow}$} 
& \multirow{2}{*}{\Sim$^{\uparrow}$} 
& \multirow{2}{*}{\Nov$^{\uparrow}$}
& \multirow{2}{*}{\SAS$^{\downarrow}$}
& \multirow{2}{*}{\RI$^{\uparrow}$}
& \multicolumn{3}{c}{\APS}
\\
\cmidrule(){8-11} 
& & & & & & & BBBP$^\uparrow$ & HIA$^\uparrow$ & Mutag$^\downarrow$ & QED$^\uparrow$ 
\\
\midrule

\rowcolor{lightgray}
\multicolumn{11}{c}{\textbf{General-purpose LLMs}} 
\\
Mistral (0-shot) & 12.71 & 76.27 & 0.73 & \textbf{\underline{100.00}} & 3.56 & 1.90 & 0.32 & 0.64 & 0.37 & 0.24 \\
Llama (0-shot) & 11.02 & 92.37 & 0.74 & \textbf{\underline{100.00}} & 4.39 & 0.68 & 0.28 & 0.63 & 0.45 & 0.20 \\
Claude-3.5 (0-shot) & 38.98 & 94.92 & 0.51 & \textbf{\underline{100.00}} & 2.93 & 2.35 & 0.49 & 0.85 & 0.36 & 0.52 \\
GPT-4o (0-shot) & 17.80 & 94.92 & 0.66 & \textbf{\underline{100.00}} & 3.70 & 0.65 & 0.32 & 0.67 & 0.37 & 0.43 \\
Mistral (5-shot) & 34.75 & 86.44 & 0.70 & \textbf{\underline{100.00}} & 3.36 & 1.31 & 0.42 & 0.70 & 0.39 & 0.40 \\
Llama (5-shot) & 36.44 & 92.37 & 0.67 & 97.67 & 3.78 & 1.13 & 0.37 & 0.64 & 0.39 & 0.34 \\
Claude-3.5 (5-shot) & 52.54 & 95.76 & 0.48 & \textbf{\underline{100.00}} & \textbf{\underline{2.78}} & \underline{2.52} & 0.50 & \textbf{\underline{0.92}} & 0.37 & \textbf{\underline{0.58}} \\
GPT-4o (1-shot) & 14.41 & 92.37 & \textbf{\underline{0.75}} & \textbf{\underline{100.00}} & 3.65 & 0.67 & 0.28 & 0.60 & 0.39 & 0.34 \\

\rowcolor{lightgray}
\multicolumn{11}{c}{\textbf{Foundational LLMs for Chemistry}} 
\\
\ChemLLM  & 1.69 & 12.71 & 0.53 & \textbf{\underline{100.00}} & 3.10 & 0.74 & \underline{0.54} & 0.54 & \textbf{\underline{0.22}} & 0.32 \\
\cellcolor{yellow!20}\LlaSMolM & \underline{\cellcolor{yellow!20}53.39} & \underline{\cellcolor{yellow!20}96.61} & \cellcolor{yellow!20}0.62 & \textbf{\underline{\cellcolor{yellow!20}100.00}} & \cellcolor{yellow!20}3.16 & \cellcolor{yellow!20}1.14 & \cellcolor{yellow!20}0.37 & \cellcolor{yellow!20}0.69 & \cellcolor{yellow!20}0.41 & \cellcolor{yellow!20}0.45 \\

\rowcolor{lightgray}
\multicolumn{11}{c}{\textbf{Generalist LLMs}} 
\\

\mollmSixGenM & 86.44 & 98.31 & 0.54 & \textbf{100.00} & 3.19 & 2.58 & 0.60 & 0.84 & 0.37 & 0.51 \\

\cellcolor{blue!10}\mollmSixGenL & \textbf{\cellcolor{blue!10}93.22} & \textbf{\cellcolor{blue!10}100.00} & \cellcolor{blue!10}0.49 & \cellcolor{blue!10}99.09 & \cellcolor{blue!10}3.02 & \textbf{\cellcolor{blue!10}3.57} & \textbf{\cellcolor{blue!10}0.64} & \textbf{\cellcolor{blue!10}0.92} & \cellcolor{blue!10}0.34 & \textbf{\cellcolor{blue!10}0.58} \\
\hline
\ImpG & 74.6 & 3.5 & -21.0 & -0.9 & 4.4 & 213.2 & 73.0 & 33.3 & -17.1 & 28.9 \\

\bottomrule
\end{tabular}

\begin{tablenotes}[normal,flushleft]
\footnotesize
\item $^\uparrow$ and $^\downarrow$ denote whether a higher or lower value of the metric is desirable, respectively.
For each task,
the best baseline performance is \underline{underlined} 
and the best overall performance is in \textbf{bold}
for each metric.
{\ImpG} denotes the percentage improvement from the \colorbox{blue!10}{best generalist LLM} over the 
\colorbox{yellow!20}{best baseline},
where the best models are selected based on {\SR} for each task.
\end{tablenotes}

\end{threeparttable}
\end{small}
\end{table*}
\begin{table*}[h!]
\centering
\caption{Overall Performance on \BMPQ}
\setlength{\tabcolsep}{0pt}%
\label{tbl:bmpq_ood}
\begin{small}
\begin{threeparttable}

\begin{tabular}{
    @{\hspace{9pt}}l@{\hspace{9pt}}
    @{\hspace{9pt}}r@{\hspace{9pt}}
    @{\hspace{9pt}}r@{\hspace{9pt}}
    @{\hspace{9pt}}r@{\hspace{9pt}}
    @{\hspace{9pt}}r@{\hspace{9pt}}
    @{\hspace{9pt}}r@{\hspace{9pt}}
    @{\hspace{9pt}}r@{\hspace{9pt}}
    @{\hspace{5pt}}r@{\hspace{5pt}}
    @{\hspace{5pt}}r@{\hspace{5pt}}
    @{\hspace{5pt}}r@{\hspace{5pt}}
    @{\hspace{5pt}}r@{\hspace{5pt}}
}
\toprule
\multirow{2}{*}{Model} 
& \multirow{2}{*}{\SR$^{\uparrow}$}
& \multirow{2}{*}{\Val$^{\uparrow}$} 
& \multirow{2}{*}{\Sim$^{\uparrow}$} 
& \multirow{2}{*}{\Nov$^{\uparrow}$}
& \multirow{2}{*}{\SAS$^{\downarrow}$}
& \multirow{2}{*}{\RI$^{\uparrow}$}
& \multicolumn{3}{c}{\APS}
\\
\cmidrule(){8-11} 
& & & & & & & BBBP$^\uparrow$ & Mutag$^\downarrow$ & plogP$^\uparrow$ & QED$^\uparrow$ 
\\
\midrule

\rowcolor{lightgray}
\multicolumn{11}{c}{\textbf{General-purpose LLMs}} 
\\
Mistral (0-shot) & 12.57 & 79.06 & 0.61 & \textbf{\underline{100.00}} & 3.18 & 0.54 & 0.60 & 0.47 & -1.07 & 0.45 \\
Llama (0-shot) & 16.75 & 93.72 & 0.51 & \textbf{\underline{100.00}} & 3.16 & 0.57 & 0.48 & 0.40 & -1.32 & 0.46 \\
Claude-3.5 (0-shot) & 44.50 & 94.76 & 0.55 & \textbf{\underline{100.00}} & 3.34 & 0.85 & 0.59 & 0.44 & -0.55 & 0.51 \\
GPT-4o (0-shot) & 25.13 & 93.19 & 0.67 & \textbf{\underline{100.00}} & 3.52 & 0.35 & 0.48 & 0.54 & -1.66 & 0.50 \\
Mistral (5-shot) & 49.21 & 95.81 & 0.62 & 96.81 & 3.30 & 0.73 & 0.63 & 0.46 & -0.93 & 0.55 \\
Llama (5-shot) & 31.94 & 96.34 & 0.66 & 96.72 & 3.40 & 0.60 & 0.60 & 0.48 & -1.02 & 0.49 \\
Claude-3.5 (5-shot) & 52.36 & 92.15 & 0.46 & \textbf{\underline{100.00}} & 2.97 & \underline{1.08} & 0.69 & 0.37 & \textbf{\underline{0.43}} & \underline{0.61} \\
GPT-4o (1-shot) & 24.08 & 90.58 & \textbf{\underline{0.68}} & \textbf{\underline{100.00}} & 3.50 & 0.45 & 0.49 & 0.52 & -1.41 & 0.47 \\

\rowcolor{lightgray}
\multicolumn{11}{c}{\textbf{Foundational LLMs for Chemistry}} 
\\
\ChemLLM & 5.24 & 19.37 & 0.24 & \textbf{\underline{100.00}} & \underline{2.93} & 0.72 & \textbf{\underline{0.80}} & \textbf{\underline{0.32}} & -0.83 & 0.57 \\

\cellcolor{yellow!20}\LlaSMolM & \underline{\cellcolor{yellow!20}64.92} & \underline{\cellcolor{yellow!20}98.95} & \cellcolor{yellow!20}0.58 & \cellcolor{yellow!20}99.19 & \cellcolor{yellow!20}3.14 & \cellcolor{yellow!20}0.57 & \cellcolor{yellow!20}0.56 & \cellcolor{yellow!20}0.49 & \cellcolor{yellow!20}-0.91 & \cellcolor{yellow!20}0.57 \\

\rowcolor{lightgray}
\multicolumn{11}{c}{\textbf{Generalist LLMs}} 
\\

\mollmSixGenM & 91.10 & \textbf{100.00} & 0.53 & 99.43 & 3.04 & 1.06 & 0.74 & 0.40 & -0.09 & 0.62 \\

\cellcolor{blue!10}\mollmSixGenL & \textbf{\cellcolor{blue!10}95.29} & \cellcolor{blue!10}98.95 & \cellcolor{blue!10}0.49 & \cellcolor{blue!10}99.45 & \textbf{\cellcolor{blue!10}2.87} & \textbf{\cellcolor{blue!10}1.20} & \cellcolor{blue!10}0.76 & \cellcolor{blue!10}0.37 & \cellcolor{blue!10}0.29 & \textbf{\cellcolor{blue!10}0.65} \\
\hline
\ImpG & 46.8 & 0.0 & -15.5 & 0.3 & 8.6 & 110.5 & 35.7 & -24.5 & 131.9 & 14.0 \\

\bottomrule
\end{tabular}

\begin{tablenotes}[normal,flushleft]
\footnotesize
\item $^\uparrow$ and $^\downarrow$ denote whether a higher or lower value of the metric is desirable, respectively.
For each task,
the best baseline performance is \underline{underlined} 
and the best overall performance is in \textbf{bold}
for each metric.
{\ImpG} denotes the percentage improvement from the \colorbox{blue!10}{best generalist LLM} over the 
\colorbox{yellow!20}{best baseline},
where the best models are selected based on {\SR} for each task.
\end{tablenotes}

\end{threeparttable}
\end{small}
\end{table*}
\begin{table*}[h!]
\centering
\caption{Overall Performance on \HMPQ}
\setlength{\tabcolsep}{0pt}%
\label{tbl:hmpq_ood}
\begin{small}
\begin{threeparttable}

\begin{tabular}{
    @{\hspace{9pt}}l@{\hspace{9pt}}
    @{\hspace{9pt}}r@{\hspace{9pt}}
    @{\hspace{9pt}}r@{\hspace{9pt}}
    @{\hspace{9pt}}r@{\hspace{9pt}}
    @{\hspace{9pt}}r@{\hspace{9pt}}
    @{\hspace{9pt}}r@{\hspace{9pt}}
    @{\hspace{9pt}}r@{\hspace{9pt}}
    @{\hspace{5pt}}r@{\hspace{5pt}}
    @{\hspace{5pt}}r@{\hspace{5pt}}
    @{\hspace{5pt}}r@{\hspace{5pt}}
    @{\hspace{5pt}}r@{\hspace{5pt}}
}
\toprule
\multirow{2}{*}{Model} 
& \multirow{2}{*}{\SR$^{\uparrow}$}
& \multirow{2}{*}{\Val$^{\uparrow}$} 
& \multirow{2}{*}{\Sim$^{\uparrow}$} 
& \multirow{2}{*}{\Nov$^{\uparrow}$}
& \multirow{2}{*}{\SAS$^{\downarrow}$}
& \multirow{2}{*}{\RI$^{\uparrow}$}
& \multicolumn{3}{c}{\APS}
\\
\cmidrule(){8-11} 
& & & & & & & HIA$^\uparrow$ & Mutag$^\downarrow$ & plogP$^\uparrow$ & QED$^\uparrow$ 
\\
\midrule

\rowcolor{lightgray}
\multicolumn{11}{c}{\textbf{General-purpose LLMs}} 
\\
Mistral (0-shot) & 21.88 & 84.38 & \textbf{\underline{0.72}} & \textbf{\underline{100.00}} & 3.62 & 0.72 & 0.66 & 0.50 & -1.41 & 0.36 \\
Llama (0-shot) & 15.62 & 91.67 & 0.47 & \textbf{\underline{100.00}} & 3.14 & 0.60 & 0.78 & \textbf{\underline{0.33}} & -1.25 & 0.50 \\
Claude-3.5 (0-shot) & 38.54 & 96.88 & 0.54 & \textbf{\underline{100.00}} & 3.42 & 1.01 & 0.75 & 0.42 & -0.91 & 0.45 \\
GPT-4o (0-shot) & 20.83 & 92.71 & 0.67 & \textbf{\underline{100.00}} & 3.64 & 0.37 & 0.70 & 0.50 & -2.29 & 0.42 \\
Mistral (5-shot) & 46.88 & 89.58 & 0.66 & 97.78 & 3.68 & 0.91 & 0.73 & 0.49 & -1.72 & 0.42 \\
Llama (5-shot) & 33.33 & 93.75 & 0.68 & \textbf{\underline{100.00}} & 3.66 & 0.61 & 0.71 & 0.46 & -1.73 & 0.39 \\
Claude-3.5 (5-shot) & \underline{\cellcolor{yellow!20}65.62} & \cellcolor{yellow!20}96.88 & \cellcolor{yellow!20}0.48 & \textbf{\underline{\cellcolor{yellow!20}100.00}} & \cellcolor{yellow!20}3.12 & \underline{\cellcolor{yellow!20}1.32} & \underline{\cellcolor{yellow!20}0.87} & \cellcolor{yellow!20}0.41 & \underline{\cellcolor{yellow!20}-0.43} & \underline{\cellcolor{yellow!20}0.56} \\
GPT-4o (1-shot) & 25.00 & 92.71 & 0.70 & \textbf{\underline{100.00}} & 3.74 & 0.54 & 0.62 & 0.46 & -2.00 & 0.40 \\

\rowcolor{lightgray}
\multicolumn{11}{c}{\textbf{Foundational LLMs for Chemistry}} 
\\
\ChemLLM & 3.12 & 22.92 & 0.58 & \textbf{\underline{100.00}} & \textbf{\underline{3.02}} & 0.51 & 0.54 & 0.36 & -1.90 & 0.54 \\
\LlaSMolM & 53.12 & \underline{98.96} & 0.62 & \textbf{\underline{100.00}} & 3.37 & 0.70 & 0.74 & 0.50 & -1.89 & 0.48 \\

\rowcolor{lightgray}
\multicolumn{11}{c}{\textbf{Generalist LLMs}} 
\\

\mollmSixGenM & 91.67 & \textbf{100.00} & 0.55 & \textbf{100.00} & 3.34 & 1.42 & 0.91 & 0.41 & -0.70 & 0.56 \\

\cellcolor{blue!10}\mollmSixGenL & \textbf{\cellcolor{blue!10}97.92} & \textbf{\cellcolor{blue!10}100.00} & \cellcolor{blue!10}0.46 & \cellcolor{blue!10}98.94 & \cellcolor{blue!10}3.06 & \textbf{\cellcolor{blue!10}1.76} & \textbf{\cellcolor{blue!10}0.94} & \cellcolor{blue!10}0.39 & \textbf{\cellcolor{blue!10}-0.30} & \textbf{\cellcolor{blue!10}0.64} \\
\hline
\ImpG & 49.2 & 3.2 & -4.2 & -1.1 & 1.9 & 33.3 & 8.0 & -4.9 & 30.2 & 14.3 \\

\bottomrule
\end{tabular}

\begin{tablenotes}[normal,flushleft]
\footnotesize
\item $^\uparrow$ and $^\downarrow$ denote whether a higher or lower value of the metric is desirable, respectively.
For each task,
the best baseline performance is \underline{underlined} 
and the best overall performance is in \textbf{bold}
for each metric.
{\ImpG} denotes the percentage improvement from the \colorbox{blue!10}{best generalist LLM} over the 
\colorbox{yellow!20}{best baseline},
where the best models are selected based on {\SR} for each task.
\end{tablenotes}

\end{threeparttable}
\end{small}
\end{table*}

\subsection{Generalizability to Unseen Instructions}
\label{sec:app:results:uninst}

Tables~\ref{tbl:bdp_uninst}, \ref{tbl:bdq_uninst}, \ref{tbl:bpq_uninst}, \ref{tbl:dpq_uninst} and \ref{tbl:bdpq_uninst} present the performance comparison of {\mollm}s
with baselines on all 5 IND tasks when prompted with unseen instructions and unseen property names.
than those used during instruction-tuning. 
This evaluation is meaningful as it mimics real-world scenarios where users may describe optimization tasks using varying terminologies, requiring models to understand the underlying semantics of the task rather than relying on exact token matching.

\begin{table*}[h!]
\centering
\caption{Performance on Unseen Instructions for \BDP}
\label{tbl:bdp_uninst}
\begin{small}
\begin{threeparttable}

\begin{tabular}{
    @{\hspace{3pt}}l@{\hspace{3pt}}
    @{\hspace{3pt}}l@{\hspace{6pt}}
    @{\hspace{6pt}}r@{\hspace{6pt}}
    @{\hspace{6pt}}r@{\hspace{6pt}}
    @{\hspace{6pt}}r@{\hspace{6pt}}
    @{\hspace{6pt}}r@{\hspace{6pt}}
    @{\hspace{6pt}}r@{\hspace{6pt}}
    @{\hspace{6pt}}r@{\hspace{6pt}}
    @{\hspace{3pt}}r@{\hspace{3pt}}
    @{\hspace{3pt}}r@{\hspace{3pt}}
    @{\hspace{3pt}}r@{\hspace{3pt}}
}
\toprule
\multirow{2}{*}{Model}
& \multirow{2}{*}{Instr}
& \multirow{2}{*}{\SR$^{\uparrow}$} 
& \multirow{2}{*}{\Val$^{\uparrow}$} 
& \multirow{2}{*}{\Sim$^{\uparrow}$} 
& \multirow{2}{*}{\Nov$^{\uparrow}$}
& \multirow{2}{*}{\SAS$^{\downarrow}$}
& \multirow{2}{*}{\RI$^{\uparrow}$}
& \multicolumn{3}{c}{\APS}
\\
\cmidrule(){9-11} 
& & & & & & & & BBBP$^\uparrow$ & DRD2$^\uparrow$ & plogP$^\uparrow$
\\
\midrule

\multirow{2}{*}{\mollmTripleTaskM}
& seen & 84.80 & 96.80 & 0.47 & 100.00 & 3.06 & 4.30 & 0.77 & 0.23 & 0.46 \\
 & unseen & \textbf{89.60} & 97.60 & 0.45 & 99.55 & 3.05 & \textbf{5.11} & 0.79 & \textbf{0.28} & 0.47 \\
 \hline

\multirow{2}{*}{\mollmTripleTaskL}
& seen & 86.80 & 99.00 & \textbf{0.48} & 99.31 & 3.01 & 4.38 & 0.76 & 0.25 & 0.42 \\
 & unseen & 85.40 & 98.80 & 0.44 & 99.30 & 2.90 & \textbf{4.69} & 0.78 & \textbf{0.28} & \textbf{0.64} \\
 
 \hline


\multirow{2}{*}{\mollmSixGenM}
& seen & \textbf{83.00} & 98.80 & 0.57 & 99.76 & 3.17 & \textbf{3.60} & 0.72 & \textbf{0.17} & \textbf{0.12} \\
 & unseen & 75.80 & 98.60 & 0.59 & 99.74 & 3.24 & 3.15 & 0.70 & 0.14 & -0.12 \\
 \hline
 
 \multirow{2}{*}{\mollmSixGenL}
 & seen & \textbf{77.00} & 99.40 & 0.53 & 99.48 & 3.00 & \textbf{3.73} & 0.74 & \textbf{0.18} & \textbf{0.25} \\
 & unseen & 64.60 & 99.00 & 0.53 & 99.69 & 2.99 & 3.06 & 0.74 & 0.14 & 0.06 \\
\bottomrule
\end{tabular}

\begin{tablenotes}[normal,flushleft]
\footnotesize
\setlength\labelsep{0pt}
    \item ``seen" and ``unseen" indicate whether the {\mollm}s 
are evaluated with seen and unseen instructions, respectively.
$^\uparrow$ and $^\downarrow$ denote whether a higher or lower value of the metric is desirable, respectively.
The best-performing {\mollm} in each row block is in \textbf{bold} 
if the performance difference between the models evaluated with seen and unseen instructions exceeds 5\%. 
\par
\end{tablenotes}

\end{threeparttable}
\end{small}
\end{table*}
\begin{table*}[h!]
\centering
\caption{Performance on Unseen Instructions for \BDQ}
\label{tbl:bdq_uninst}
\begin{small}
\begin{threeparttable}

\begin{tabular}{
    @{\hspace{3pt}}l@{\hspace{3pt}}
    @{\hspace{3pt}}l@{\hspace{6pt}}
    @{\hspace{6pt}}r@{\hspace{6pt}}
    @{\hspace{6pt}}r@{\hspace{6pt}}
    @{\hspace{6pt}}r@{\hspace{6pt}}
    @{\hspace{6pt}}r@{\hspace{6pt}}
    @{\hspace{6pt}}r@{\hspace{6pt}}
    @{\hspace{6pt}}r@{\hspace{6pt}}
    @{\hspace{3pt}}r@{\hspace{3pt}}
    @{\hspace{3pt}}r@{\hspace{3pt}}
    @{\hspace{3pt}}r@{\hspace{3pt}}
}
\toprule
\multirow{2}{*}{Model}
& \multirow{2}{*}{Instr}
& \multirow{2}{*}{\SR$^{\uparrow}$} 
& \multirow{2}{*}{\Val$^{\uparrow}$} 
& \multirow{2}{*}{\Sim$^{\uparrow}$} 
& \multirow{2}{*}{\Nov$^{\uparrow}$}
& \multirow{2}{*}{\SAS$^{\downarrow}$}
& \multirow{2}{*}{\RI$^{\uparrow}$}
& \multicolumn{3}{c}{\APS}
\\
\cmidrule(){9-11} 
& & & & & & & & BBBP$^\uparrow$ & DRD2$^\uparrow$ & QED$^\uparrow$
\\
\midrule

\multirow{2}{*}{\mollmTripleTaskM}
& seen & 87.00 & 98.80 & \textbf{0.47} & 100.00 & 2.90 & 5.61 & 0.79 & 0.22 & 0.49 \\
 & unseen & 87.40 & 99.00 & 0.44 & 100.00 & 2.83 & \textbf{6.29} & 0.81 & \textbf{0.25} & 0.50 \\
 \hline

\multirow{2}{*}{\mollmTripleTaskL}
& seen & 90.00 & 99.40 & 0.46 & 99.78 & 2.93 & 5.66 & 0.81 & 0.23 & 0.51 \\
 & unseen & 90.40 & 99.80 & 0.46 & 99.56 & 2.83 & 5.68 & 0.81 & 0.23 & 0.51 \\
 
 \hline


\multirow{2}{*}{\mollmSixGenM}
& seen & \textbf{85.80} & 99.60 & 0.59 & 100.00 & 2.94 & \textbf{4.78} & 0.75 & \textbf{0.19} & 0.40 \\
 & unseen & 80.40 & 99.40 & 0.59 & 99.75 & 2.93 & 4.54 & 0.74 & 0.17 & 0.39 \\
 \hline
 
 \multirow{2}{*}{\mollmSixGenL}
  & seen & \textbf{79.60} & 99.80 & 0.56 & 100.00 & 2.87 & \textbf{5.05} & 0.76 & \textbf{0.20} & 0.42 \\
 & unseen & 73.40 & 99.80 & 0.57 & 100.00 & 2.85 & 4.56 & 0.75 & 0.19 & 0.41 \\
\bottomrule
\end{tabular}

\begin{tablenotes}[normal,flushleft]
\footnotesize
\setlength\labelsep{0pt}
    \item ``seen" and ``unseen" indicate whether the {\mollm}s 
are evaluated with seen and unseen instructions, respectively.
$^\uparrow$ and $^\downarrow$ denote whether a higher or lower value of the metric is desirable, respectively.
The best-performing {\mollm} in each row block is in \textbf{bold} 
if the performance difference between the models evaluated with seen and unseen instructions exceeds 5\%. 
\par
\end{tablenotes}

\end{threeparttable}
\end{small}
\end{table*}
\begin{table*}[h!]
\centering
\caption{Performance on Unseen Instructions for \BPQ}
\label{tbl:bpq_uninst}
\begin{small}
\begin{threeparttable}

\begin{tabular}{
    @{\hspace{3pt}}l@{\hspace{3pt}}
    @{\hspace{3pt}}l@{\hspace{6pt}}
    @{\hspace{6pt}}r@{\hspace{6pt}}
    @{\hspace{6pt}}r@{\hspace{6pt}}
    @{\hspace{6pt}}r@{\hspace{6pt}}
    @{\hspace{6pt}}r@{\hspace{6pt}}
    @{\hspace{6pt}}r@{\hspace{6pt}}
    @{\hspace{6pt}}r@{\hspace{6pt}}
    @{\hspace{3pt}}r@{\hspace{3pt}}
    @{\hspace{3pt}}r@{\hspace{3pt}}
    @{\hspace{3pt}}r@{\hspace{3pt}}
}
\toprule
\multirow{2}{*}{Model}
& \multirow{2}{*}{Instr}
& \multirow{2}{*}{\SR$^{\uparrow}$} 
& \multirow{2}{*}{\Val$^{\uparrow}$} 
& \multirow{2}{*}{\Sim$^{\uparrow}$} 
& \multirow{2}{*}{\Nov$^{\uparrow}$}
& \multirow{2}{*}{\SAS$^{\downarrow}$}
& \multirow{2}{*}{\RI$^{\uparrow}$}
& \multicolumn{3}{c}{\APS}
\\
\cmidrule(){9-11} 
& & & & & & & & BBBP$^\uparrow$ & plogP$^\uparrow$ & QED$^\uparrow$
\\
\midrule

\multirow{2}{*}{\mollmTripleTaskM}
& seen & 93.00 & 97.40 & 0.46 & 99.57 & 3.14 & 1.49 & 0.77 & \textbf{-0.16} & 0.69 \\
 & unseen & 93.00 & 97.80 & 0.45 & 98.71 & 3.13 & 1.48 & 0.78 & -0.13 & 0.69 \\
 \hline

\multirow{2}{*}{\mollmTripleTaskL}
& seen & 94.00 & 98.20 & 0.50 & 98.94 & 3.13 & 1.38 & 0.77 & \textbf{-0.22} & 0.67 \\
 & unseen & 93.80 & 98.60 & 0.49 & 98.72 & 3.07 & 1.42 & 0.77 & -0.11 & 0.68 \\
 
 \hline


\multirow{2}{*}{\mollmSixGenM}
& seen & 96.80 & 99.20 & 0.53 & 99.38 & 3.23 & 1.48 & 0.78 & -0.34 & 0.68 \\
 & unseen & 96.20 & 98.80 & 0.54 & 98.96 & 3.22 & 1.42 & 0.77 & \textbf{-0.46} & 0.67 \\
 \hline
 
 \multirow{2}{*}{\mollmSixGenL}
& seen & 95.00 & 98.60 & 0.47 & 99.58 & 2.98 & 1.66 & 0.81 & \textbf{0.16} & 0.71 \\
 & unseen & 95.60 & 98.40 & 0.47 & 99.58 & 2.98 & 1.66 & 0.81 & 0.10 & 0.71 \\
\bottomrule
\end{tabular}

\begin{tablenotes}[normal,flushleft]
\footnotesize
\setlength\labelsep{0pt}
    \item ``seen" and ``unseen" indicate whether the {\mollm}s 
are evaluated with seen and unseen instructions, respectively.
$^\uparrow$ and $^\downarrow$ denote whether a higher or lower value of the metric is desirable, respectively.
The best-performing {\mollm} in each row block is in \textbf{bold} 
if the performance difference between the models evaluated with seen and unseen instructions exceeds 5\%. 
\par
\end{tablenotes}

\end{threeparttable}
\end{small}
\end{table*}
\begin{table*}[h!]
\centering
\caption{Performance on Unseen Instructions for \DPQ}
\label{tbl:dpq_uninst}
\begin{small}
\begin{threeparttable}

\begin{tabular}{
    @{\hspace{3pt}}l@{\hspace{3pt}}
    @{\hspace{3pt}}l@{\hspace{6pt}}
    @{\hspace{6pt}}r@{\hspace{6pt}}
    @{\hspace{6pt}}r@{\hspace{6pt}}
    @{\hspace{6pt}}r@{\hspace{6pt}}
    @{\hspace{6pt}}r@{\hspace{6pt}}
    @{\hspace{6pt}}r@{\hspace{6pt}}
    @{\hspace{6pt}}r@{\hspace{6pt}}
    @{\hspace{3pt}}r@{\hspace{3pt}}
    @{\hspace{3pt}}r@{\hspace{3pt}}
    @{\hspace{3pt}}r@{\hspace{3pt}}
}
\toprule
\multirow{2}{*}{Model}
& \multirow{2}{*}{Instr}
& \multirow{2}{*}{\SR$^{\uparrow}$} 
& \multirow{2}{*}{\Val$^{\uparrow}$} 
& \multirow{2}{*}{\Sim$^{\uparrow}$} 
& \multirow{2}{*}{\Nov$^{\uparrow}$}
& \multirow{2}{*}{\SAS$^{\downarrow}$}
& \multirow{2}{*}{\RI$^{\uparrow}$}
& \multicolumn{3}{c}{\APS}
\\
\cmidrule(){9-11} 
& & & & & & & & DRD2$^\uparrow$ & plogP$^\uparrow$ & QED$^\uparrow$
\\
\midrule

\multirow{2}{*}{\mollmTripleTaskM}
& seen & 62.80 & 98.20 & \textbf{0.37} & 97.45 & 3.04 & 3.87 & 0.24 & \textbf{-0.08} & 0.64 \\
 & unseen & 64.20 & 98.40 & 0.35 & 98.44 & 2.90 & 3.95 & 0.25 & 0.39 & 0.65 \\
 \hline

\multirow{2}{*}{\mollmTripleTaskL}
& seen & 60.60 & 99.00 & \textbf{0.44} & 97.03 & 3.08 & 3.76 & 0.24 & \textbf{-0.12} & 0.59 \\
 & unseen & 63.60 & 98.60 & 0.39 & 95.91 & 2.94 & \textbf{4.36} & \textbf{0.28} & 0.10 & \textbf{0.62} \\
 
 \hline


\multirow{2}{*}{\mollmSixGenM}
& seen & \textbf{60.80} & 99.40 & 0.54 & 99.67 & 3.31 & \textbf{2.16} & \textbf{0.12} & -0.57 & 0.57 \\
 & unseen & 54.60 & 98.80 & 0.55 & 99.63 & 3.32 & 1.99 & 0.11 & \textbf{-0.82} & 0.55 \\
 \hline
 
 \multirow{2}{*}{\mollmSixGenL}
& seen & \textbf{57.00} & 99.00 & 0.49 & 99.65 & 3.14 & \textbf{2.50} & \textbf{0.14} & -0.36 & 0.58 \\
 & unseen & 53.60 & 99.60 & 0.50 & 100.00 & 3.15 & 2.15 & 0.12 & \textbf{-0.39} & 0.58 \\
\bottomrule
\end{tabular}

\begin{tablenotes}[normal,flushleft]
\footnotesize
\setlength\labelsep{0pt}
    \item ``seen" and ``unseen" indicate whether the {\mollm}s 
are evaluated with seen and unseen instructions, respectively.
$^\uparrow$ and $^\downarrow$ denote whether a higher or lower value of the metric is desirable, respectively.
The best-performing {\mollm} in each row block is in \textbf{bold} 
if the performance difference between the models evaluated with seen and unseen instructions exceeds 5\%. 
\par
\end{tablenotes}

\end{threeparttable}
\end{small}
\end{table*}
\begin{table*}[h!]
\centering
\caption{Performance on Unseen Instructions for \BDPQ}
\label{tbl:bdpq_uninst}
\begin{small}
\begin{threeparttable}

\begin{tabular}{
    @{\hspace{3pt}}l@{\hspace{3pt}}
    @{\hspace{3pt}}l@{\hspace{6pt}}
    @{\hspace{6pt}}r@{\hspace{6pt}}
    @{\hspace{6pt}}r@{\hspace{6pt}}
    @{\hspace{6pt}}r@{\hspace{6pt}}
    @{\hspace{6pt}}r@{\hspace{6pt}}
    @{\hspace{6pt}}r@{\hspace{6pt}}
    @{\hspace{6pt}}r@{\hspace{6pt}}
    @{\hspace{3pt}}r@{\hspace{3pt}}
    @{\hspace{3pt}}r@{\hspace{3pt}}
    @{\hspace{3pt}}r@{\hspace{3pt}}
    @{\hspace{3pt}}r@{\hspace{3pt}}
}
\toprule
\multirow{2}{*}{Model}
& \multirow{2}{*}{Instr}
& \multirow{2}{*}{\SR$^{\uparrow}$} 
& \multirow{2}{*}{\Val$^{\uparrow}$} 
& \multirow{2}{*}{\Sim$^{\uparrow}$} 
& \multirow{2}{*}{\Nov$^{\uparrow}$}
& \multirow{2}{*}{\SAS$^{\downarrow}$}
& \multirow{2}{*}{\RI$^{\uparrow}$}
& \multicolumn{4}{c}{\APS}
\\
\cmidrule(){9-12} 
& & & & & & & & BBBP$^{\uparrow}$ & DRD2$^\uparrow$ & plogP$^\uparrow$ & QED$^\uparrow$
\\
\midrule

\multirow{2}{*}{\mollmTripleTaskM}
& seen & 30.00 & 93.00 & \textbf{0.48} & 95.33 & 3.02 & 3.44 & 0.65 & 0.17 & \textbf{-1.55} & 0.53 \\
 & unseen & \textbf{32.80} & 90.60 & 0.45 & 93.29 & 2.98 & \textbf{3.62} & 0.63 & \textbf{0.19} & -1.24 & 0.52 \\
 \hline

\multirow{2}{*}{\mollmTripleTaskL}
& seen & \textbf{28.00} & 94.00 & 0.66 & 98.57 & 3.57 & 1.02 & 0.56 & 0.05 & -2.68 & 0.42 \\
 & unseen & 24.20 & 93.40 & 0.64 & 97.52 & 3.47 & \textbf{1.29} & 0.58 & \textbf{0.06} & -2.67 & 0.44 \\
 
 \hline


\multirow{2}{*}{\mollmSixGenM}
& seen & \textbf{54.00} & 97.40 & 0.54 & 99.26 & 3.34 & \textbf{3.09} & 0.65 & \textbf{0.16} & -0.93 & 0.48 \\
 & unseen & 49.80 & 97.20 & \textbf{0.57} & 99.20 & 3.37 & 2.81 & 0.63 & 0.14 & -0.96 & 0.48 \\
 \hline
 
 \multirow{2}{*}{\mollmSixGenL}
 & seen & \textbf{52.20} & 97.20 & 0.49 & 99.23 & 3.17 & 3.48 & 0.69 & 0.16 & -0.65 & 0.53 \\
 & unseen & 46.40 & 97.20 & 0.48 & 99.14 & 3.09 & 3.52 & 0.68 & 0.16 & -0.68 & 0.55 \\
\bottomrule
\end{tabular}

\begin{tablenotes}[normal,flushleft]
\footnotesize
\setlength\labelsep{0pt}
    \item ``seen" and ``unseen" indicate whether the {\mollm}s 
are evaluated with seen and unseen instructions, respectively.
$^\uparrow$ and $^\downarrow$ denote whether a higher or lower value of the metric is desirable, respectively.
The best-performing {\mollm} in each row block is in \textbf{bold} 
if the performance difference between the models evaluated with seen and unseen instructions exceeds 5\%. 
\par
\end{tablenotes}

\end{threeparttable}
\end{small}
\end{table*}

\subsection{Additional Case Studies}
\label{sec:app:results:additional_cases}
\begin{figure}[p]
    \centering
    
    \begin{subfigure}[t]{\columnwidth}
        \centering
        \begin{minipage}{0.43\linewidth}
            \centering
            \tiny $M_x$
            \includegraphics[width=1.0\linewidth]{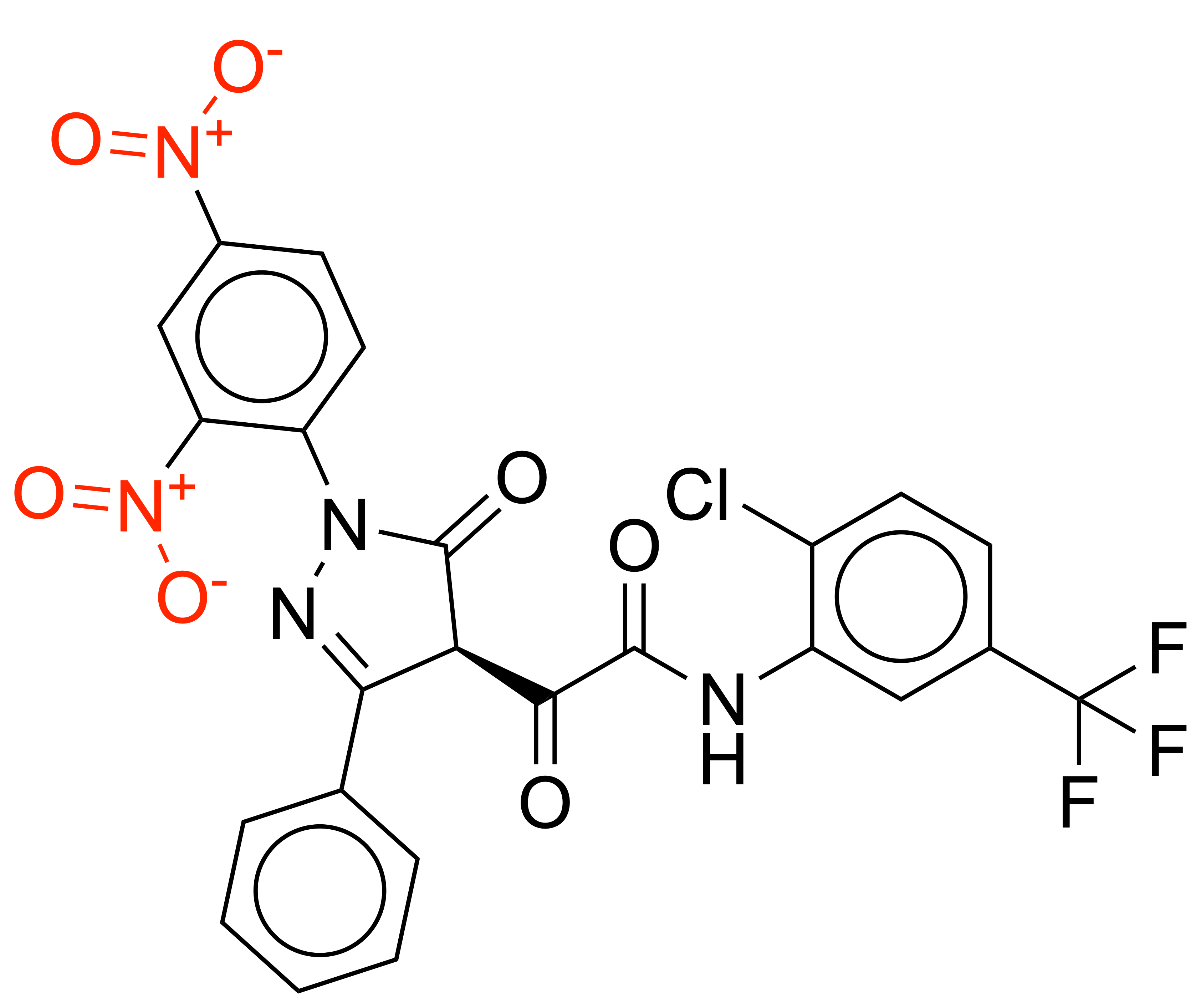}
            \par\vspace{2pt}
            \tiny BBBP=0.48, DRD2=0.00, \\
            \tiny QED=0.18
        \end{minipage}
        \hfill
        \begin{minipage}{0.1\linewidth}
            \centering \Large$\Rightarrow$\\
            \raggedright \tiny \mollm
            \raggedright \tiny \Sim=0.72
        \end{minipage}
        \hfill
        \begin{minipage}{0.44\linewidth}
            \centering
            \tiny $M_y$
            \includegraphics[width=1.0\linewidth]{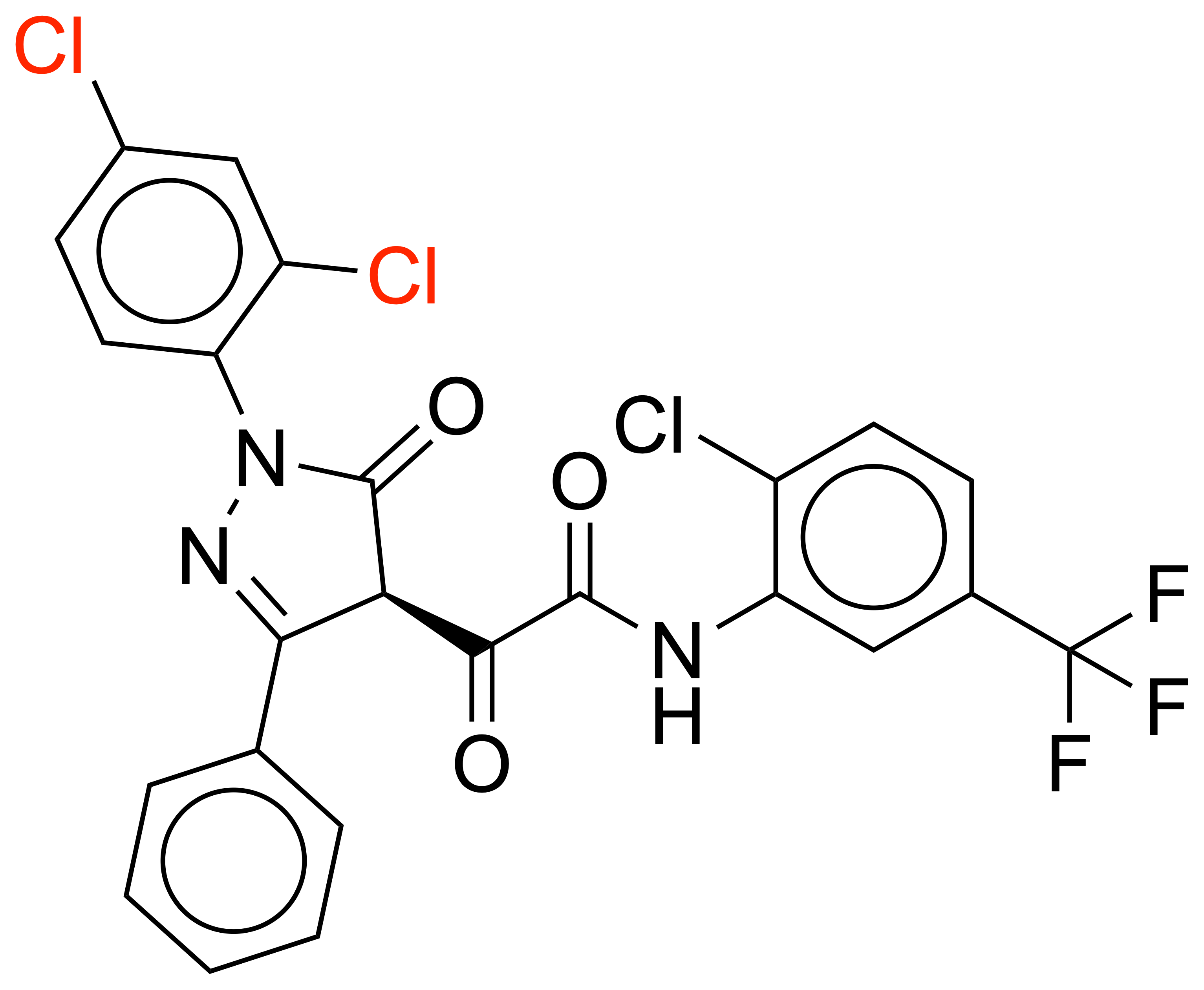}
            \par\vspace{2pt}
            \tiny BBBP=0.79 (+0.31),\\ 
            \tiny DRD2=0.01 (+0.01), QED=0.30 (+0.12)
        \end{minipage}
        \caption{{\mollmSixGenM}}
        \label{fig:BDQ_mollm_case_1}
    \end{subfigure}
    
    \vspace{1em}
    
    \begin{subfigure}[t]{\columnwidth}
        \centering
        \begin{minipage}{0.43\linewidth}
            \centering
            \tiny $M_x$
            \includegraphics[width=1.0\linewidth]{./figures/BDQ_case1_mollm_src.png}
            \par\vspace{2pt}
            \tiny BBBP=0.48, \\
            \tiny DRD2=0.00, QED=0.18
        \end{minipage}
        \hfill
        \begin{minipage}{0.1\linewidth}
            \centering \Large$\Rightarrow$\\
            \raggedright \tiny \LlaSMol
            \raggedright \tiny \Sim=0.62
        \end{minipage}
        \hfill
        \begin{minipage}{0.44\linewidth}
            \centering
            \tiny $M_y$\\
            \includegraphics[width=1.0\linewidth]{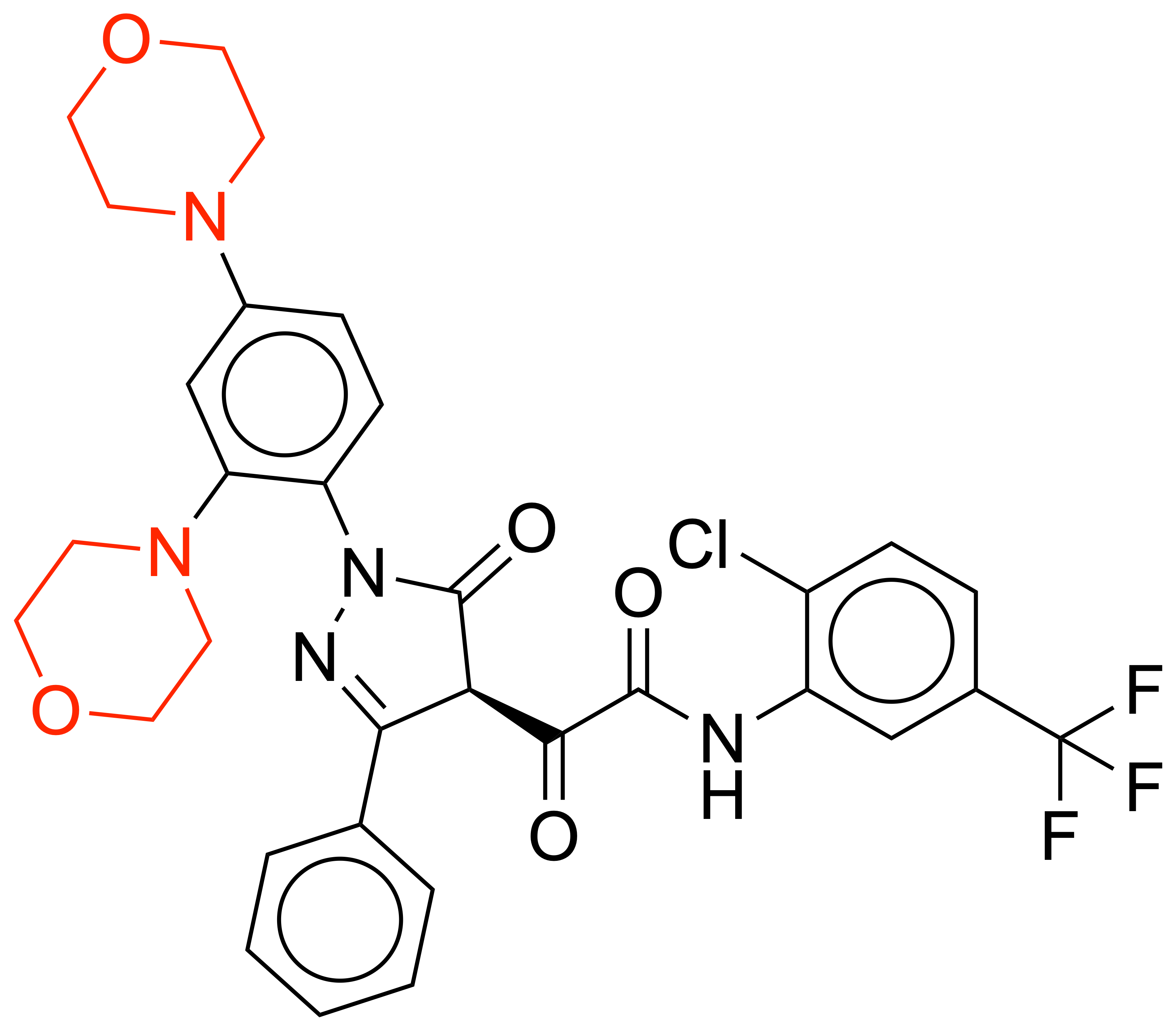}
            \par\vspace{1pt}
            \tiny BBBP=0.56 (+0.08), \\ 
            \tiny DRD2=0.03 (+0.03), QED=0.29 (+0.11)
        \end{minipage}
        \caption{{\LlaSMolM}}
        \label{fig:BDQ_llasmol_case_1}
    \end{subfigure}
    
    \caption{An optimization case on \BDQ. Modifications are highlighted in red.}
    \label{fig:BDQ_cases_1}
\end{figure}

In this section, we provide two additional cases from IND task \BDQ.
As shown in Figure~\ref{fig:BDQ_mollm_case_1}, \mollmSixGenM improves molecular properties by removing a nitro group (–NO$_{2}$) from the aromatic ring and replacing it with a chlorine atom (–Cl), while \LlaSMolM replaces the nitro group with two morpholine rings (highlighted structures).  
Removing the nitro group reduces polarity and eliminates a structural alert associated with toxicity and poor pharmacokinetics \cite{nepali2018nitro}, and the chlorine substitution enhances lipophilicity and promotes passive diffusion across the BBB \cite{plattard2021overview, rosa2024identifying}.  
As a result, \mollmSixGenM achieves a notable increase in BBBP (+0.31), despite the hit molecule already having a relatively good BBBP value of 0.48, making further optimization more challenging.  
Additionally, the nitro group removal improves drug-likeness \cite{nepali2018nitro, noriega2022diverse}, and the minimal fragment substitution preserves the molecular scaffold, resulting in higher structural similarity (0.72 vs. 0.62).  
In contrast, {\LlaSMolM}’s morpholine rings increase polarity and hydrogen bonding capacity \cite{lenci2021occurrence}, which shows limited BBBP gains (+0.08) (Figure~\ref{fig:BDQ_llasmol_case_1}).  

\begin{figure}[p]
    \centering
    
    \begin{subfigure}[t]{\columnwidth}
        \centering
        \begin{minipage}{0.43\linewidth}
            \centering
            \tiny $M_x$
            \includegraphics[width=1.0\linewidth]{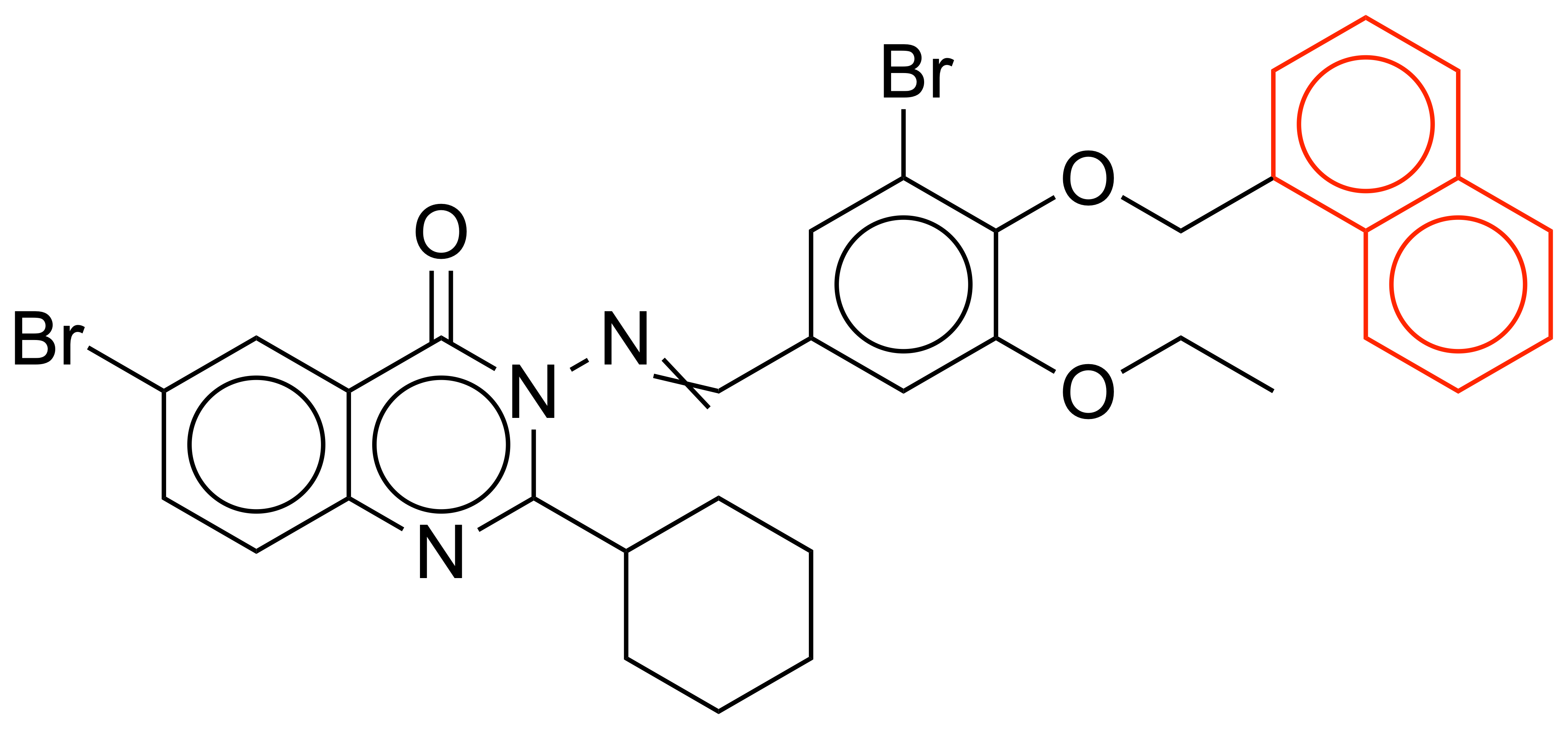}
            \par\vspace{2pt}
            \tiny BBBP=0.51, DRD2=0.06, \\
            \tiny QED=0.15
        \end{minipage}
        \hfill
        \begin{minipage}{0.1\linewidth}
            \centering \Large$\Rightarrow$\\
            \raggedright \tiny \mollm
            \raggedright \tiny \Sim=0.71
        \end{minipage}
        \hfill
        \begin{minipage}{0.44\linewidth}
            \centering
            \tiny $M_y$
            \includegraphics[width=1.0\linewidth]{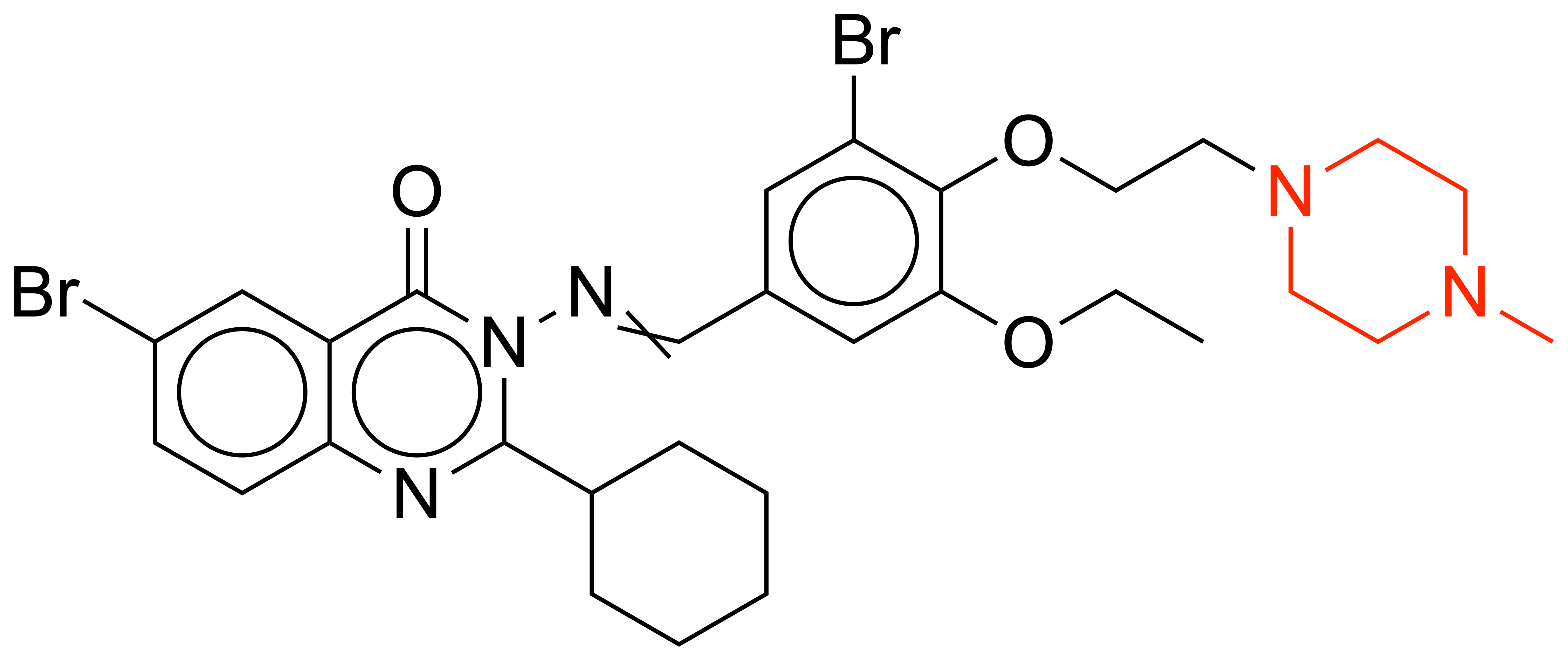}
            \par\vspace{2pt}
            \tiny BBBP=0.92 (+0.41),\\ 
            \tiny DRD2=0.47 (+0.41), QED=0.26 (+0.11)
        \end{minipage}
        \caption{{\mollmSixGenM}}
        \label{fig:BDQ_mollm_case_2}
    \end{subfigure}
    
    \vspace{1em}
    
    \begin{subfigure}[t]{\columnwidth}
        \centering
        \begin{minipage}{0.43\linewidth}
            \centering
            \tiny $M_x$
            \includegraphics[width=1.0\linewidth]{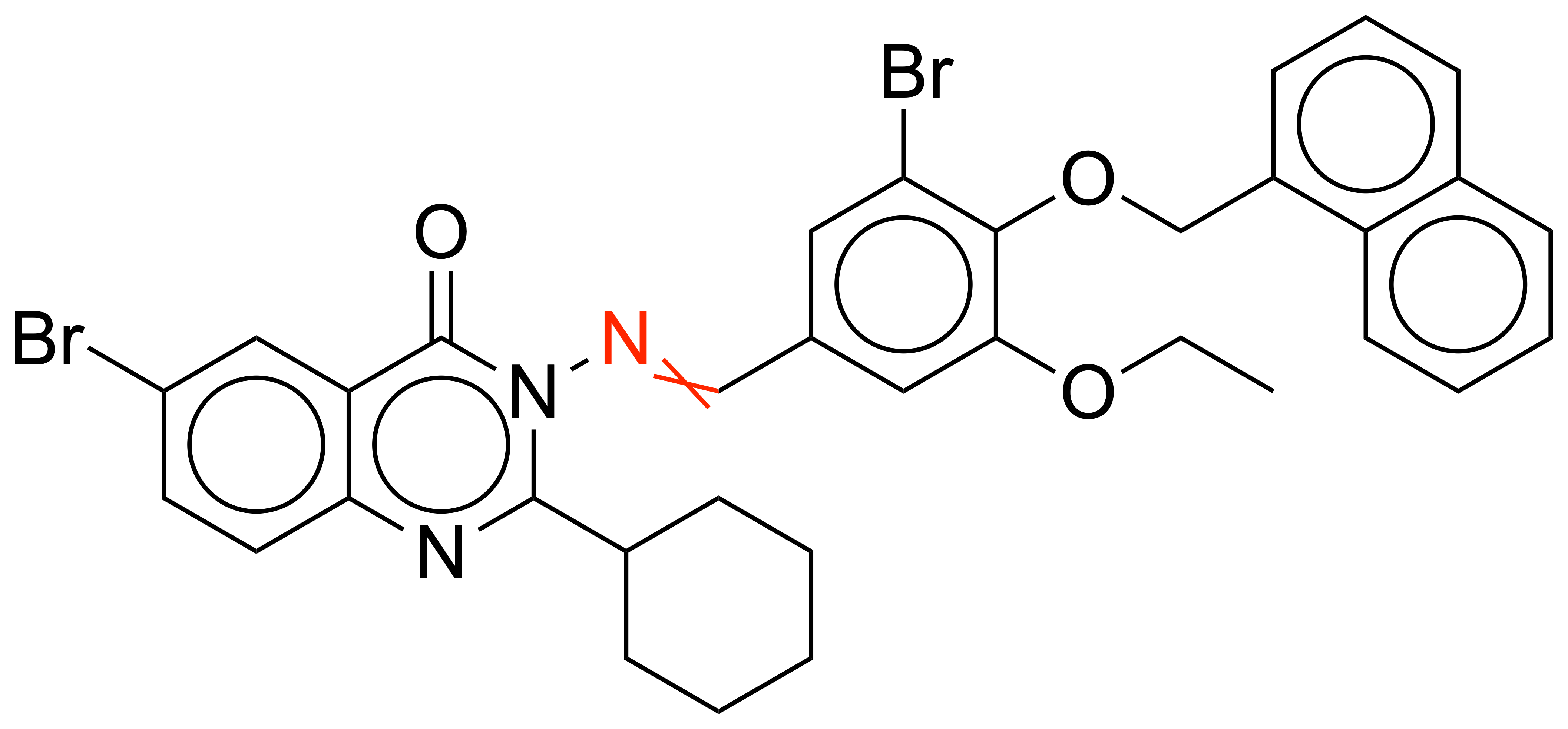}
            \par\vspace{2pt}
            \tiny BBBP=0.51, \\
            \tiny DRD2=0.06, QED=0.15
        \end{minipage}
        \hfill
        \begin{minipage}{0.1\linewidth}
            \centering \Large$\Rightarrow$\\
            \raggedright \tiny \LlaSMol
            \raggedright \tiny \Sim=0.69
        \end{minipage}
        \hfill
        \begin{minipage}{0.44\linewidth}
            \centering
            \tiny $M_y$\\
            \includegraphics[width=1.0\linewidth]{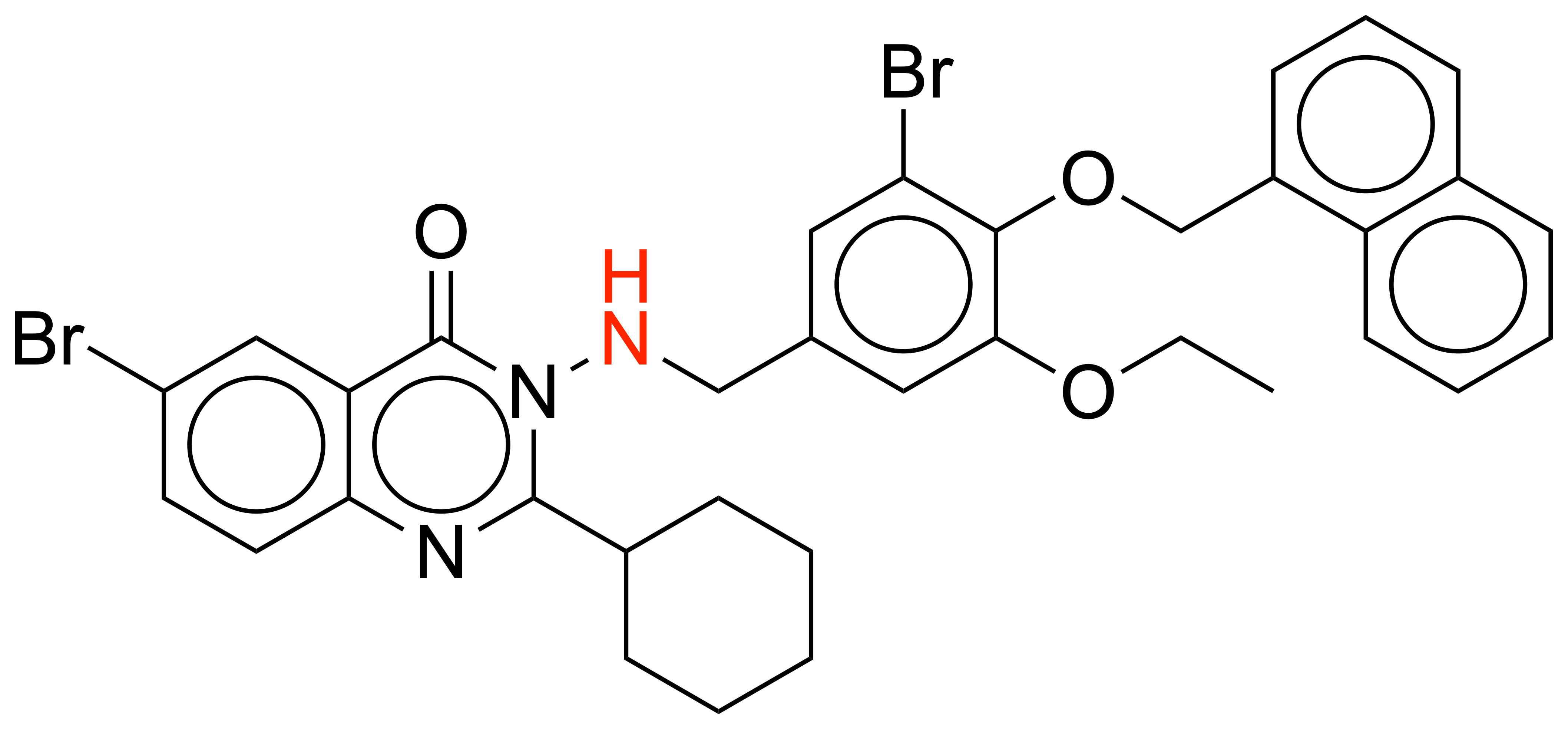}
            \par\vspace{1pt}
            \tiny BBBP=0.52 (+0.01), \\ 
            \tiny DRD2=0.07 (+0.01), QED=0.17 (+0.02)
        \end{minipage}
        \caption{{\LlaSMolM}}
        \label{fig:BDQ_llasmol_case_2}
    \end{subfigure}
    
    \caption{Another optimization case on \BDQ. Modifications are highlighted in red.}
    \label{fig:BDQ_case_2}
\end{figure}

In another case shown in Figure~\ref{fig:BDQ_mollm_case_2}, \mollmSixGenM improves the properties by replacing the hydrophobic naphthalene group in the hit molecule with a nitrogen-containing heteroaromatic ring resembling pyrazine (highlighted structures).
The naphthalene group increases hydrophobicity~\cite{bouiahya2020hydrophobic} but may disrupt the balance between lipophilicity and polarity required for efficient BBB penetration~\cite{jimenez2024molecular, cornelissen2023explaining}, while also posing potential metabolic stability concerns~\cite{kiani2019lipophilic}. 
Replacing it with a nitrogen-containing heterocyclic ring fine-tunes this lipophilicity-polarity balance, a well-established medicinal chemistry strategy to improve permeability and metabolic stability~\cite{kerru2020review}.  
This modification enables \mollmSixGenM to achieve notable improvements in BBBP (+0.41), DRD2 (+0.41), and QED (+0.11). Notably, the hit molecule’s BBBP is 0.51, a decent but not maximal value, making further optimization a nontrivial task.  
In contrast, \LlaSMolM’s adjustment is a minor local change to a single bond near the core structure (Figure~\ref{fig:BDQ_llasmol_case_2}), which neither meaningfully shifts polarity and lipophilicity nor reduces structural complexity, resulting in only negligible property improvements.


%
\section{Discussion on \PMol}
\label{sec:app:baselines}
\begin{table*}[h!]
\centering
\caption{Training Details of \PMol}
\label{tbl:pmol_training}
\begin{small}
\begin{threeparttable}
\begin{tabular}{
    @{\hspace{3pt}}l@{\hspace{3pt}}
    @{\hspace{3pt}}l@{\hspace{10pt}}
    @{\hspace{3pt}}l@{\hspace{3pt}}
    @{\hspace{3pt}}l@{\hspace{3pt}}
    @{\hspace{3pt}}l@{\hspace{3pt}}
}
\toprule
\multirow{2}{*}{Task} 
& \multicolumn{2}{c}{MGA Training} 
& \multicolumn{2}{c}{Transformer Training} 
\\
\cmidrule(lr){2-3} \cmidrule(lr){4-5}
& Data (\# of uni mols) & Optimal parameters & Data (\# of mol pairs) & Optimal parameters \\
\midrule
\BDQ & 3,691 & Epoch: 64 & 4,472 & Epoch: 73 \\
\BPQ & 5,562 & Epoch: 81 & 4,048 & Epoch: 113 \\
\BDP & 1,959 & Epoch: 115 & 2,064 & Epoch: 116 \\
\DPQ & 2,071 & Epoch: 76 & 2,114 & Epoch: 99 \\
\BDPQ & 641 & Epoch: 166 & 624 & Epoch: 249 \\
\hline
\end{tabular}
\end{threeparttable}
\end{small}
\end{table*}


\subsection{Training Details}
\label{sec:app:pmol}
We use the official implementation of \PMol, 
and follow their two-stage training process.
First, we train the multitask graph attention network (MGA) for property predictions within each task (where a task refers to multi-property optimization). This MGA is later used to generate task-specific atom embeddings for optimization guidance.
%
%
Specifically, we extract the unique molecules 
from each IND task's training and validation sets for the MGA training.
%
Secondly, we use task-specific molecule pairs, with atom embeddings provided generated by the trained MGA, to train the sequence-to-sequence transformer.
The task-specific atom embeddings are aggregated with token embeddings through summation for transformer input.
In total, \PMol has approximately 25 million parameters.
Training for each task with early stopping typically takes 1.5 hours on a single NVIDIA V100 GPU with 16GB memory,
hence totalling 7.5 GPU hours for 5 IND tasks.
Training data statistics and 
best hyper-parameters are presented in Table~\ref{tbl:pmol_training}. 
%

\subsection{Limitations}
\label{sec:app:pmol:discuss}

First, in \PMol's paper, the atom embedding produced by the MGA is claimed to be ``property-specific", implying that the embedding is trained independently for each property prediction (that is, only contains the specific property's information).
However, we find that during MGA training, this atom embedding is actually shared across all properties and is not differentiated for individual property prediction. 
This design introduces a coupling effect, where the atom embedding actually encodes information across multiple properties simultaneously.
As a result, when these embeddings are later used to guide molecular optimization towards improving a particular property, the encoded information from other properties will bias the property-specific optimization. 
Therefore, we argue that these embeddings are more accurately described as "task-specific", meaning they are only appropriate for guiding optimization under the same property combination (i.e., task) used during the transformer training.

Second, \PMol aggregates the task-specific atom embeddings and token embeddings through direct summation. 
However, this approach is problematic because the atom embeddings generated by MGA and the token embeddings reside in different latent spaces. 
Combining representations from different spaces 
directly through summation
is an ill-considered fusion strategy. 
A more principled approach would involve introducing a projection layer to align both embeddings into a common space before aggregation.

\section{Discussions on DeepSeek-R1}
\label{sec:app:deepseek}

DeepSeek-R1~\cite{deepseekai2025deepseekr1incentivizingreasoningcapability} is a recently open-sourced,
reasoning-focused LLM
trained via large-scale reinforcement learning
without relying on large amounts of supervised fine-tuning data.
Experiments demonstrated that DeepSeek-R1
has strong reasoning capabilities comparable to OpenAI-o1-1217
on tasks such as logical inference, mathematics, and coding.
Their experiments also highlight the effectiveness of distillation,
where smaller distilled models, such as Qwen2.5-14B,
outperform the larger base model QwQ-32B-Preview by a significant margin.
Given the relevance of reasoning capabilities in multi-property molecule optimization,
we chose to evaluate the distilled version of 
Llama-3.1-8B, DeepSeek-Distill-R1-Llama-8B, 
as it is the only version directly comparable to the Llama-based models in our experiments.

Following the recommendation of DeepSeek's authors in their paper,
we avoided using system prompts and few-shot prompting,
as such settings have been shown to degrade the model's performance.
Additionally, using few-shot prompts would significantly increase the cost and resource requirements, 
as it resulted in considerably longer response generation times in our preliminary demonstrations. 
Thus, we employed a zero-shot setting,
balancing efficiency and adherence to best practices.

Initially, we experimented with the same instruction template used in our general-purpose LLM evaluations.
However, DeepSeek-R1-Distill-Llama-8B 
consistently failed to optimize or generate modified molecules. 
Instead, it simply echoed the input molecule as its response, even after increasing the token limit. 
This behavior is likely because the prompt asked the model to only generate SMILES strings without explicitly mentioning step-by-step reasoning (i.e., chain-of-thought). 
Figure~\ref{fig:deepseek_1} provides an example of such a failure case.
\begin{figure*}
\begin{tcolorbox}[
  colback=lightergray, 
  colframe=black, 
  sharp corners, 
  boxrule=0.5pt, 
  width=\textwidth,
  left=1mm, 
  right=1mm, 
  top=1mm, 
  bottom=1mm 
]
\begin{lstlisting}[
  language=, 
  basicstyle=\ttfamily\footnotesize, 
  breaklines=true, 
  breakindent=0pt, % Disable indentation for wrapped lines
  showstringspaces=false, % Remove visible spaces
  xleftmargin=0pt, % Remove left margin
  xrightmargin=0pt, % Remove right margin
  aboveskip=0pt,belowskip=0pt
]
[INST]
Your task is to modify the given molecule to adjust specific molecular properties while keeping structural changes as minimal as possible. Use the examples (if provided) as a guide. Your response should only contain a valid SMILES representation of the modified molecule enclosed with <SMILES> </SMILES> tag.

Task:
%%% Input : <SMILES> CC(=O)OCC1=C(C(=O)O)N2C(=O)[C@@H](NC(=O)CCC[C@H](N)C(=O)O)[C@H]2SC1 </SMILES>
%%% Adjust: increase BBB permeability, increase DRD2 inhibition, increase Penalized octanol-water partition coefficient (penalized logP) and increase QED

[/INST]
%%% Response:
<SMILES>CC(=O)OCC1=C(C(=O)O)N2C(=O)[C@@H](NC(=O)CCC[C@H](N)C(=O)O)[C@H]2SC1</SMILES>
</br>
</br>
</br>
...
\end{lstlisting}
\end{tcolorbox}
\vspace{-10pt}
\caption{A failure case of DeepSeek-R1-Distill-Llama-8B on task \BDPQ}
\label{fig:deepseek_1}
\end{figure*}

We then modified the prompt structure, 
removing the clear separation between task instructions, input, and output, and explicitly instructing the model to "think step by step".
As shown in Figure~\ref{fig:deepseek_2}, 
this adjustment enabled DeepSeek-R1-Distill-Llama-8B to generate longer reasoning processes.
Interestingly, the model demonstrated self-reflection and explored multiple reasoning trajectories, akin to an "aha moment" expressed in Table 3 in the original paper~\cite{deepseekai2025deepseekr1incentivizingreasoningcapability}.
Despite this, the final molecule was still not optimized for the desired properties,
and the generation required significantly more tokens
-- up to 20K tokens more than those used in our instruction-tuned LLMs. 

This token inefficiency makes large-scale evaluation of DeepSeek-R1 using such prompts infeasible.
Due to these limitations, we decided not to pursue
further experiments, such as instruction-tuning on
top of DeepSeek-R1-Distill-Llama-8B or other distilled models. 
%
While the reasoning capabilities of DeepSeek-R1 are impressive, its current limitations in token efficiency, prompt sensitivity, and molecular optimization restrict its applicability in this domain.
We plan to explore this further in future work.
\begin{figure*}
\begin{tcolorbox}[
  colback=lightergray, 
  colframe=black, 
  sharp corners, 
  boxrule=0.5pt, 
  width=\textwidth,
  left=1mm, 
  right=1mm, 
  top=1mm, 
  bottom=1mm 
]
\begin{lstlisting}[
  language=, 
  basicstyle=\ttfamily\footnotesize, 
  breaklines=true, 
  breakindent=0pt, % Disable indentation for wrapped lines
  showstringspaces=false, % Remove visible spaces
  xleftmargin=0pt, % Remove left margin
  xrightmargin=0pt, % Remove right margin
  aboveskip=0pt,belowskip=0pt,
  escapeinside={(*@}{@*)}
]
Modify the molecule <SMILES> CC(=O)OCC1=C(C(=O)O)N2C(=O)[C@@H](NC(=O)CCC[C@H](N)C(=O)O)[C@H]2SC1 </SMILES> to increase its BBB permeability value, and increase its QED value. Keep the modifications to the molecule structure as minimal as possible. Think step by step and then output the modified molecule. 

%%% Response:
Okay, so I need to modify the given molecule to increase both its BBB permeability and QED value, while keeping the changes minimal. Let me start by understanding what BBB and QED mean.
...
I notice it's a cyclic structure with some substituents. Let me try to visualize it. It seems to have a central ring, maybe a thiazole or similar, with various substituents including ester groups, amides, and a sulfur atom.
...
(*@\textcolor{blue}{Another idea is to reduce the number of amide bonds.}@*) Amides are hydrophilic, but they can make the molecule bulkier and more polar, which might not always be good for BBB permeability. 

(*@\textcolor{red}{Wait, the chain is NC(=O)CCC[C@H](N)C(=O)O. That's an amide group connected...}@*)
...
(*@\textcolor{blue}{One approach could be to remove some of the ester or amide groups}@*)
...
Alternatively, maybe removing one of the carbonyl groups. 
...
(*@\textcolor{red}{Wait, but replacing a carbonyl with a CH2 would change the structure significantly. I need to make sure that the molecule still retains its functional groups necessary for its activity.}@*)
...
(*@\textcolor{blue}{Another thought: the molecule has multiple ester and amide groups,}@*)
...
So, the final modified molecule would have the NC(=O) chain shortened from four carbons to three carbons.
</think>

The molecule has been modified by shortening the NC(=O)CCC[C@H](N)C(=O)O chain to NC(=O)CC[C@H](N)C(=O)O, reducing the carbon count from four to three. This change decreases the molecular weight and increases hydrophilicity, enhancing both BBB permeability and QED.

SMILES: CC(=O)OCC1=C(C(=O)O)N2C(=O)[C@@H](NC(=O)CC[C@H](N)C(=O)O)[C@H]2SC1

\end{lstlisting}
\end{tcolorbox}
\vspace{-10pt}
\caption{"aha moment" even in the distilled models, DeepSeek-R1-Distill-Llama-8B, where it does \textcolor{red}{self-reflection} on its own and even \textcolor{blue}{explores
multiple reasoning} trajectories or chain-of-thoughts. Still fails to optimize on the same case as in Figure~\ref{fig:deepseek_1}.}
\label{fig:deepseek_2}
\end{figure*}



\end{document}